%% file: neurips_2026.tex
\documentclass{article}
\PassOptionsToPackage{numbers,sort&compress}{natbib}
\usepackage[main,final]{neurips_2026}
\usepackage{float}
\usepackage[utf8]{inputenc}
\usepackage[T1]{fontenc}
\usepackage{hyperref}
\usepackage{url}
\usepackage{booktabs}
\usepackage{amsfonts}
\usepackage{amsmath}
\usepackage{amssymb}
\usepackage{nicefrac}
\usepackage{microtype}
\usepackage{xcolor}
\usepackage{graphicx}
\usepackage{multirow}
\usepackage{subcaption}
\usepackage{pifont}
\usepackage{textcomp}
\usepackage{wrapfig}
\usepackage{tikz}
\usetikzlibrary{positioning, arrows.meta, shapes.geometric, calc}

\title{Visual Text Compression as Measure Transport}

\author{%
  Lv Tang$^{1*}$, Tianyi Zheng$^{3,*}$, Yang Liu$^4$, Bo Li$^{3,\dagger}$, Xingyu Li$^{1,2\dagger}$ \\
  Department of Electrical and Computer Engineering, University of Alberta$^1$, \\ Alberta Machine Intelligence Institute (Amii)$^2$, \\ vivo BlueImage Lab, vivo Mobile Communication Co., Ltd.$^3$, \\ School of Vehicle and Mobility, Tsinghua University$^4$ \\ 
   \texttt{luckybird1994@gmail.com, libra@vivo.com, xingyu@ualberta.ca}
}

\begin{document}
\maketitle
{\renewcommand{\thefootnote}{\fnsymbol{footnote}}%
 \footnotetext[1]{Equal contribution}%
 \footnotetext[2]{Corresponding authors}}

\begin{abstract}
Visual text compression (VTC) promises efficient long-context processing by rendering text into an image and re-encoding it with a vision-language model, often producing $3$--$20\times$ fewer decoder tokens than subword tokenization.
Yet token savings do not translate predictably into downstream utility: on some tasks the visual path matches or exceeds the text path, on others it collapses, and the compression ratio itself does not predict which regime will occur.
The missing quantity is therefore not another summary of efficiency, but a principled measure of task-relevant information loss induced by visual encoding.
We address this problem by formulating VTC in the language of measure transport.
Treating text and visual tokens as empirical probability measures, we show that the ViT patch encoder induces a push-forward map whose transport cost decomposes into a precision cost from within-patch aggregation and a coverage cost from cross-patch fragmentation.
Both terms are estimable from downstream-label-free probes.
This formulation yields two operational consequences: a downstream-label-free routing criterion that selects whether to use the visual path for a given input or benchmark instance, and a transport-informed foveation mechanism that re-encodes high-cost regions at higher resolution.
Across $24$ NLP datasets at Qwen3-4B, our label-free rule matches the per-dataset oracle on $17/24$ datasets ($70.8\%$), and improves the average task score by $+3.3\%$ with $-10.3\%$ average tokens relative to a pure-LLM.
\end{abstract}

\section{Introduction}

Visual text compression (VTC) has recently emerged as a practical route to long-context processing.
By rendering text into images and re-encoding it through a vision-language model (VLM), VTC replaces long subword-token sequences with substantially fewer visual tokens, often at compression ratios of 3--20$\times$~\cite{DBLP:journals/corr/abs-2510-18234,DBLP:journals/corr/abs-2510-17800,DBLP:journals/corr/abs-2510-18840,DBLP:journals/corr/abs-2502-00791,DBLP:journals/corr/abs-2601-10378,DBLP:journals/corr/abs-2510-18279}.
DeepSeek-OCR~\cite{DBLP:journals/corr/abs-2510-17800} and Glyph~\cite{DBLP:journals/corr/abs-2510-18234} show that VLMs can preserve strong OCR fidelity and competitive long-context modeling even under heavy compression.
These results establish the efficiency promise of VTC.
They do not resolve the more fundamental question that determines whether VTC is actually usable: once text is rendered and re-encoded, what task-relevant information is lost, and when does that loss matter?

We answer this question by systematically evaluating VTC on a diverse suite of NLP benchmarks spanning understanding, reasoning, and generation.
The outcome is sharply non-uniform (Tab.~\ref{tab:intro-4b}).
On some tasks the visual path substantially outperforms the text path, while on others it falls far below.
The amount of compression achieved has no predictive power over which outcome occurs.
Datasets with the largest token savings can degrade the most, while datasets with much smaller savings can improve by several points.
The central empirical fact is therefore not that VTC saves tokens.
It clearly does.
The central problem is that the same rendering-and-encoding pipeline helps some tasks and hurts others, with no principled way to predict which case applies.

Existing reports suggest the same instability but stop short of explaining it.
VTCBench~\cite{DBLP:journals/corr/abs-2512-15649} documents deficits in associative reasoning and long-term memory, while PixelWorld~\cite{lyu2025pixelworld} reports pronounced drops on mathematics.
Performance can collapse under semantic disruption~\cite{DBLP:journals/corr/abs-2601-03714}, and simple mean pooling can rival vision-based compression on token-level reconstruction~\cite{DBLP:journals/corr/abs-2512-03643}.
Together, these findings show that compression ratio is only an efficiency statistic: it tells us how many tokens are removed, but not what information is discarded, or when the visual path should be trusted.

\begin{table*}[t]
\centering
\setlength\tabcolsep{15pt}
\renewcommand\arraystretch{1.35}
\caption{Performance and token savings of the text path (LLM) versus the visual path (VLM) on 10 representative datasets using each task's native metric, evaluated with the Qwen3-4B model~\cite{DBLP:journals/corr/abs-2505-09388}.}
\label{tab:intro-4b}
\resizebox{\textwidth}{!}{%
\begin{tabular}{@{}llccc@{\hspace{18pt}}c@{}}
\toprule[1.5pt]
\multirow{2}{*}{Dataset} & \multirow{2}{*}{Task type} & \multirow{2}{*}{Metric}
 & \multicolumn{2}{c}{Score} & \multirow{2}{*}{\shortstack[c]{Token savings\\(\%)}} \\
\cmidrule(lr){4-5}
 & & & LLM & VLM & \\
\midrule[1.2pt]

\multicolumn{6}{@{}l}{\textit{Visual path wins}} \\
\midrule

HotpotQA~\cite{DBLP:conf/emnlp/Yang0ZBCSM18}        & Multi-hop QA            & F1       & 13.1 & 35.4 & 45.0 \\
SciFact~\cite{DBLP:conf/emnlp/WaddenLLWZCH20}       & Fact Verification       & F1       & 51.3 & 71.4 & 20.7 \\
MuSiQue~\cite{DBLP:journals/tacl/TrivediBKS22}      & Multi-hop QA            & F1       &  8.0 & 19.0 & 39.3 \\
Yahoo Answers~\cite{DBLP:conf/nips/ZhangZL15}       & Topic Classification    & Acc.     & 54.3 & 64.1 & 52.4 \\
DBpedia~\cite{DBLP:conf/nips/ZhangZL15}             & Ontology Classification & Acc.     & 88.7 & 93.8 & 35.5 \\

\midrule[1.2pt]

\multicolumn{6}{@{}l}{\textit{Text path wins}} \\
\midrule

ReCoRD~\cite{DBLP:journals/corr/abs-1810-12885}     & Cloze-style QA          & F1       & 51.3 & 26.5 & 12.7 \\
DREAM~\cite{sun2019dream}                           & Multi-choice RC         & Acc.     & 88.0 & 71.1 & 12.8 \\
IMDB~\cite{DBLP:conf/acl/MaasDPHNP11}               & Sentiment               & Acc.     & 94.6 & 82.4 & 56.0 \\
LogiQA~\cite{DBLP:conf/ijcai/LiuCLHWZ20}            & Logical Reasoning       & Acc.     & 50.2 & 38.9 & 12.8 \\
CNN/DM~\cite{DBLP:conf/nips/HermannKGEKSB15}        & Summarization           & Rouge-1  & 34.0 & 24.1 & 68.6 \\

\bottomrule[1.5pt]
\end{tabular}}
\vspace{-0.6cm}
\end{table*}

Our starting point is that this failure of compression ratio is not accidental.
Compression ratio measures savings, not task-relevant information loss.
What is missing is a structured quantity that explains how the visual encoder changes the information available to the downstream task.
We show that such a quantity arises naturally in the language of measure transport.
After text is rendered, each textual unit obtains a spatial support through its glyph region in the image.
Treating these spatially localized textual units and the resulting visual tokens as empirical probability measures, the ViT patch mechanism is not merely an opaque compressor.
It can be modeled as a push-forward map that merges nearby textual units into coarser visual units and changes the geometry in which downstream reasoning is performed.
This is the central analytical move of the paper.
It turns VTC from an empirical compression trick into an object whose cost can be defined and estimated.

From this perspective, the transport cost induced by visual encoding admits a principled decomposition into two terms.
The first is a precision cost induced by within-patch aggregation, which smooths fine-grained lexical distinctions.
The second is a coverage cost induced by cross-patch fragmentation, which disperses evidence that a task may need to integrate across positions or pages.
Both terms are estimable from downstream-label-free probes.
Together, they provide the missing explanatory quantity behind the contradictions in Tab.~\ref{tab:intro-4b}.
This transport-cost model does more than explain the phenomenon.
Because the cost is label-free and input-computable, it yields operational tools directly.
We derive a downstream-label-free decision criterion that selects whether the visual path should be used for a given input or task instance.
We also derive a transport-informed foveation mechanism that identifies high-cost regions and selectively re-encodes them at higher resolution.
The routing rule and the foveation module are therefore not separate heuristics.
They are two consequences of the same transport-theoretic formulation.
Our contributions are listed as:

\begin{itemize}

    \item We show that VTC exhibits strong task-dependent instability across NLP benchmarks, and that compression ratio is an efficiency statistic rather than a predictor of downstream utility.

    \item We provide a transport-theoretic account of this instability by recasting ViT-based visual encoding as a push-forward map and deriving a decomposed transport cost with precision and coverage terms, both estimable without downstream labels.

    \item We show that the cost model yields downstream-label-free routing and transport-informed foveation on 24 benchmarks, where routing improves the accuracy-token tradeoff over a pure-LLM baseline and foveation provides bounded local repair within the visual path.

\end{itemize}

\begin{figure}[!t]
\centering
\includegraphics[width=\linewidth]{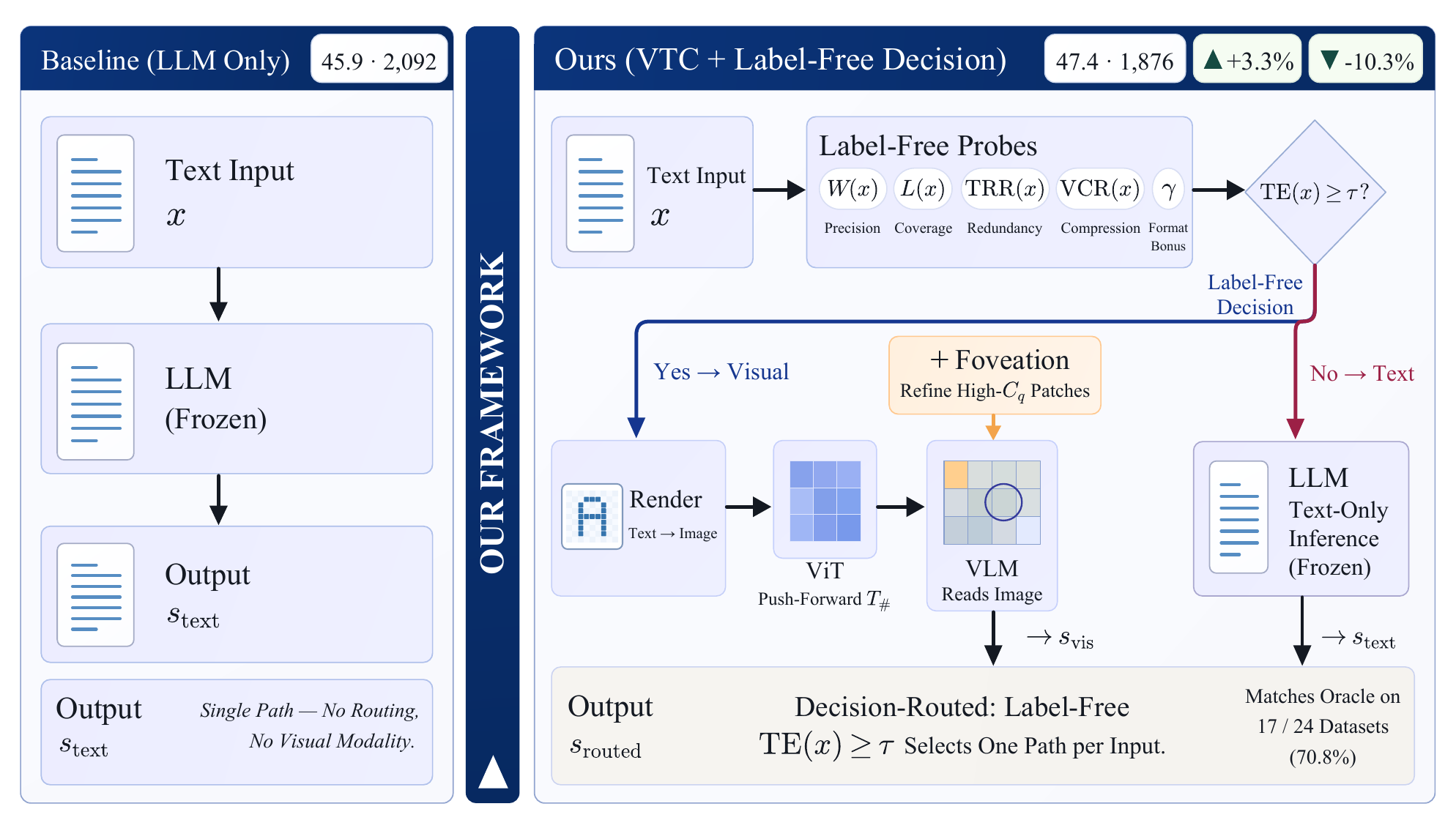}
\caption{\emph{Left}: the text-only baseline. \emph{Right}: our VTC framework. Given $x$, we first compute label-free probes $(W, L, \mathrm{TRR}, \mathrm{VCR}, \gamma)$, combine them into the transport-efficiency score $\mathrm{TE}(x)$, and route the instance under the rule $\mathrm{TE}(x) \geq \tau$ (Sec.~\ref{sec:te}). If the visual path is selected, $x$ is rendered to an image, encoded by the ViT push-forward map, and read by a VLM augmented with foveation that re-encodes high-$C_q$ patches (Sec.~\ref{sec:fov}). Across the 24-benchmark suite, the label-free rule matches the per-dataset oracle on $17/24$ datasets ($70.8\%$), raises the macro-average score from $45.9$ to $47.4$ ($+1.5$ points, $+3.3\%$ relative), and uses $10.3\%$ fewer average input tokens than pure-LLM.}
\label{fig:framework}
\vspace{-0.3cm}
\end{figure}

\section{Related Work} \label{sec:related}

Recent works treat visual re-encoding as a practical interface for reducing textual sequence length. 
Frameworks like DeepSeek-OCR~\cite{DBLP:journals/corr/abs-2510-17800} and Glyph~\cite{DBLP:journals/corr/abs-2510-18234} instantiate this idea for long-context modeling, while later variants adapt it to dense retrieval, streaming, and mixed-modality inputs~\cite{DBLP:journals/corr/abs-2510-18840,DBLP:journals/corr/abs-2502-00791,DBLP:journals/corr/abs-2601-10378,DBLP:journals/corr/abs-2510-18279}. 
The same pattern has also been repurposed beyond plain document ingestion, including long-context understanding~\cite{DBLP:journals/corr/abs-2405-14213}, CoT compression~\cite{DBLP:journals/corr/abs-2601-14750,chen2026imgcot}, and agent-memory compression~\cite{DBLP:journals/corr/abs-2601-04786}. 
A separate line of work probes the failure modes of this paradigm. VTCBench~\cite{DBLP:journals/corr/abs-2512-15649} surfaces deficits in associative reasoning and long-term memory, PixelWorld~\cite{lyu2025pixelworld} reports weakness on mathematics, semantic perturbation studies expose brittleness beyond OCR fidelity~\cite{DBLP:journals/corr/abs-2601-03714}, and reconstruction analyses show that naive pooling can rival vision-based compression on some token-level objectives~\cite{DBLP:journals/corr/abs-2512-03643}. 
What remains missing across these lines is not another application or stress test, but a structural account of the ViT-induced loss itself. Our method fills this gap by modeling visual encoding as a push-forward map and deriving a label-free transport cost that predicts when compression remains usable. Fig.~\ref{fig:framework} summarizes the resulting framework-level view of this formulation. 

\section{Method} \label{sec:method}

We begin by formalizing VTC as a push-forward map induced by the rendering-and-encoding pipeline.
For a text input $x$, the text path produces subword tokens $(x_1,\ldots,x_n)$.
Rendering assigns each token $x_i$ a spatial glyph support in the image.
The input therefore defines an empirical text measure
\begin{equation}
\mu_x \;=\; \frac{1}{n}\sum_{i=1}^{n}\delta_{x_i}.
\end{equation}
The visual path renders $x$ and encodes the image with a ViT~\cite{DBLP:conf/iclr/DosovitskiyB0WZ21}, producing visual tokens $(z_1,\ldots,z_m)$ with $m\ll n$.
We view this pipeline as a many-to-one map $T$ from localized textual units to visual-token supports.
For any measurable set $B$ in visual-token space, $T$ induces the push-forward map
\begin{equation}
(T_{\#}\mu_x)(B) \;=\; \mu_x(T^{-1}(B)).
\end{equation}

In the decoder-visible token view, suppressing preimage weights gives
\begin{equation}
\nu_x \;\approx\; \frac{1}{m}\sum_{j=1}^{m}\delta_{z_j},
\end{equation}
which keeps the visual-token support seen by the decoder.
This rendering-induced push-forward reduces token count but changes the downstream information geometry.
We therefore need a cost measuring task-relevant loss from replacing $\mu_x$ with $\nu_x$.

\subsection{Operationalized Transport Cost \texorpdfstring{$C(x)$}{C(x)}}
\label{sec:cost}

A natural choice for quantifying the transport cost induced by $T_{\#}$ is the $p$-Wasserstein distance~\cite{DBLP:journals/ftml/PeyreC19},
\begin{equation}
\label{eq:wasserstein}
W_p(\mu_x, \nu_x) \;=\; \inf_{\gamma \in \Pi(\mu_x, \nu_x)}
\left( \int \|u - v\|^p \, d\gamma(u, v) \right)^{1/p},
\end{equation}
which measures the minimum total cost of rearranging $\mu_x$ into $\nu_x$ under a chosen ground metric $\|u-v\|^p$, over all couplings $\gamma \in \Pi(\mu_x,\nu_x)$ with marginals $\mu_x$ and $\nu_x$.
This is the right mathematical object for transport, but it is not yet the right object for deployment.
Three obstacles prevent us from adopting $W_p$ directly.
First, it suffers from \textbf{\textit{ground-metric circularity}}.
A ground metric that reflects task-relevant loss must already know which lexical distinctions matter, such as ``2023'' versus ``2024'', but identifying exactly those distinctions is the purpose of $C(x)$.
Second, it suffers from \textbf{\textit{task-agnosticity}}.
The same $W_p$ value may be harmless for multiple-choice reading comprehension and catastrophic for extractive span prediction.
Third, it provides \textbf{\textit{no downstream-label-free target}}.
Any usable threshold on $W_p$ would have to be fit from labelled degradation curves, which violates our downstream-label-free requirement that no downstream label be consulted during inference.

\paragraph{Operational target.}
We use the score gap between the text path and the visual path as an operational measure of the loss introduced by the visual encoding pipeline.
Let $s_{\text{text}}(x)$ and $s_{\text{vis}}(x)$ denote the task score of the text path and the visual path on the same input $x$.
The quantity
\begin{equation}
\label{eq:operational}
\Delta(x) \;\triangleq\; s_{\text{text}}(x) - s_{\text{vis}}(x)
\end{equation}
measures the performance loss associated with replacing the text-token interface by the rendered visual-token interface.
$\Delta(x)$ resolves the first two obstacles.
It requires no ground metric, and it inherits task dependence directly from the task scores themselves.
It does not resolve the third obstacle (\textbf{\textit{no downstream-label-free target}}), because computing $\Delta(x)$ still requires labels on each path's output.
We therefore treat $\Delta(x)$ as a label-requiring ground-truth target and introduce a label-free proxy, a function $C(x)$ that is computable from the input alone and approximates $\Delta(x)$ without running either path.
During validation, labels are used only to verify that $C(x)$ tracks $\Delta(x)$.
During inference, only the proxy $C(x)$ is consulted.
We emphasize that the resulting $C(x)$ is an engineering operationalization of the transport-loss target $\Delta(x)$, not a numerical solution to the optimal-transport problem in Eq.~\ref{eq:wasserstein}.
Its functional form is motivated by the push-forward map structure derived next, but its scalars are estimated from label-free probes rather than from solving $W_p$, so all transport-cost statements in this paper refer to the proxy rather than to a computed Wasserstein distance.

\paragraph{Structure of Push-forward Map.}
To derive $C(x)$, we exploit the structure of encoder rather than fitting an unconstrained predictor.
A ViT encoder does not apply $T_{\#}$ as a single map.
It factorizes as
\begin{equation}
\label{eq:factorization}
T_{\#} \;=\; T_{\text{inter}} \circ T_{\text{intra}},
\end{equation}
where $T_{\text{intra}}$ performs local aggregation by merging the spatially localized textual units inside each image patch into one patch embedding, and $T_{\text{inter}}$ performs global arrangement by organizing the resulting patch embeddings under a 2D positional scheme.
In this second stage, attention replaces the 1D linear order available to the text path.
This factorization induces two distinct failure modes.
The first is \textbf{\textit{within-patch smoothing}}.
Fine-grained lexical distinctions inside the same patch, such as ``2023'' versus ``2024'', collapse into a shared averaged embedding.
This harms tasks that require exact lexical fidelity.
The second is \textbf{\textit{cross-patch fragmentation}}.
Evidence distributed across patches, or across rendered pages for long inputs, must now be recomposed through 2D spatial attention rather than the native 1D positional structure of the text path.
This weakens long-range evidence integration.

Because these two distortions act at different scales, we model the total transport loss additively:
\begin{equation}
\label{eq:additive}
\Delta(x) \;\approx\; \Delta_{\text{intra}}(x) + \Delta_{\text{inter}}(x),
\end{equation}
with each term depending only on the corresponding stage of the push-forward map.

\begin{figure}[!t]
\centering
\includegraphics[width=\linewidth]{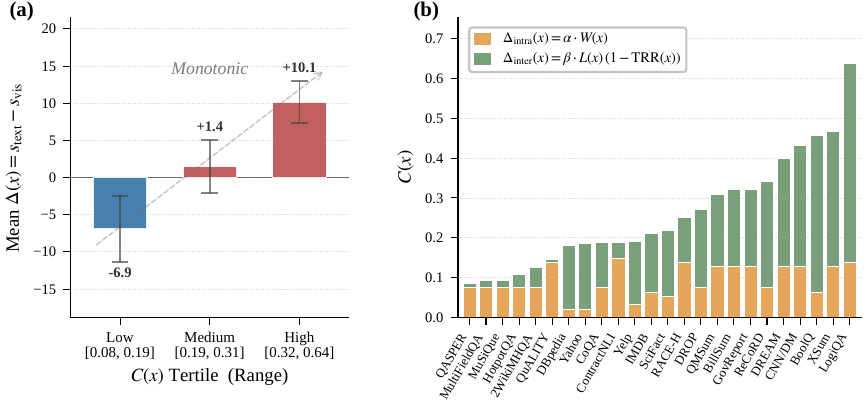}
\caption{The label-free proxy $C(x)$ tracks the labelled operational gap $\Delta(x)$ on 24 NLP benchmarks with the Qwen3-4B backbone (LLM and VLM). \textbf{(a)} Mean $\Delta(x)=s_{\mathrm{text}}-s_{\mathrm{vis}}$ within each $C(x)$ tertile, with 8 datasets per bin and error bars of $\pm 1$ standard error. $\Delta$ is reported in the native task metric, all on the same 0 to 100 scale. The mean $\Delta$ increases monotonically from $-6.9$ in the low-$C$ tertile to $+10.1$ in the high-$C$ tertile, showing that the label-free proxy orders the $C(x)$ tertiles consistently with the labelled operational gap. \textbf{(b)} Per-dataset decomposition of $C(x)$ into its intra-patch component $\Delta_{\mathrm{intra}}(x)=\alpha \cdot W(x)$ (\textcolor[HTML]{C2843A}{Orange}) and inter-patch component $\Delta_{\mathrm{inter}}(x)=\beta \cdot L(x)\,(1-\mathrm{TRR}(x))$ (\textcolor[HTML]{5B7F5E}{Green}), matching Eq.~\ref{eq:cost}. The composition varies systematically with task structure. $\Delta_{\mathrm{inter}}$ is largest on long-document summarization such as CNN/DM~\cite{DBLP:conf/nips/HermannKGEKSB15}, XSum~\cite{DBLP:conf/emnlp/NarayanCL18}, GovReport~\cite{DBLP:conf/naacl/HuangCPJW21}, and BillSum~\cite{DBLP:journals/corr/abs-1910-00523}, and on several high-coverage reasoning or classification inputs such as LogiQA~\cite{DBLP:conf/ijcai/LiuCLHWZ20} and BoolQ~\cite{DBLP:conf/naacl/ClarkLCK0T19}. $\Delta_{\mathrm{intra}}$ remains substantial on precision-sensitive tasks such as QuALITY~\cite{DBLP:conf/naacl/PangPJNPCPMT0B22}, RACE-H~\cite{DBLP:conf/emnlp/LaiXLYH17}, ContractNLI~\cite{DBLP:conf/emnlp/KoreedaM21}, and IMDB~\cite{DBLP:conf/acl/MaasDPHNP11}. This empirical separation is consistent with the factorization $T_{\#}=T_{\mathrm{inter}}\circ T_{\mathrm{intra}}$ in Eq.~\ref{eq:factorization}.}
\label{fig:cost-delta}
\vspace{-0.6cm}
\end{figure}

\paragraph{Label-free Proxy Function.}
Eq.~\ref{eq:additive} specifies the structure of the loss.
We now instantiate it as a computable proxy.
Each term is parameterized as the product of a label-free feature that measures sensitivity to the corresponding distortion and a model-specific scalar weight that measures how strongly that distortion hurts a given backbone.
This yields
\begin{equation}
\label{eq:cost}
C(x) \;=\; \underbrace{\alpha \cdot W(x)}_{\Delta_{\text{intra}}(x)} \;+\; \underbrace{\beta \cdot L(x) \cdot \bigl(1 - \mathrm{TRR}(x)\bigr)}_{\Delta_{\text{inter}}(x)}.
\end{equation}

The precision feature $W(x) \in [0, 1]$ measures sensitivity to token-level distinctions in the required answer.
It is high for extractive, generative, numerical, or option-level decisions where small lexical differences can change correctness, and low for coarse semantic classifications whose labels are robust to local lexical smoothing.
$W(x)$ is determined from the task specification alone, without consulting any sample's ground-truth label.

The coverage feature $L(x) \in [0,1]$ measures how widely task-relevant evidence is distributed across the rendered input.
To estimate it, we compute BM25~\cite{DBLP:journals/ftir/RobertsonZ09} relevance scores between the question and text segments, normalize these scores into a relevance distribution over segments, and summarize the dispersion of this relevance mass.
Higher $L(x)$ indicates that task-relevant evidence is spread across more segments, making it more vulnerable to the 2D rearrangement induced by $T_{\text{inter}}$.
The text redundancy rate $\mathrm{TRR}(x) \in [0,1]$, estimated by the gzip compression ratio~\cite{DBLP:journals/corr/abs-0809-2553} of $x$, discounts the coverage term.
When the text is highly redundant, duplicate carriers of the same information buffer fragmentation loss, so only the irreducible fraction $1-\mathrm{TRR}(x)$ contributes to the cost.

All three features are available before any downstream label is observed.
Scalar weights $\alpha$ and $\beta$ are model-specific because the severity of each distortion depends on the ViT architecture and its coupling to the decoder.
They are estimated from independent synthetic probes and never see downstream-task labels.
$C(x)$ is therefore fully label-free at deployment.
$\Delta(x)$ enters only as the validation target against which $C(x)$ is checked, never as an input to inference.

\paragraph{Empirical Check.}
Fig.~\ref{fig:cost-delta} verifies that the derived proxy tracks the labelled operational gap on the benchmark suite.
Panel~\textbf{(a)} groups 24 benchmarks into $C(x)$ tertiles and shows that the mean gap $\Delta(x)$ increases monotonically from the low-$C$ bin to the high-$C$ bin.
This confirms that $C(x)$ orders the tertiles consistently with $\Delta(x)$ without consulting downstream labels.
Panel~\textbf{(b)} decomposes $C(x)$ for every dataset and shows that the inter-patch term is largest on long-document and high-coverage inputs, while precision-sensitive tasks retain substantial intra-patch cost.
The empirical split is consistent with the theoretical factorization in Eq.~\ref{eq:factorization}, indicating that the two terms in $C(x)$ correspond to distinct failure modes of the ViT-induced push-forward map rather than to compression ratio alone.

\subsection{Path Selection via Transport Efficiency (TE)} \label{sec:te}

Path selection must balance the transport loss measured by $C(x)$ against  token savings from visual compression.
We define a TE score that combines surviving information with compression benefit.

\paragraph{Compression Benefit and Structure Bonus.}
The benefit term is the Visual Compression Ratio
\begin{equation}
\label{eq:vcr}
\mathrm{VCR}(x) \;\triangleq\; \frac{n(x)}{m(x)},
\end{equation}
where $n(x)$ is the number of text tokens produced by subword tokenization and $m(x)$ is the number of visual tokens produced by the ViT.
$\mathrm{VCR}(x) > 1$ means rendering shortens the sequence the decoder must process.
A second effect is specific to structured inputs.
When $x$ contains tabular or list layout, rendering preserves 2D spatial relations that a flat token sequence cannot express, and the visual path can exploit these relations at decode time.
We summarize this effect by a nonnegative structured-format bonus $\gamma$, estimated from an independent probe.

\setlength\intextsep{2pt}
\setlength\abovecaptionskip{2pt}
\begin{wrapfigure}{!t}{0.5\linewidth}
\centering
\includegraphics[width=\linewidth]{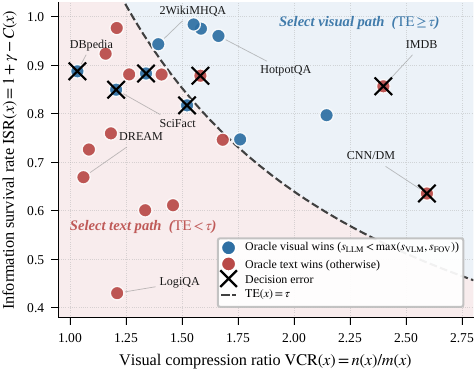}
\caption{TE decision plane at 4B scale. Each of 24 benchmarks is placed at its $(\mathrm{VCR}(x), \mathrm{ISR}(x))$ coordinates and coloured by the oracle preference. The dashed curve is the $\mathrm{TE}(x) = \tau$ contour, a hyperbola because $\mathrm{TE}(x) = \mathrm{ISR}(x) \cdot \mathrm{VCR}(x)$. The two shaded regions are the select-visual and select-text assignments of the rule $\mathrm{TE}(x) \geq \tau$, and decision errors are marked with an $\times$. For oracle labeling, the visual arm is the better of VLM and VLM+Fov, matching the upper-bound definition in Tab.~\ref{tab:main-4b}. The single scalar $\tau$ is fixed across all benchmarks. The rule is correct on $17/24$ datasets ($70.8\%$) at Qwen3-4B, raises the macro-average score from $45.9$ to $47.4$ ($+1.5$ points, $+3.3\%$ relative), and uses $10.3\%$ fewer average input tokens than the LLM-only baseline model.}
\label{fig:te-plane}
\end{wrapfigure}

\paragraph{TE.}
We normalize the text path to a baseline information value of $1$.
Relative to that baseline, the visual path loses information through cost $C(x)$ and may gain information through the structured-format bonus $\gamma$.
We call the resulting quantity the Information Survival Rate,
\begin{equation}
\label{eq:isr}
\mathrm{ISR}(x) \;\triangleq\; 1 + \gamma - C(x),
\end{equation}
which is below $1$ when transport loss dominates and above $1$ when the layout bonus dominates.
Multiplying surviving information by compression benefit gives the TE,
\begin{equation}
\begin{split}
\label{eq:te}
\mathrm{TE}(x) & \;\triangleq\; \mathrm{ISR}(x) \cdot \mathrm{VCR}(x) \\ 
&\;=\; \bigl(1 + \gamma - C(x)\bigr) \cdot \mathrm{VCR}(x),
\end{split}
\end{equation}
which measures the net value of selecting the visual path for $x$.
High $\mathrm{TE}(x)$ means that the visual path preserves enough information to justify its token savings.
Low $\mathrm{TE}(x)$ means that the push-forward map destroys more information than the compression benefit can repay.

Routing then reduces to a single threshold:
\begin{equation}
\label{eq:decision}
\text{select the visual path for } x \iff \mathrm{TE}(x) \geq \tau,
\end{equation}
where $\tau$ is a fixed decision margin that absorbs residual uncertainty in the probe-based parameter estimates.
The parameters $(\alpha, \beta, \gamma)$ are model-specific and estimated by the probes.

\paragraph{Decision geometry.}
According to Eq.~\ref{eq:te}, the threshold condition $\mathrm{TE}(x)\geq\tau$ traces a hyperbola in the $(\mathrm{VCR},\mathrm{ISR})$ plane:
\[
\mathrm{ISR}(x)=\frac{\tau}{\mathrm{VCR}(x)}.
\]
This hyperbola partitions the plane into a select-visual region above the curve and a select-text region below it.
Fig.~\ref{fig:te-plane} visualizes this partition on the 24-benchmark suite at Qwen3-4B scale.
Each dataset's position relative to the curve encodes the trade-off between transport loss and compression benefit.
The visual separation allows a single scalar $\tau$, shared across all 24 benchmarks at this scale, to carve out the two regimes without per-benchmark tuning.
The 4B operating point $\tau=1.28$ is fixed before downstream evaluation.
App.~\ref{sec:appx-rule} reports only post-hoc sensitivity diagnostics and cross-scale behavior.

\subsection{Localized Inverse Transport}
\label{sec:fov}

The scalar $C(x)$ decides whether the visual path is viable.
It does not assume that transport loss is spatially uniform.
In practice, some patches survive aggregation gracefully, while others concentrate exactly the distinctions that $T_{\text{intra}}$ washes out.
For inputs already routed to the visual path, we partially recover this local loss by re-encoding selected regions at finer resolution and appending the additional tokens to the base visual measure.
We call this foveation.
In the transport view, foveation is a localized inverse of the push-forward map:
\begin{equation}
\label{eq:nu-fov}
\nu_{\text{fov}} \;=\; \nu_{\text{base}} \;+\; \sum_{k} \Delta\nu_k.
\end{equation}
The remaining questions are where to refine, when refinement is worthwhile, and how the refinement mechanism relates to the same cost model used for routing.

\paragraph{Transport cost \texorpdfstring{$C_q$}{C\_q}.}
We spatialize the global proxy by defining, for each patch $q$ in rendered image,
\begin{equation}
\label{eq:cq}
C_q \;=\; \alpha \cdot W_q \;+\; \beta \cdot L_q \cdot (1 - \mathrm{TRR}_q),
\end{equation}
using same weights $\alpha$ and $\beta$ as in Eq.~\ref{eq:cost}, together with patch-level features that mirror their input-level counterparts.
$W_q$ is the normalized variance of ViT features inside patch $q$.
Higher variance indicates more local detail that $T_{\text{intra}}$ is likely to smooth out.
$L_q$ is the local contribution of patch $q$ to the segment-level BM25 relevance distribution used for $L(x)$.
Higher $L_q$ marks patches that are more likely to contain task-relevant evidence.
$\mathrm{TRR}_q$ is the gzip compression ratio of the text rendered inside patch $q$.
It discounts patches whose content is redundantly available elsewhere in the document.

The text-to-patch alignment is deterministic.
Rendering is a fixed function of $x$ with known fonts, page width, and line spacing, so each line of text occupies a known pixel region and maps into a known set of ViT patches.
No learned alignment is required.
The result is a cost map $\{C_q\}$ in which regions concentrating task-relevant lexical detail or dense evidence appear as hot spots.

\paragraph{Foveation trigger.}
Foveation is useful only if the information it recovers exceeds the extra token cost it introduces.
Suppose we select a set $\{k\}$ of regions, yielding a total cost recovery $\sum_k \Delta C_k$ at the price of $n_c$ additional visual tokens.
The post-foveation transport efficiency is
\begin{equation}
\label{eq:te-fov}
\mathrm{TE}_{\text{fov}}(x) \;=\; \bigl(\mathrm{ISR}(x) + \sum_{k} \Delta C_k\bigr) \cdot \frac{n_t}{n_v + n_c},
\end{equation}
where $n_t$ and $n_v$ are the text-path and base visual-path token counts.
Requiring $\mathrm{TE}_{\text{fov}}(x) > \mathrm{TE}(x)$ and simplifying gives
\begin{equation}
\label{eq:fov-trigger}
\frac{\sum_{k} \Delta C_k}{\mathrm{ISR}(x)} \;>\; \frac{n_c}{n_v}.
\end{equation}
The left-hand side is the fraction of surviving information recovered by foveation.
The right-hand side is the token overhead relative to the base visual encoding.
This inequality is the sole trigger for refinement.
It inherits all of its parameters from Sec.~\ref{sec:cost} and Sec.~\ref{sec:te}.

\paragraph{Region selection.}
Given the trigger, we still need to choose which patches to refine.
We sort patches by $C_q$ in descending order, apply non-maximum suppression at a fixed spatial radius to avoid overlapping crops, and add regions greedily until either Eq.~\ref{eq:fov-trigger} is satisfied or the crop budget is exhausted.
The crop budget caps extra tokens at $n_c \leq 0.25 \cdot n_v$, a fixed engineering constant used in all experiments.
If the greedy procedure terminates before the trigger is met, foveation is skipped and the base visual encoding is used unchanged.

\section{Experiments} \label{sec:exp}

\subsection{Experimental Setup} \label{sec:exp-setup}

We evaluate four inference strategies on the 24-benchmark suite of Sec.~\ref{sec:bench}: LLM (text-only), VLM (text rendered to an image), VLM+Fov (visual path with localized inverse transport, Sec.~\ref{sec:fov}), and Decision-routed, which selects a path using $\mathrm{TE}(x) \geq \tau$ with $\tau = 1.28$ fixed before downstream evaluation (Sec.~\ref{sec:te}). Downstream outcomes are used only for the post-hoc sensitivity analysis in App.~\ref{sec:appx-rule}, not to choose the reported operating point.
We also report an Oracle-routed upper bound (UB) that picks the better of LLM and the best visual arm post hoc. We use the Qwen3 Instruct model series as the shared backbone because it provides LLMs and VLMs at matched parameter counts, making two paths directly comparable. We run experiments at 4B, 8B, and 32B. \textbf{\textit{Due to space constraints, the main text reports the 4B results, while the 8B and 32B results are deferred to the App. \ref{sec:appx-crossscale}.}} All experiments are run on NVIDIA L40S GPUs. Scores are per-dataset means of the metric listed in Tab.~\ref{tab:bench-metadata} (App. \ref{sec:bench}). Tokens are mean input tokens per sample. We report macro-averages within the two scenarios S2 (visual-friendly, $n\!=\!10$) and S3 (text-friendly, $n\!=\!14$).

\subsection{Main Results} \label{sec:exp-main}

Tab.~\ref{tab:main-4b} reports the 4B results, aggregated over S2, S3, and all 24 benchmarks. \textbf{\textit{The per-dataset breakdown and finer-grained analysis stratified by scenario sub-types are deferred to the appendix.}} The discussion below is organized around four claims that match the theoretical structure of Sec.~\ref{sec:method}.

\begin{table}[t]
\centering
\small
\renewcommand\arraystretch{1.2}
\caption{Main results at the Qwen3-4B scale. Scores are macro-averages of per-dataset metrics (Tab.~\ref{tab:bench-metadata} in App. \ref{sec:bench}). Tokens are the mean input-token count per sample. \textbf{\textit{Oracle-routed}} picks the better of LLM and the best visual arm post hoc and gives an upper bound (UB).}
\label{tab:main-4b}
\begin{tabular*}{\textwidth}{@{\extracolsep{\fill}} l ccc r c @{}}
\toprule[1.2pt]
& \multicolumn{3}{c}{Score} & & \\
\cmidrule(lr){2-4}
Method & S2 & S3 & All & Tokens & $\Delta$ vs LLM \\
\midrule
LLM                       & 34.7 & 54.0 & 45.9 & 2{,}092 & --- \\
VLM                       & 44.0 & 44.6 & 44.4 & 1{,}486 & $-1.5$\,/\,$-29.0\%$ \\
VLM + Foveation           & 44.7 & 44.8 & 44.8 & 1{,}521 & $-1.1$\,/\,$-27.3\%$ \\
\midrule
Decision-routed (Ours)    & 41.1 & 51.9 & \textbf{47.4} & \textbf{1{,}876} & $\mathbf{+1.5}\,/\,\mathbf{-10.3\%}$ \\
\midrule
\emph{Oracle-routed (UB)} & 44.7 & 54.0 & \emph{50.2} & \emph{1{,}894} & \emph{$+4.3\,/\,-9.5\%$} \\
\bottomrule[1.2pt]
\end{tabular*}
\vspace{-0.3cm}
\end{table}

\paragraph{Claim 1: Naive VTC has task-dependent transport cost.}
Across all 24 benchmarks, VLM scores $1.5$ points below LLM on average ($44.4$ vs. $45.9$) while consuming $29.0\%$ fewer tokens. The gap is highly non-uniform. On S2, the visual path outperforms LLM by $+9.3$ points, showing that rendering can preserve useful structure when relevant evidence remains visually recoverable. On S3, the visual path falls $9.4$ points below LLM, consistent with the loss of fine-grained lexical and logical cues under within-patch aggregation. This confirms the central prediction of Sec.~\ref{sec:method}: the ViT push-forward map is not uniformly beneficial or harmful, but induces a task-dependent transport cost.

\paragraph{Claim 2: Foveation recovers part of the transport cost but is insufficient alone.}
Adding localized inverse transport (Sec.~\ref{sec:fov}) lifts the visual path from $44.4$ to $44.8$, reducing the average deficit to LLM by $0.4$ points at $<\!3\%$ extra token cost ($1{,}521$ vs. $1{,}486$). The gains concentrate where a few high-$C_q$ patches carry most of the useful evidence ($+0.7$ on S2) and fade where the cost is more diffuse ($+0.2$ on S3). Foveation therefore complements the visual path by locally repairing high-cost regions, but it cannot replace a global path-selection decision.

\paragraph{Claim 3: Label-free routing converts the two paths into a net win.}
Our Decision-routed strategy, selecting the visual arm iff $\mathrm{TE}(x) \geq \tau$ with a single $\tau = 1.28$ fixed a priori, matches the oracle choice on $17/24$ datasets ($70.8\%$; Fig.~\ref{fig:te-plane}) and achieves $+1.5$ average score over LLM-only at $-10.3\%$ tokens, without consulting any training labels at inference. The gain splits as expected: on S3, the rule preserves $96\%$ of LLM's score while still reducing tokens by about $2\%$; on S2, it captures $64\%$ of the visual-path upside relative to the oracle.

\paragraph{Claim 4: A $2.8$-point headroom remains to the oracle.}
The Oracle-routed row shows the label-aware upper bound (UB) at $+4.3$ points over LLM, while our label-free rule reaches $+1.5$ points and captures about $35\%$ of this gain. The remaining $2.8$-point gap is concentrated in borderline cases where the current proxy under-estimates structural loss or benefit. This suggests that the cost model captures the dominant routing signal, but finer sample-level structure could further close the gap.

\subsection{Ablations and Sensitivity} \label{sec:exp-ablation}

\setlength\intextsep{2pt}
\setlength\abovecaptionskip{2pt}
\begin{wraptable}{r}{0.60\linewidth}
\centering
\footnotesize
\renewcommand\arraystretch{1.15}
\caption{\textbf{\textit{Path-selection predictors at 4B}}. Accuracy vs.\ the per-dataset oracle on 24 benchmarks. \textbf{\textit{Labels}} marks whether the predictor sees task labels. Supervised rows use LOOCV.}
\label{tab:ablation-predictors}
\begin{tabular*}{\linewidth}{@{\extracolsep{\fill}} l c c @{}}
\toprule[1.0pt]
Predictor & Labels & Acc. \\
\midrule
VCR-only (best $\theta$)                             & \ding{51}        & $66.7\%$ \\
$W$-only (best $\theta$)                             & \ding{51}        & $66.7\%$ \\
LogReg ($W, L, \mathrm{TRR}, \mathrm{VCR}$)          & \ding{51}        & $66.7\%$ \\
LogReg (5 features)                                  & \ding{51}        & $\mathbf{70.8\%}$ \\
DecisionTree (depth $3$)                             & \ding{51}        & $37.5\%$ \\
MLP (hidden $16$)                                    & \ding{51}        & $50.0\%$ \\
SVM-RBF                                              & \ding{51}        & $58.3\%$ \\
\midrule
\textbf{\textit{OT rule}} (ours, label-free)         & \ding{55}        & $\mathbf{70.8\%}$ \\
\bottomrule[1.0pt]
\end{tabular*}
\end{wraptable}

\paragraph{Our proposed label-free rule can match the supervised predictor.}
Tab.~\ref{tab:ablation-predictors} compares our downstream-label-free OT rule against supervised baselines trained leave-one-dataset-out (LOOCV) across 24 benchmarks, with the per-dataset oracle (``visual arm beats LLM'') as the binary label. 
Single-feature rows pick the best label-fitting threshold on the listed feature. The 4-feature classifiers take $(W, L, \mathrm{TRR}, \mathrm{VCR})$. The 5-feature row adds LLM input length. 
Our rule reaches $70.8\%$ \textbf{\textit{without labels}}, matching the best supervised method (5-feature LogReg) and showing that the transport-cost structure pre-encodes most of the label-selection signal. 
The 4-feature LogReg ties single-feature thresholds at $66.7\%$ because
the four probes carry the same per-dataset rank as $\mathrm{VCR}$ alone. 
The fifth feature lifts accuracy by distinguishing long-context inputs
whose transport benefit exceeds compression-ratio alone. 
Decision tree (depth $3$) and MLP underperform because $24$-sample LOOCV gives too little supervision for axis-aligned splits or non-linear models to beat a linear discriminant.

\setlength\intextsep{2pt}
\setlength\abovecaptionskip{2pt}
\begin{wraptable}{l}{0.40\linewidth}
\centering
\footnotesize
\renewcommand\arraystretch{1.15}
\caption{Rendering robustness at the Qwen3-4B scale for three font sizes.}
\label{tab:ablation-rendering}
\begin{tabular*}{\linewidth}{@{\extracolsep{\fill}} l c r @{}}
\toprule[1.0pt]
Font size & Scores & Tokens \\
\midrule
font10                              & 42.4                   & 1{,}185 \\
font12 (Default)  & 44.4 & 1{,}486 \\
font14                              & 45.3                   & 1{,}716 \\
\bottomrule[1.0pt]
\end{tabular*}
\end{wraptable}

\paragraph{The framework shows predictable rendering sensitivity.}
We sweep font size in $\{10,12,14\}$ at the Qwen3-4B scale while holding other rendering parameters fixed (Tab.~\ref{tab:ablation-rendering}). 
The macro score changes monotonically with font size: shrinking from $12$\,pt to $10$\,pt drops the score by $2.0$ points and reduces mean input tokens by $20\%$, while enlarging to $14$\,pt raises the score by $0.9$ points at a roughly $16\%$ token cost. 
The total variation is about $3$ points across the sweep and shows no direction reversal, indicating that the visual path follows a consistent score-token tradeoff rather than relying on a tuned rendering setting. 
This direction is consistent with the push-forward map: smaller fonts pack more characters into each ViT patch and therefore amplify lossy within-patch aggregation. 
Full details of the rendering pipeline are provided in App.~\ref{sec:appx-rendering}.

\section{Conclusion} \label{sec:conclusion}

We cast visual text compression as a measure-transport problem and showed that the ViT push-forward map decomposes into precision and coverage costs. These costs are estimated from downstream-label-free probes, yielding a scalar transport-efficiency score $\mathrm{TE}(x)$ whose threshold defines a label-free path-selection rule. The same cost model extends to the patch level through $C_q$, motivating bounded foveation that re-encodes only costly regions. Because global routing and local refinement share the same probe-derived cost parameters $(\alpha,\beta,\gamma)$, while routing alone introduces the decision threshold $\tau$, the downstream-label-free design is preserved across both stages. On 24 NLP benchmarks at the 4B scale, the rule matches the per-dataset oracle on $17/24$ datasets ($70.8\%$), improves the macro-average score by $+1.5$ points ($+3.3\%$ relative), uses $10.3\%$ fewer average input tokens than a pure-LLM baseline, and matches the best supervised predictor without downstream labels. These results suggest that VTC's task-dependent contradictions arise from non-uniform transport cost rather than compression ratio alone. We hope that the modeling perspective and experimental design developed here can provide useful starting points for future research on visual text compression.

\section{Acknowledgment}
This work was supported by the National Natural Science Foundation of China (Nos. T2588101, 52572334) and Xishan-Tsinghua
Special Funding for the Deep Integration of Industry, Academia and Research. This work was also supported in part by the Canada CIFAR AI Chairs Program, the Alberta Machine Intelligence Institute (Amii), and the Natural Sciences and Engineering Research Council of Canada (NSERC)

{\small
\bibliographystyle{plainnat}
\bibliography{reference}
}

\newpage

\clearpage

\appendix

\input{Appendix/appx_overview.tex}
\input{Appendix/appx_related_ot.tex}
\input{Appendix/appx_benchmarks.tex}
\input{Appendix/appx_probes.tex}
\input{Appendix/appx_crossscale.tex}
\input{Appendix/appx_rule.tex}
\input{Appendix/appx_fov_fail.tex}
\input{Appendix/appx_rendering.tex}
\input{Appendix/appx_latency}
\input{Appendix/appx_crossarch.tex}

\end{document}

%% file: Appendix/appx_overview.tex
\section{Overview of Supplementary Experiments and Analyses}
\label{sec:appx-overview}

\noindent\textbf{The purpose of this section:}
This section provides a roadmap for the appendix.
The main paper presents the transport-cost framework, the label-free path-selection rule, and the headline 4B routing and foveation results.
The appendix opens with a positioning of the framework against prior optimal-transport work in deep learning, then extends the empirical evidence by documenting the evaluation protocol, probe calibration, cross-scale behavior, foveation limitations, rendering sensitivity, runtime measurements, and cross-architecture validation.
Together, these analyses are intended to support the \textbf{\textit{reproducibility}}, \textbf{\textit{downstream-label-free design}}, and \textbf{\textit{transport-cost interpretation}} of the main results.

\vspace{-0.5em}

\paragraph{What is documented in the appendix.}
Beyond this overview, the supplementary material is organized into the following technical sections, each addressing a distinct concern that a careful reader may have after the main text.

App.~\ref{sec:appx-related-ot} situates the measure-transport formulation within the broader use of optimal transport in deep learning, organizes prior work into four families (cross-modal alignment, visual-token selection, MoE routing, and information-bottleneck/compression), identifies the structural difference between each family and our framework, addresses two recent VTC critiques on the boundary of our claim, and summarizes what is and is not new in our contribution.
The goal is to make explicit that our framework treats OT as something to diagnose, not optimize, so that a reader familiar with any one of these families does not read our work as a variant.

App.~\ref{sec:bench} documents the 24-benchmark suite, including dataset selection, native metrics, answer formats, and the unified prompt templates used for the LLM and visual paths.
The goal is to make the headline numbers reproducible at the data layer and to rule out prompt design as a source of the observed LLM/VLM gap.

App.~\ref{sec:appx-probes} reports the construction and fits of the three downstream-label-free synthetic probes used to estimate $(\alpha, \beta, \gamma)$.
It also provides a sample-paired validation showing that the fixed cost ordering $C(x)$ predicts VLM-vs-LLM behavior on real evaluation samples.
The goal is to show that the parameters underlying the rule are not fit on the 24 downstream benchmarks, and that the resulting cost is aligned with observed visual advantage after the cost has already been fixed.

App.~\ref{sec:appx-crossscale} reports per-family theory-data validation and the full per-dataset grid at 4B, 8B, and 32B for LLM, VLM, and VLM with foveation.
The goal is to verify that the qualitative regimes implied by the cost decomposition, including intra-dominated, inter-dominated, low-cost, and mixed regimes, appear at the family and dataset levels rather than only in macro-averages.

App.~\ref{sec:appx-rule} reports threshold sensitivity, cross-scale routing diagnostics, and an external path-identification diagnostic on VTCBench.
The goal is to show that the 4B operating point is not a fragile threshold choice, to diagnose how the same transport-efficiency framework behaves under scale-dependent probe regimes at 8B and 32B, and to test subtask-level path identification on an external benchmark not used in calibration.

App.~\ref{sec:appx-fov-fail} reports per-scale foveation effects, foveation trigger statistics, and qualitative case studies.
The goal is to characterize when foveation helps, when it is limited, and how its scale dependence follows the same parameter regimes identified by the probes.

App.~\ref{sec:appx-rendering} documents the deterministic rendering protocol and reports a font-size sensitivity sweep at all three scales.
The goal is to show that the rendering setup is fully specified, applied uniformly to every benchmark, and that the visual path responds to font size in the direction predicted by within-patch aggregation rather than depending on dataset-specific rendering choices.

App.~\ref{sec:appx-latency} reports wall-clock latency and peak GPU memory for the four inference strategies on six representative datasets at the 4B scale.
The goal is to verify that the token-count savings translate into measurable runtime savings, and that the decision-routed strategy does not introduce a runtime cost that erases its accuracy gains.

App.~\ref{sec:appx-crossarch} reports a sample-paired cross-architecture replication of the 4B, 8B and 32B baseline experiments using InternVL3.5, which pairs an InternViT visual encoder with the same Qwen3 LLM backbones used in the headline runs.
The goal is to test whether the framework's task-dependent visual-versus-text pattern is specific to the Qwen3-VL family or transfers to a different visual encoder, while keeping the LLM backbone matched and using the result as a cross-architecture diagnostic of the visual encoding stack.

\paragraph{What this paper does and does not claim.}
We do not claim a state-of-the-art result on any individual benchmark in this suite.
Several included benchmarks have stronger task-specific systems in the literature, and our objective is not to compete with them at the leaderboard level.
The contribution of this paper is upstream of any specific benchmark number.
We recast visual text compression as a measure-transport problem, derive a structured transport cost with two separately estimable components, and use it to obtain a label-free path-selection rule and a transport-informed foveation mechanism that share their parameters.
The experiments in the main text and appendix are designed to test whether this framework explains the task-dependent behavior of VTC, not to set a leaderboard.
We hope that the modeling perspective and the supplementary evidence collected here provide a useful starting point for future work on visual text compression.

\paragraph{Scope and limitations.}
We summarize the deliberate scope choices that shape this paper and the resulting limitations a reader should keep in mind when interpreting the results.

First, the transport cost $C(x)$ is an engineering operationalization of the labelled gap $\Delta(x)$, not a numerical solution to the optimal-transport problem of Eq.~\ref{eq:wasserstein}.
Its functional form is motivated by the push-forward map structure of Sec.~\ref{sec:method}, but its parameters are estimated from label-free probes rather than from a Wasserstein computation.

Second, the operating point $\tau=1.28$ is selected at the Qwen3-4B scale and is not claimed as a universal cross-scale constant.
The 8B and 32B columns of App.~\ref{sec:appx-rule} are diagnostic, and report scale-specific thresholds where applicable.

Third, the decomposition $C(x)=\alpha W+\beta L(1-\mathrm{TRR})$ is the simplest two-term proxy that the push-forward factorization motivates.
We do not claim it is the only or optimal feature combination.

Fourth, the rendering protocol is a single global setting, using Roboto-Regular, $12$\,pt, and a fixed page sizing rule, applied uniformly to all benchmarks.
We do not search rendering settings per dataset, and the App.~\ref{sec:appx-rendering} sweep is a sensitivity test, not a tuning sweep.

Fifth, foveation is a localized inverse of the push-forward map and is gated by the trigger of Eq.~\ref{eq:fov-trigger}.
It is not expected to repair diffuse inter-dominated cases, such as long-document summarization tasks where evidence is distributed across many rendered regions.
On such inputs, the rule deliberately abstains rather than spending extra visual tokens when the cost map provides no localized target.
QMSum is the main exception in our suite because its query-specific evidence is often concentrated around a small number of dialogue turns, as discussed in App.~\ref{sec:appx-fov-trigger}.

Sixth, the headline experiments use a single backbone family, Qwen3-VL/Qwen3 at three scales.
A cross-architecture replication on InternVL3.5 is reported in App.~\ref{sec:appx-crossarch}.
Macros agree to within $\pm 2.5$ points and the per-task pattern shows ViT-family-specific differences on long-document QA and binary classification, so the framework's qualitative behavior transfers but exact gap magnitudes do not.
The framework's portability beyond these two ViT families is best understood as a hypothesis derived from the push-forward map structure rather than as an experimental claim of this paper.

Seventh, the multiplicative form of the routing rule, $\mathrm{TE}(x)=(1+\gamma-C(x))\cdot\mathrm{VCR}(x)$, has no explicit mechanism to detect when the text path is already near its score ceiling.
On near-ceiling, high-VCR external benchmarks such as the memory subtask of VTCBench, the compression factor can dominate the cost term and push $\mathrm{TE}(x)$ above the operating threshold even when the text path remains stronger.
App.~\ref{sec:appx-rule-vtcbench} documents this as a structural failure mode of the present rule and as a candidate for ceiling-aware extensions in future work.

\textbf{\textit{We mark these as limitations rather than weaknesses.
Each scope choice is the price of obtaining a transparent, downstream-label-free framework, and each is the natural starting point for follow-up work on visual text compression.
We hope that the modeling perspective and experimental design developed here can provide useful starting points for future research on visual text compression.}}

\clearpage

%% file: Appendix/appx_related_ot.tex
\section{Related Work on Optimal Transport and Measure-Transport Views}
\label{sec:appx-related-ot}

\noindent\textbf{The purpose of this section:}
This section clarifies how our measure-transport formulation of VTC relates to prior uses of OT in deep learning.
We do not claim novelty in OT theory, Wasserstein distances, or Sinkhorn-style solvers~\cite{DBLP:journals/ftml/PeyreC19}.
Our contribution is to identify the ViT visual encoder in VTC as a fixed push-forward map from rendered text units to visual tokens.
We then decompose the induced distortion into precision and coverage costs, and use the resulting downstream-label-free proxy to derive a path-selection rule and a foveation mechanism.
This distinction matters because much prior OT work optimizes an alignment, a pruning subset, an expert assignment, a bottleneck variable, or a model-fusion plan.
Our framework instead diagnoses when an already-fixed visual encoding channel is worth using.

\subsection{Main Families of OT in Deep Learning}
\label{sec:appx-related-ot-families}

\paragraph{OT for cross-modal alignment.}
Several works use OT to align representations across modalities.
Joint Wasserstein autoencoders learn image and text embeddings with a shared latent prior and paired supervision for cross-modal retrieval and phrase localization~\cite{mahajan2019joint}.
Graph Optimal Transport formulates cross-domain alignment between image objects and text words as graph matching, using Wasserstein distance for node matching and Gromov--Wasserstein distance for structure matching as regularizers in vision-language and language tasks~\cite{DBLP:conf/icml/ChenG0LC020}.
PLOT uses OT inside prompt learning for CLIP-like models, matching local visual features to multiple learned textual prompts with a Sinkhorn-based inner loop~\cite{DBLP:conf/iclr/0002YSLR023}.
Cross-modal OT has also been used to transfer linguistic information from pretrained language models to CTC-based ASR acoustic encoders~\cite{DBLP:conf/asru/LuSTK23}, and to improve CLIP robustness by restoring image-text alignment under perturbations at test time~\cite{zhu2025enhancing}.

These works treat OT as an alignment mechanism between two representation sets.
The optimized object is a cross-modal coupling, prompt alignment, training regularizer, or inference-time alignment score.
Our problem is not to align image and text representations into a shared embedding space.
In VTC, the text is first rendered into an image and then encoded by a fixed ViT.
The key question is what task-relevant information is lost by this fixed channel.
Thus, our transport object is the encoder-induced map $T$ and the decoder-visible visual measure $\nu_x$ derived from its push-forward, rather than a learned or solved coupling between independently produced image and text representations.

\paragraph{OT and visual-token pruning.}
Visual-token pruning methods reduce the number of visual tokens processed by a multimodal model.
DivPrune is a closely related non-OT baseline in this space: it selects a diverse subset of visual tokens to reduce redundancy and preserve representative visual information~\cite{alvar2025divprune}.
OTPrune makes the OT connection explicit by formulating pruning as distribution alignment: it minimizes a surrogate of the $2$-Wasserstein distance between the full visual-token distribution and a pruned subset, yielding a training-free token-selection criterion~\cite{chen2026otprune}.

This line is adjacent but structurally different from ours.
Token-pruning methods operate after the visual encoder has already produced visual tokens, and their objective is to approximate the full visual-token distribution with fewer tokens.
They do not decide whether the input should be read through the visual path at all.
They also do not model the text-to-vision transformation that produced the visual tokens.
Our rule acts before consulting downstream labels or path scores: from input-computable features and probe-calibrated parameters, it decides whether VTC is likely to preserve enough task-relevant information to justify using the visual path.
Our foveation module then adds local high-resolution evidence only when the same cost model estimates that local refinement can repay its token overhead.

\paragraph{OT for mixture-of-experts routing.}
Another family uses OT for token-to-expert assignment in mixture-of-experts (MoE) models.
Empirical studies of vision MoE routers compare softmax, expert-choice, token-choice, Sinkhorn, and soft-routing variants, including routers motivated by balanced assignment and OT-style constraints~\cite{liu2024routers}.
Selective Sinkhorn Routing formulates sparse token-to-expert assignment as an entropy-regularized OT problem and uses the transport map to promote balanced expert utilization without auxiliary load-balancing losses~\cite{nguyen2025selective}.

The shared word ``routing'' can be misleading here.
MoE routers choose which expert processes each token inside a model forward pass.
Our decision rule chooses between two input interfaces, text tokens and rendered visual tokens, before consulting downstream labels or path scores.
The MoE transport plan is the routing decision itself.
Our $\mathrm{TE}(x)$ rule is a diagnostic decision derived from the estimated transport cost of a fixed visual channel.
The units, timing, and objectives are therefore different: token-to-expert balancing inside a network versus input-to-path selection for VTC.

\paragraph{OT for information bottlenecks and model fusion.}
Information-bottleneck OT recasts the IB problem from a posterior-probability perspective and uses entropy-regularized OT with generalized Sinkhorn-style optimization to trace the bottleneck tradeoff~\cite{chen2023information}.
Model fusion via OT aligns neurons or layer-wise units across independently trained networks before averaging their parameters, producing a post-hoc fusion plan for model merging or compression~\cite{singh2020model}.

These works use OT to optimize a latent bottleneck or a model-parameter alignment.
Our paper does not learn a bottleneck variable, solve for an optimal compressed representation, compute mutual information, or fuse model weights.
The ViT encoder is already given.
We analyze the distortion induced by that encoder when text is rendered as pixels, and use a downstream-label-free proxy $C(x)$ to decide when the resulting visual path is worth using.

\subsection{What Our Formulation Adds}
\label{sec:appx-related-ot-position}

Across the families above, OT is usually used as an optimization tool for domain alignment, subset selection, expert-load balancing, IB-curve tracing, or model-parameter fusion.
Our use is diagnostic and operational.
We model an existing ViT patch encoder as a push-forward map and ask when its induced loss matters for downstream utility.
Three aspects separate the proposed framework from the OT uses above.

\paragraph{Encoder-induced push-forward rather than learned alignment.}
The transformation from text to visual tokens in VTC is not an abstract cross-modal matching problem.
Rendering gives each textual unit a spatial glyph support, and the ViT patch encoder aggregates these localized textual units into visual-token supports.
As formalized in Sec.~\ref{sec:method}, this rendering-and-encoding pipeline can be viewed as a many-to-one map $T$ that induces the push-forward map $T_{\#}\mu_x$.
In the decoder-visible token view, this push-forward is represented by the visual-token support measure $\nu_x$ after suppressing preimage weights.
This structure lets us factor the induced push-forward as
\[
T_{\#}=T_{\mathrm{inter}}\circ T_{\mathrm{intra}},
\]
where $T_{\mathrm{intra}}$ induces within-patch precision loss and $T_{\mathrm{inter}}$ induces cross-patch coverage fragmentation.
Prior alignment works can say whether two independently produced representation sets can be matched.
They do not provide this decomposition of the fixed VTC channel from rendered text units to visual tokens.

\paragraph{Task-aware proxy rather than task-agnostic Wasserstein distance.}
A raw Wasserstein distance depends on a ground metric and does not by itself encode which textual distinctions matter for a task.
In VTC, the same visual compression can be harmless for coarse topic classification but damaging for candidate-list cloze QA, numerical QA, or long-document summarization.
For this reason, our cost is not a numerical solution to an OT problem:
\[
C(x)=\alpha W(x)+\beta L(x)\bigl(1-\mathrm{TRR}(x)\bigr).
\]
It is an operational, task-aware proxy for the labelled gap $\Delta(x)=s_{\mathrm{text}}(x)-s_{\mathrm{vis}}(x)$.
Its form is motivated by the push-forward factorization, while the scalars $(\alpha,\beta,\gamma)$ are estimated from synthetic downstream-label-free probes rather than from downstream benchmark labels.

\paragraph{Path-level decision and patch-level repair share one cost model.}
The same probe-derived cost parameters underlie two decisions.
Globally, $\mathrm{TE}(x)=(1+\gamma-C(x))\mathrm{VCR}(x)$ selects whether to use the visual path.
Locally, the patch cost $C_q$ identifies high-cost regions for foveation when the trigger in Sec.~\ref{sec:fov} estimates that extra visual tokens can recover enough information.
This shared structure is different from a pipeline that first prunes tokens by one criterion and later routes by another.
It is also different from MoE routing, where the transport plan directly assigns tokens to experts but does not estimate whether an alternative input channel should be used.

\subsection{Adjacent VTC Critiques}
\label{sec:appx-related-ot-adjacent}

Two recent critiques of visual text interfaces help define the boundary of our claim.
One recent study~\cite{DBLP:journals/corr/abs-2512-03643} argues that rendering subword embeddings as pixels and re-encoding them can behave like an inefficient autoencoder, and that simple pooling can be competitive on token-level reconstruction objectives.
This does not contradict our claim, because our objective is not token reconstruction.
The framework predicts that the visual path should be avoided when the rendered channel destroys task-relevant distinctions, and used only when preserved layout or evidence locality outweighs transport cost.
The structured-format bonus $\gamma$ and the routing rule make this distinction explicit rather than assuming that visual compression is always useful.

Work on DeepSeek-OCR's linguistic-crutch behavior shows that high OCR or reconstruction quality does not guarantee reliable reasoning from visual text~\cite{DBLP:journals/corr/abs-2601-03714}.
Our precision term $\alpha W(x)$ targets this type of failure mode.
Tasks whose answers require lexical, numerical, or candidate-level fidelity receive higher precision cost and are more likely to be routed to the text path.
Thus, the framework does not deny these failures.
It turns them into a measurable cost that governs path selection.

\subsection{What Is and Is Not New}
\label{sec:appx-related-ot-new}

We do not claim novelty in OT itself, in Wasserstein distances as similarity measures, or in Sinkhorn-style optimization.
We also do not claim to compute a true Wasserstein distance between text and visual-token measures for each input.
The novelty is the use of the measure-transport viewpoint to identify the right diagnostic object for VTC and to turn that object into operational decisions.
Concretely, the paper contributes:
\begin{itemize}
\item A push-forward view of ViT-based VTC, with the encoder factorized into within-patch aggregation and cross-patch arrangement.
\item A downstream-label-free proxy $C(x)$ whose precision and coverage terms are separately estimable and whose ordering is validated against paired LLM/VLM behavior.
\item A path-selection rule that trades transport cost against visual-token savings and decides whether to use the visual path without downstream-label calibration.
\item A foveation mechanism that reuses the same cost parameters at the patch level, so local repair is tied to the same transport-cost structure as global routing.
\end{itemize}

This is why the closest OT literature should be read as methodological context rather than as prior versions of the same contribution.
Existing OT works optimize alignments, subsets, expert assignments, bottlenecks, or model parameters.
Our work instead diagnoses the task-relevant loss of a fixed visual text compression channel, uses this diagnosis to decide when visual compression is worth using, and applies the same cost structure to repair localized visual-token loss.

In short, the contribution is not a new OT solver, but a transport-based diagnostic framework for deciding when and where visual text compression should be trusted.

\clearpage

%% file: Appendix/appx_benchmarks.tex
\section{Benchmark Suite and Evaluation Protocol} 
\label{sec:bench}

\noindent\textbf{The purpose of this section:}
This section makes the empirical evidence and the \textbf{\textit{downstream-label-free claim}} reproducible by documenting the full evaluation suite, input statistics, answer formats used to assign $W(x)$, and the prompt protocol used for all LLM/VLM comparisons.

\subsection{Benchmark Suite} \label{sec:bench-suite}

\vspace{1.0em}

\begin{table*}[!htbp]
\centering
\footnotesize
\setlength\tabcolsep{4pt}
\renewcommand\arraystretch{1.1}
\caption{The 24-benchmark evaluation suite. \textbf{\textit{Samples}} is the number of evaluation instances used in every run reported here. \textbf{\textit{Avg. tokens}} is the mean LLM-path input length per sample, computed with the Qwen3 tokenizer and including the prompt template. \textbf{\textit{Answer format}} is the output constraint enforced by the instruction and provides the basis for assigning the precision feature $W(x)$ in Eq.~\ref{eq:cost}. Metric abbreviations: Acc.\ = accuracy, R1 = Rouge-1. Datasets are grouped by broad task family.}
\label{tab:bench-metadata}
\vspace{0.5em}
\begin{tabular*}{\textwidth}{@{\extracolsep{\fill}} l l c r r l @{}}
\toprule[1.2pt]
Dataset & Task family & Metric & Samples & Avg. tokens & Answer format \\
\midrule
RACE-H~\cite{DBLP:conf/emnlp/LaiXLYH17}                 & Multi-choice RC       & Acc.\    &  3{,}451 &     469 & One letter A / B / C / D \\
QuALITY~\cite{DBLP:conf/naacl/PangPJNPCPMT0B22}         & Multi-choice RC       & Acc.\    &  2{,}086 &  5{,}039 & One letter A / B / C / D \\
LogiQA~\cite{DBLP:conf/ijcai/LiuCLHWZ20}                & Multi-choice RC       & Acc.\    &     651 &     237 & One letter A / B / C / D \\
DREAM~\cite{sun2019dream}                               & Multi-choice RC       & Acc.\    &  2{,}041 &     235 & One letter A / B / C \\
\midrule
BoolQ~\cite{DBLP:conf/naacl/ClarkLCK0T19}               & Binary QA             & Acc.\    &  3{,}270 &     189 & Yes / No \\
IMDB~\cite{DBLP:conf/acl/MaasDPHNP11}                   & Binary sentiment      & Acc.\    & 25{,}000 &     334 & Positive / Negative \\
\midrule
DBpedia-14~\cite{DBLP:conf/nips/ZhangZL15}              & 14-way topic          & Acc.\    & 70{,}000 &     148 & One of 14 category names \\
Yahoo Answers~\cite{DBLP:conf/nips/ZhangZL15}           & 10-way topic          & Acc.\    & 60{,}000 &     200 & One of 10 category names \\
Yelp Full~\cite{DBLP:conf/nips/ZhangZL15}               & 5-way sentiment       & Acc.\    & 50{,}000 &     253 & Integer 1--5 \\
\midrule
ContractNLI~\cite{DBLP:conf/emnlp/KoreedaM21}           & 3-way NLI             & Acc.\    &  1{,}037 &  2{,}370 & Entailment / Contradiction / N.M. \\
SciFact~\cite{DBLP:conf/emnlp/WaddenLLWZCH20}           & Claim verification    & F1       &     300 &     595 & Rationale + SUPPORT / CONTRADICT \\
\midrule
HotpotQA~\cite{DBLP:conf/emnlp/Yang0ZBCSM18}            & Multi-hop QA          & F1       &  7{,}405 &  1{,}516 & Free-form short \\
MuSiQue~\cite{DBLP:journals/tacl/TrivediBKS22}          & Multi-hop QA          & F1       &  2{,}417 &  2{,}584 & Free-form short \\
2WikiMultiHopQA~\cite{DBLP:conf/coling/HoNSA20}         & Multi-hop QA          & F1       & 12{,}576 &  1{,}070 & Free-form short \\
CoQA~\cite{DBLP:journals/tacl/ReddyCM19}                & Conversational QA     & F1       &  7{,}983 &     404 & Free-form short \\
DROP~\cite{DBLP:conf/naacl/DuaWDSS019}                  & Numerical QA          & F1       &  9{,}535 &     344 & Number / span / date \\
MultiFieldQA-en~\cite{DBLP:conf/acl/BaiLZL0HDLZHDTL24}  & Long-doc. QA          & F1       &     150 &  4{,}905 & Free-form short \\
QASPER~\cite{DBLP:conf/naacl/DasigiLBCSG21}             & Scientific QA         & F1       &  1{,}451 &  4{,}228 & Yes / No / Unanswerable / free-form \\
ReCoRD~\cite{DBLP:journals/corr/abs-1810-12885}         & Cloze QA              & F1       & 10{,}000 &     341 & Entity from candidate list \\
\midrule
CNN/DailyMail~\cite{DBLP:conf/nips/HermannKGEKSB15}     & News summ.\           & R1       & 11{,}490 &     732 & Free-form summary \\
XSum~\cite{DBLP:conf/emnlp/NarayanCL18}                 & News summ.\           & R1       & 11{,}334 &     519 & Single-sentence summary \\
BillSum~\cite{DBLP:journals/corr/abs-1910-00523}        & Legal summ.\          & R1       &  3{,}269 &  2{,}080 & Free-form summary \\
GovReport~\cite{DBLP:conf/naacl/HuangCPJW21}            & Long-doc.\ summ.\     & R1       &     972 &  9{,}177 & Executive summary \\
QMSum~\cite{DBLP:conf/naacl/ZhongYYZMJACLQR21}          & Query-based summ.\    & R1       &     272 & 12{,}241 & Query-specific summary \\
\bottomrule[1.2pt]
\end{tabular*}
\vspace{0.6em}
\end{table*}

\vspace{1.0em}

\noindent\textbf{The purpose of this subsection:}
This subsection specifies 24 evaluation datasets and explains why the suite is suitable for testing our cost model: it spans \textbf{\textit{precision-sensitive}} tasks, where $W(x)$ should dominate, and \textbf{\textit{coverage-sensitive}} tasks, where $L(x)(1-\mathrm{TRR}(x))$ should dominate.

We evaluate on 24 established English NLP benchmarks spanning six broad task families:
(i) multi-choice reading comprehension,
(ii) binary classification,
(iii) multi-class classification,
(iv) natural-language inference and claim verification,
(v) short-answer question answering, and
(vi) long-document summarization.
The suite covers tasks with short, label-structured outputs, such as single letters, category names, and Yes/No answers, and longer free-form outputs, such as multi-sentence summaries and span-style answers.
This diversity is important for our cost model: the precision feature $W(x)$ in Eq.~\ref{eq:cost} depends on answer format, while the coverage feature $L(x)$ depends on how widely task-relevant evidence is distributed across the input.
Tab.~\ref{tab:bench-metadata} summarizes the task family, metric, sample count, average input length, and answer format for each dataset.

\clearpage

\subsection{Prompt Parity between the Text and Visual Paths} \label{sec:bench-prompt}

\vspace{1.0em}

\begin{table}[H]
\centering
\footnotesize
\setlength\tabcolsep{6pt}
\renewcommand\arraystretch{1.2}
\caption{LLM and VLM instruction templates using RACE-H~\cite{DBLP:conf/emnlp/LaiXLYH17} as an example. The two prompts differ only in the italicized surface pointer to the input channel. The task description and answer-format constraint are otherwise identical.}
\label{tab:prompt-parity}
\vspace{0.3em}
\begin{tabular}{@{}p{0.12\linewidth}p{0.82\linewidth}@{}}
\toprule[1.2pt]
\textbf{LLM} & \textit{Read the following passage} and question, then choose the correct answer from A, B, C, or D. Reply with only one letter: A, B, C, or D. \\
\midrule
\textbf{VLM} & \textit{Read the passage shown in the image} and the following question, then choose the correct answer from A, B, C, or D. Reply with only one letter: A, B, C, or D. \\
\bottomrule[1.2pt]
\end{tabular}
\vspace{0.2em}
\end{table}

\vspace{1.0em}

\noindent\textbf{The purpose of this subsection:}
This subsection establishes that the LLM and VLM paths use \textbf{\textit{matched task semantics}}, so the measured gap $\Delta(x)=s_{\text{text}}(x)-s_{\text{vis}}(x)$ can be interpreted as an operational estimate of visual-encoding cost rather than a prompt artifact.

Every LLM/VLM comparison in this paper uses textually parallel and semantically matched prompts.
Both paths receive the same task instruction, the same question, and the same passage content.
The controlled distinction is the channel through which the passage enters the model.
For each task family, we maintain one instruction template whose LLM and VLM variants differ only in a surface pointer to the input channel: the LLM variant introduces the passage with \textbf{\textit{``Read the following passage''}}, while the VLM variant introduces the identical passage, rendered as an image, with \textbf{\textit{``Read the passage shown in the image''}}.
An example for the RACE-H benchmark appears in Tab.~\ref{tab:prompt-parity}.
Three properties follow from this protocol and jointly support the fairness of comparisons.

\textbf{(1) Task semantics are matched.}
The instruction that defines the task, including what question to answer and what answer format to use, is identical up to the channel pointer.
Neither path receives extra guidance, label hints, or chain-of-thought scaffolding.

\textbf{(2) The underlying passage content is held fixed.}
The VLM image is rendered from exactly the same passage that the LLM receives as text.
Thus, before model-specific encoding, the two paths are given the same underlying content.
Any performance difference is not caused by changing the passage, adding evidence, or removing evidence, but by changing the representation channel through which the passage is consumed.

\textbf{(3) The controlled distinction is the visual encoding pipeline.}
Because the task instruction, question, answer constraint, and underlying passage content are held fixed, the relevant distinction between $s_{\text{text}}(x)$ and $s_{\text{vis}}(x)$ is whether the passage is consumed directly as subword tokens or first rendered and encoded through the visual pipeline modeled as the ViT push-forward map in Sec.~\ref{sec:method}.
We therefore interpret the empirical gap $\Delta(x)=s_{\text{text}}(x)-s_{\text{vis}}(x)$ as an operational measure of the cost introduced by the \textbf{\textit{visual encoding pipeline}}, which is the quantity that the proxy $C(x)$ is designed to estimate.

\clearpage

\begin{table*}[t]
\centering
\footnotesize
\setlength\tabcolsep{6pt}
\renewcommand\arraystretch{1.2}
\caption{Representative LLM-path and VLM-path instruction templates for the six broad task families. For each family, the LLM variant introduces the passage with ``\emph{Read the following $\langle$X$\rangle$}'', while the VLM variant introduces the identical passage, rendered as an image, with ``\emph{Read the $\langle$X$\rangle$ shown in the image}''. The remaining task description and answer-format constraint are otherwise matched across the two paths. The precision feature $W(x)$ in Eq.~\ref{eq:cost} reads directly from the \textbf{\textit{Answer format}} column.}
\label{tab:prompt-family}
\vspace{0.5em}
\begin{tabular}{@{}l l p{0.56\linewidth} l@{}}
\toprule[1.2pt]
Task family & Path & Representative instruction & Answer format \\
\midrule
\multirow{2}{*}{Multi-choice RC} 
& LLM 
& \emph{Read the following passage and question, then choose the correct answer from A, B, C, or D. Reply with only one letter: A, B, C, or D.} 
& \multirow{2}{*}{One letter} \\
& VLM 
& \emph{Read the passage shown in the image and the following question, then choose the correct answer from A, B, C, or D. Reply with only one letter: A, B, C, or D.} 
& \\
\midrule
\multirow{2}{*}{Yes/No QA} 
& LLM 
& \emph{Read the following passage and answer the question with Yes or No. Reply with exactly one word: Yes or No.} 
& \multirow{2}{*}{One word} \\
& VLM 
& \emph{Read the passage shown in the image and answer the question with Yes or No. Reply with exactly one word: Yes or No.} 
& \\
\midrule
\multirow{2}{*}{Multi-class} 
& LLM 
& \emph{Classify the following text into one of these 14 categories: Company, \ldots, Written Work. Reply with exactly one category name.} 
& \multirow{2}{*}{One category name} \\
& VLM 
& \emph{Classify the text shown in the image into one of these 14 categories: Company, \ldots, Written Work. Reply with exactly one category name.} 
& \\
\midrule
\multirow{2}{*}{NLI / verif.} 
& LLM 
& \emph{Given a contract and a hypothesis about the contract, determine whether the hypothesis is Entailment, Contradiction, or Not mentioned. Reply with exactly one of: Entailment, Contradiction, or Not mentioned.} 
& \multirow{2}{*}{One of 3 labels} \\
& VLM 
& \emph{Given the contract shown in the image and a hypothesis about the contract, determine whether the hypothesis is Entailment, Contradiction, or Not mentioned. Reply with exactly one of: Entailment, Contradiction, or Not mentioned.} 
& \\
\midrule
\multirow{2}{*}{Short-answer QA} 
& LLM 
& \emph{Read the following paragraphs and answer the question. The question may require combining information from multiple paragraphs. Give a concise, direct answer.} 
& \multirow{2}{*}{Free-form short} \\
& VLM 
& \emph{Read the paragraphs shown in the image and answer the question. The question may require combining information from multiple paragraphs. Give a concise, direct answer.} 
& \\
\midrule
\multirow{2}{*}{Summarization} 
& LLM 
& \emph{Summarize the following text concisely. Write a summary that captures the key points.} 
& \multirow{2}{*}{Free-form summary} \\
& VLM 
& \emph{Summarize the text shown in the image concisely. Write a summary that captures the key points.} 
& \\
\bottomrule[1.2pt]
\end{tabular}
\end{table*}

\subsection{Prompt Design Rationale} 
\label{sec:bench-rationale}

\noindent\textbf{The purpose of this subsection:}
This subsection explains how the prompts are constructed without downstream-label tuning, and why the answer-format column in Tab.~\ref{tab:bench-metadata} provides a reproducible basis for assigning the precision feature $W(x)$.

For each task, we follow two design rules.
First, the \textbf{\textit{label space}} is inherited directly from the dataset's official schema: for example, DREAM uses A/B/C because the dataset has three options, DBpedia lists all 14 official category names, and Yelp Full uses the 1--5 rating scale.
Second, the \textbf{\textit{instruction phrasing}} follows common lm-evaluation-harness~\cite{DBLP:journals/corr/abs-2405-14782} and LongBench~\cite{DBLP:conf/acl/BaiLZL0HDLZHDTL24} conventions for single-token classification outputs and concise free-form QA outputs, respectively.
These rules ensure that the label-free precision feature $W(x)$ in Eq.~\ref{eq:cost} maps cleanly from the answer-format column of Tab.~\ref{tab:bench-metadata} to a task-sensitivity score, without introducing prompt-engineering degrees of freedom tuned on downstream labels.
One LLM/VLM instruction pair per family is shown in Tab.~\ref{tab:prompt-family}, confirming that the channel-pointer substitution of Sec.~\ref{sec:bench-prompt} is applied uniformly across all six families.
The remaining datasets follow the same within-family templates, and their full instruction text and output parsers will be released with the code.

%% file: Appendix/appx_probes.tex
\section{Probe-Based Calibration of \texorpdfstring{$(\alpha, \beta, \gamma)$}{(alpha, beta, gamma)} and Sample-Paired Validation}
\label{sec:appx-probes}

\noindent\textbf{The purpose of this section:}
This section makes the calibration of the transport-cost model traceable and downstream-label-free.
The decomposition $C(x)=\alpha W(x)+\beta L(x)(1-\mathrm{TRR}(x))$ depends on calibrated scalars $(\alpha,\beta,\gamma)$ whose values are used in the main text but not derived there.
We first describe the three synthetic probes used to estimate these scalars at each backbone scale.
The probes use synthetic gold answers only inside the probe tasks and never use labels, scores, or outcomes from the 24 downstream benchmarks.
We then perform a separate post-hoc validation to test whether the resulting fixed $C(x)$ predicts paired VLM-vs-LLM behavior on real evaluation samples.
This second analysis is \textbf{\textit{validation, not calibration}}: it does not modify $C(x)$, $(\alpha,\beta,\gamma)$, the routing threshold $\tau$, or any foveation decision.
Its role is only to address the concern that $C(x)$ might be a pure task-family lookup rather than a label-free transport-cost ordering.

\subsection{Synthetic Probes for \texorpdfstring{$(\alpha, \beta, \gamma)$}{(alpha, beta, gamma)}}
\label{sec:appx-probes-design}

\noindent\textbf{The purpose of this subsection:}
This subsection defines the three synthetic probe datasets, the tier structure of each probe, the metric used per tier, and the regression fit that converts per-tier VLM-LLM gaps into one scalar.
All three probes use information-dense synthetic content, paired VLM-vs-LLM evaluation under matched prompts, and per-tier accuracy aggregated into one parameter through a no-intercept least-squares fit.
These probes are the only source of $(\alpha,\beta,\gamma)$.

\paragraph{$\alpha$ probe, generation fidelity.}
The $\alpha$ probe measures how the precision of VLM output decays as the answer becomes more specific.
We construct $80$ information-dense needles, each a short biographical fact about a person, university, year, and award.
The model performs one of three output tasks with increasing precision demand: extract a single keyword (Low, $W=0.10$), produce a one-sentence summary (Medium, $W=0.35$), or restate the full needle in roughly $50$ words (High, $W=0.65$).
Each needle is rendered into both a $1{,}000$-character and a $4{,}000$-character haystack, giving $80\times 3\times 2 = 480$ trials per model.
Low-tier responses are scored by exact-substring match, and the medium and high tiers by ROUGE-L F1 against the gold needle.
Per tier we compute the gap
\[
g_W = \max\bigl(0, 1 - \mathrm{VLM}_{\mathrm{avg}}/\mathrm{LLM}_{\mathrm{avg}}\bigr),
\]
then fit
\[
\alpha = \frac{\sum_W W \cdot g_W}{\sum_W W^2}, \qquad R^2 = 1 - \frac{\sum_W (g_W - \alpha W)^2}{\sum_W g_W^2}.
\]
This is a no-intercept least-squares fit of $g_W$ on $W$ over tier means.

\paragraph{$\beta$ probe, multi-fact aggregation.}
The $\beta$ probe measures how VLM performance degrades when the answer requires aggregating evidence dispersed across the rendered page.
We generate $60$ city-temperature aggregation tasks, such as asking which listed city had the highest temperature.
Each task has three locality tiers: a single fact placed in one location (Low, $L=0.10$), three facts scattered across the page (Medium, $L=0.40$), and five facts placed at $10\%$, $30\%$, $50\%$, $70\%$, and $90\%$ of the document (High, $L=0.70$).
Each task is rendered in two haystack lengths, giving $60\times 3\times 2 = 360$ trials per model.
Numeric answers are scored with a $\pm 15\%$ tolerance and name answers by exact match.
The fit is identical to the $\alpha$ case but with $L$ replacing $W$:
\[
\beta = \frac{\sum_L L \cdot g_L}{\sum_L L^2}, \qquad R^2 = 1 - \frac{\sum_L (g_L - \beta L)^2}{\sum_L g_L^2}.
\]

\paragraph{$\gamma$ probe, structured-format bonus.}
The $\gamma$ probe isolates the visual encoder's ability to exploit structured layout.
We generate $60$ tasks where the same set of facts about three people, their award years, and their award names is presented in two formats: a Markdown table (\emph{structured}) and a plain paragraph (\emph{flat}).
The model is asked who received their award most recently, scored by name-substring match.
Across two haystack lengths this gives $60\times 2\times 2 = 240$ trials per model.
We compute the per-format VLM-LLM accuracy ratios
\[
r_{\mathrm{struct}} = \mathrm{VLM}_{\mathrm{struct}} / \mathrm{LLM}_{\mathrm{struct}}, \qquad
r_{\mathrm{flat}} = \mathrm{VLM}_{\mathrm{flat}} / \mathrm{LLM}_{\mathrm{flat}},
\]
and define the structured bonus as
\[
\gamma = \max\bigl(0,\; r_{\mathrm{struct}} - r_{\mathrm{flat}}\bigr).
\]
Unlike $\alpha$ and $\beta$, $\gamma$ has no R$^2$ because there are only two format conditions.

\begin{table}[!htbp]
\centering
\footnotesize
\renewcommand\arraystretch{1.10}
\caption{Per-tier VLM/LLM accuracies and fitted $(\alpha,\beta)$ across model scales. ``Low/Med/High'' are the three tier values of $W$ for the $\alpha$ probe or $L$ for the $\beta$ probe. The gap $g$ is $\max(0,\,1-\mathrm{VLM}_{\mathrm{avg}}/\mathrm{LLM}_{\mathrm{avg}})$. Each tier aggregates $160$ trials per model for the $\alpha$ probe and $120$ for the $\beta$ probe. The fitted parameter is a no-intercept least-squares fit of $g$ on the tier value, with R$^2$ measuring the fraction of $g^2$ explained by the linear fit.}
\label{tab:appx-probe-pertier}
\vspace{0.5em}
\begin{tabular*}{\linewidth}{@{\extracolsep{\fill}} l c c c c c c c c @{}}
\toprule[1.2pt]
& \multicolumn{2}{c}{Low (tier $=0.10$)} & \multicolumn{2}{c}{Medium ($0.35$ / $0.40$)} & \multicolumn{2}{c}{High ($0.65$ / $0.70$)} & \multicolumn{2}{c}{Fit} \\
\cmidrule(lr){2-3} \cmidrule(lr){4-5} \cmidrule(lr){6-7} \cmidrule(lr){8-9}
Probe @ scale & VLM & LLM & VLM & LLM & VLM & LLM & param & R$^2$ \\
\midrule
$\alpha$ @ 4B  & $0.794$ & $1.000$ & $0.531$ & $0.642$ & $0.651$ & $0.690$ & $0.213$ & $0.33$ \\
$\alpha$ @ 8B  & $0.906$ & $1.000$ & $0.526$ & $0.632$ & $0.547$ & $0.764$ & $0.455$ & $0.98$ \\
$\alpha$ @ 32B & $0.944$ & $1.000$ & $0.628$ & $0.566$ & $0.843$ & $0.875$ & $\mathbf{0.053}$ & $0.35$ \\
\midrule
$\beta$ @ 4B   & $0.975$ & $1.000$ & $0.650$ & $0.883$ & $0.442$ & $0.783$ & $0.627$ & $0.99$ \\
$\beta$ @ 8B   & $0.992$ & $1.000$ & $0.800$ & $0.850$ & $0.717$ & $0.733$ & $\mathbf{0.061}$ & $0.61$ \\
$\beta$ @ 32B  & $0.992$ & $1.000$ & $0.825$ & $0.933$ & $0.742$ & $0.875$ & $0.233$ & $0.98$ \\
\bottomrule[1.2pt]
\end{tabular*}
\vspace{0.6em}
\end{table}

\begin{table}[!htbp]
\centering
\footnotesize
\renewcommand\arraystretch{1.10}
\caption{$\gamma$ probe per-format ratios and the fitted $(\alpha,\beta,\gamma)$ triplet used in $C(x)$ and $\mathrm{TE}(x)$. The ratio $r=\mathrm{VLM}/\mathrm{LLM}$ is computed separately on the structured Markdown-table condition and the flat paragraph condition. The $\gamma$ value is the non-negative excess ratio. The right block restates the calibrated triplet at each scale.}
\label{tab:appx-probe-fits}
\vspace{0.5em}
\begin{tabular*}{\linewidth}{@{\extracolsep{\fill}} l c c c c c c c @{}}
\toprule[1.2pt]
& \multicolumn{3}{c}{$\gamma$ probe ratios} & \multicolumn{4}{c}{Calibrated $(\alpha,\beta,\gamma)$} \\
\cmidrule(lr){2-4} \cmidrule(lr){5-8}
Scale & $r_{\mathrm{struct}}$ & $r_{\mathrm{flat}}$ & $\gamma$ & $\alpha$ & $\beta$ & $\gamma$ & Cost-map regime \\
\midrule
4B  & $0.961$ & $0.892$ & $0.069$ & $0.213$ & $0.627$ & $0.069$ & both terms active \\
8B  & $0.887$ & $0.889$ & $0.000$ & $0.455$ & $\mathbf{0.061}$ & $0.000$ & precision-dominated \\
32B & $1.031$ & $0.790$ & $0.241$ & $\mathbf{0.053}$ & $0.233$ & $0.241$ & coverage-dominated \\
\bottomrule[1.2pt]
\end{tabular*}
\vspace{0.6em}
\end{table}

\paragraph{Per-tier accuracies and fitted parameters.}
Tab.~\ref{tab:appx-probe-pertier} reports the per-tier VLM and LLM accuracies and the resulting gap for the $\alpha$ and $\beta$ probes at all three scales, together with the fitted $\alpha$, $\beta$, and the corresponding R$^2$.
Tab.~\ref{tab:appx-probe-fits} reports the per-format ratios for the $\gamma$ probe and the final $(\alpha,\beta,\gamma)$ triplet used in $C(x)$ and $\mathrm{TE}(x)$ throughout the rest of the paper.
The same calibrated values are used by the scale-dependent foveation analysis in App.~\ref{sec:appx-fov-fail}.

Two patterns in Tab.~\ref{tab:appx-probe-pertier} and Tab.~\ref{tab:appx-probe-fits} are worth noting because they motivate App.~\ref{sec:appx-fov-fail}.
At 8B, the $\beta$ probe yields $\beta=0.061$ with $R^2=0.61$.
The per-tier gaps are weak and non-monotonic ($0.008$, $0.059$, $0.023$), so the linear fit is small and only weakly supported by the data.
At 32B, the $\alpha$ probe yields $\alpha=0.053$ with $R^2=0.35$.
The medium-tier VLM accuracy exceeds the LLM accuracy ($0.628$ vs. $0.566$), so the corresponding gap collapses to zero and the linear fit explains little residual variance.
We use these probe-level R$^2$ values as evidence that the cost decomposition becomes scale-dependent.
That same scale dependence drives the foveation behavior analyzed in App.~\ref{sec:appx-fov-fail}.

\clearpage

\subsection{Sample-Paired Validation of \texorpdfstring{$C(x)$}{C(x)}}
\label{sec:appx-probes-instance}

\begin{table}[!htbp]
\centering
\footnotesize
\renewcommand\arraystretch{1.10}
\caption{Mean per-sample visual advantage $A(x) = s_{\mathrm{VLM}}(x) - s_{\mathrm{LLM}}(x)$ within each $C(x)$ quantile bucket on 4B suite. The analysis uses approximately $230{,}000$ paired sub-samples drawn from the 4B evaluation outputs. Win rate is the fraction of samples in the bucket where VLM strictly beats LLM.}
\label{tab:appx-probe-instance-buckets}
\vspace{0.5em}
\begin{tabular*}{\linewidth}{@{\extracolsep{\fill}} c c r r r l @{}}
\toprule[1.2pt]
Bucket & $C(x)$ range & $n$ & Mean $A$ & Win rate & Task families \\
\midrule
1 & $[0.085, 0.181)$ & $18{,}680$ & $+0.170$ & $39.3\%$ & long-context QA + multi-choice RC \\
2 & $[0.181, 0.186)$ & $70{,}000$ & $+0.052$ & $9.3\%$  & topic classification \\
3 & $[0.186, 0.191)$ & $61{,}037$ & $+0.095$ & $17.7\%$ & topic classification + NLI \\
4 & $[0.191, 0.639]$ & $79{,}948$ & $-0.085$ & $20.5\%$ & short-context classification + QA \\
\bottomrule[1.2pt]
\end{tabular*}
\vspace{0.6em}
\end{table}

\begin{table}[!htbp]
\centering
\footnotesize
\renewcommand\arraystretch{1.10}
\caption{Mean visual advantage $A(x)$ within a $3\times 3$ joint partition of samples by $\mathrm{VCR}(x)$ tertile (rows) and $C(x)$ tertile (columns). High-$C$ cells have the lowest mean visual advantage within each VCR row, showing that $C(x)$ predicts VLM-vs-LLM behavior beyond compression ratio alone. Cell counts are shown in parentheses.}
\label{tab:appx-probe-instance-vcrgrid}
\vspace{0.5em}
\begin{tabular*}{\linewidth}{@{\extracolsep{\fill}} l c c c @{}}
\toprule[1.2pt]
& Low $C$ & Mid $C$ & High $C$ \\
\midrule
Low VCR  & $+0.252$ ($n=6{,}785$)  & $+0.067$ ($n=45{,}696$) & $-0.097$ ($n=22{,}529$) \\
Mid VCR  & $+0.153$ ($n=8{,}608$)  & $+0.075$ ($n=50{,}876$) & $-0.011$ ($n=18{,}853$) \\
High VCR & $+0.046$ ($n=3{,}287$)  & $+0.074$ ($n=34{,}465$) & $-0.113$ ($n=38{,}566$) \\
\bottomrule[1.2pt]
\end{tabular*}
\vspace{0.6em}
\end{table}

\noindent\textbf{The purpose of this subsection:}
This subsection tests whether the probe-calibrated cost $C(x)$ predicts real benchmark behavior without fitting any downstream labels.
The analysis is sample-paired because it compares VLM and LLM scores on the same evaluation samples.
At the same time, the current implementation assigns one $C(x)$ value per dataset, since $W$ is task-level and $(L,\mathrm{TRR})$ are summarized at the dataset level.
The goal is therefore to test whether this \textbf{\textit{label-free dataset-level cost ordering}} remains predictive over many paired samples, rather than to claim that within-dataset sample variation is fully captured.
Crucially, this subsection is not used to calibrate the method.
It is a post-hoc validation of a cost function that has already been fixed by the synthetic probes.

\paragraph{Setup.}
We use 4B calibrated parameters $(\alpha,\beta)=(0.213, 0.627)$ from Tab.~\ref{tab:appx-probe-fits}.
These parameters are fixed before this analysis and are estimated only from synthetic probes, not from the 24 downstream benchmarks.
We then collect aligned per-sample LLM and VLM scores from the 4B evaluation outputs solely as a \textbf{\textit{post-hoc validation target}} for the cost ordering.
This yields approximately $230{,}000$ paired samples.
For each paired sample, we assign $C(x)$ from the dataset's $(W,L,\mathrm{TRR})$ triplet using the same formula as in the main method.
This assignment uses only task and input-side features, and never uses model scores or ground-truth performance outcomes.
No downstream score, including the visual advantage $A(x)$, is used to fit $C(x)$, $\alpha$, $\beta$, $\gamma$, the routing threshold $\tau$, or any foveation decision.
Equivalently, removing this validation analysis would not change any calibrated parameter, routing choice, foveation choice, or main result.
We compute the per-sample visual advantage
\[
A(x) = s_{\mathrm{VLM}}(x) - s_{\mathrm{LLM}}(x).
\]
Positive $A(x)$ means the visual path outperforms the text path, while negative $A(x)$ means the text path outperforms the visual path.

\paragraph{Spearman correlation.}
The Spearman rank correlation between $C(x)$ and $A(x)$ across all of these paired samples is
\[
\rho\bigl(C(x), A(x)\bigr) = -0.125, \qquad p < 10^{-300}.
\]
The negative sign matches the qualitative claim: higher-cost datasets have systematically lower visual advantage.
Thus, this test should be read as a \textbf{\textit{sample-paired validation of cost ordering}}, not as evidence that the current proxy captures all within-dataset variation.

\paragraph{Quantile bucketing on $C(x)$.}
We partition the samples into four quantile buckets by $C(x)$.
The four buckets are reported in Tab.~\ref{tab:appx-probe-instance-buckets}.
The lowest-$C$ bucket has the largest mean visual advantage, $+0.170$, while the highest-$C$ bucket is the only bucket with negative mean visual advantage, $-0.085$.
The middle buckets remain positive and close in magnitude, which reflects the coarse dataset-level nature of $C(x)$ in this validation.

\paragraph{Controlling for the visual compression ratio.}
A reviewer might worry that the negative correlation in Tab.~\ref{tab:appx-probe-instance-buckets} is driven entirely by the visual compression ratio $\mathrm{VCR}(x)$, since samples with different compression ratios may also have different task profiles.
To rule this out, we partition these paired samples into a $3\times 3$ grid: tertiles of $\mathrm{VCR}(x)$ along the rows and tertiles of $C(x)$ along the columns.
Tab.~\ref{tab:appx-probe-instance-vcrgrid} reports the within-cell mean visual advantage $A(x)$.
Within every VCR row, the high-$C$ cell has the lowest mean visual advantage: $-0.097$ in the low-VCR row, $-0.011$ in the mid-VCR row, and $-0.113$ in the high-VCR row.
Thus, $C(x)$ contributes information about visual advantage that is not absorbed by compression ratio.

Together, the probe-derived calibration in App.~\ref{sec:appx-probes-design} and the sample-paired validation in Tab.~\ref{tab:appx-probe-instance-buckets} and Tab.~\ref{tab:appx-probe-instance-vcrgrid} make the cost decomposition traceable end to end.
The parameters come from independent synthetic probes with measured R$^2$ values, and the resulting $C(x)$ predicts real-data visual advantage over paired samples even after controlling for visual compression ratio.
This validation does not participate in calibration.
It only tests whether the already-fixed cost ordering has the expected relationship with downstream behavior.
The decision-rule sensitivity in App.~\ref{sec:appx-rule} and the foveation limitations in App.~\ref{sec:appx-fov-fail} both use the probe calibration defined in this section.

\clearpage

%% file: Appendix/appx_crossscale.tex
\section{Per-Dataset Cross-Scale Performance}
\label{sec:appx-crossscale}

\noindent\textbf{The purpose of this section:}
This section tests whether the transport-cost explanation holds beyond the macro-averaged 4B result in the main paper.
It reports cross-scale results at Qwen3-4B, Qwen3-8B, and Qwen3-32B and asks whether the qualitative LLM/VLM preference predicted by $C(x)$ persists at the task-family and per-dataset levels.
The main paper reports macro-averaged Qwen3-4B results in Tab.~\ref{tab:main-4b}.
We first aggregate the per-dataset results by task family and compare the observed LLM/VLM gaps with the regimes predicted by the cost decomposition in Eq.~\ref{eq:cost}.
The full per-dataset grid is provided in the following subsection.

\subsection{Theory--Data Validation by Task Family} \label{sec:appx-perfamily}

\vspace{0.6em}

\begin{table}[!htbp]
\centering
\footnotesize
\caption{Per-family theory-data validation. Predicted regimes are derived from the dominant term of $C(x)$ in Eq.~\ref{eq:cost}. Pred.\ sign denotes the expected LLM/VLM gap direction. ``$+$'' means LLM wins, ``$-$'' means VLM is non-degraded or better, and ``N/A'' means no fixed prediction for the mixed regime. The $\Delta$ columns report observed family-average gaps at Qwen3-4B, Qwen3-8B, and Qwen3-32B. Positive values mean LLM beats VLM. Family $\Delta$ values are computed from full-precision LLM and VLM scores and rounded to two decimals for display. Recomputing $\Delta$ from the 1dp scores in Tab.~\ref{tab:appx-perdataset} can give values differing by up to $0.06$ across the 18 cells of this table due to per-dataset rounding accumulation.}
\label{tab:appx-perfamily}
\vspace{0.5em}
\begin{tabular*}{\textwidth}{@{\extracolsep{\fill}} l c c r r r c @{}}
\toprule[1.2pt]
Task family & Predicted regime & Pred.\ sign & 4B $\Delta$ & 8B $\Delta$ & 32B $\Delta$ & Match? \\
\midrule
Multi-choice RC & intra ($\uparrow W$) & $+$ & $+9.07$ & $+2.65$ & $+8.28$ & \ding{51} \\
Binary & intra ($\uparrow W$) & $+$ & $+9.64$ & $+5.38$ & $+5.29$ & \ding{51} \\
Multi-class & low cost & $-$ & $-2.45$ & $-3.74$ & $-0.33$ & \ding{51} \\
NLI/verif. & mixed & N/A & $-5.26$ & $+2.13$ & $+4.42$ & N/A \\
Short-answer QA & mixed & N/A & $-2.98$ & $-0.02$ & $+1.89$ & N/A \\
Summarization & inter ($\uparrow L$) & $+$ & $+4.73$ & $+2.25$ & $+1.69$ & \ding{51} \\
\midrule
All ($n\!=\!24$) & N/A & N/A & $+1.56$ & $+1.07$ & $+3.13$ & N/A \\
\bottomrule[1.2pt]
\end{tabular*}
\vspace{0.6em}
\end{table}

\vspace{1.0em}

\noindent\textbf{The purpose of this subsection:}
This subsection validates the \textbf{\textit{explanatory content}} of the cost decomposition rather than the final routing rule.
Specifically, it asks whether the dominant term in $C(x)$ predicts the sign of the LLM/VLM gap at the task-family level.

The cost decomposition
\begin{equation}
C(x)=\alpha W(x)+\beta L(x)(1-\mathrm{TRR}(x))
\end{equation}
implies different qualitative regimes across task families.
When $W(x)$ is high and $L(x)$ is small, the dominant loss is \emph{intra-patch} lexical smoothing, so the text path is expected to outperform the visual path.
When $L(x)$ is high, the dominant loss is \emph{inter-patch} fragmentation, so the text path is again expected to be favored.
When both terms are small, the visual path should remain non-degraded or better, because the induced transport cost is low enough that the visual representation remains usable.
Finally, when the two terms are comparable, the family falls into a \emph{mixed} regime, where the sign of the LLM/VLM gap is not expected to be fixed and may depend on model scale.

Tab.~\ref{tab:appx-perfamily} reports the predicted regime, the corresponding expected sign of the LLM/VLM gap, and the observed family-average gap
$\Delta_{\mathrm{LLM-VLM}}=s_{\mathrm{text}}-s_{\mathrm{vis}}$
at all three scales.
Positive $\Delta_{\mathrm{LLM-VLM}}$ means that the LLM path outperforms the VLM path, while negative values mean that the VLM path performs better.
For mixed regimes, we mark the predicted sign as N/A because the cost decomposition does not imply a fixed direction.

The observed signs align with the model's directional expectations for all four families where the theory makes a fixed prediction.
\emph{Multi-choice RC} and \emph{Binary} fall into the \emph{intra} regime.
Their outputs require exact option-level or label-level distinctions, making them sensitive to within-patch lexical smoothing.
Accordingly, the LLM path consistently outperforms the VLM path, with 4B gaps of $+9.07$ and $+9.64$ points.
\emph{Summarization} falls into the \emph{inter} regime.
Task-relevant evidence is distributed across long inputs, making cross-patch fragmentation costly.
This yields an LLM advantage of $+4.73$ points at 4B, with the same sign preserved at 8B and 32B.
In contrast, \emph{Multi-class} classification falls into the \emph{low-cost} regime.
Inputs are short, labels are coarse, and local lexical smoothing rarely changes the required category.
The VLM path therefore remains non-degraded and slightly outperforms the LLM path across all three scales.

The two families identified as \emph{mixed}, \emph{NLI/verif.} and \emph{Short-answer QA}, show bidirectional behavior as expected.
At 4B, the visual path wins by $-5.26$ and $-2.98$ points, respectively, but the gaps shrink or flip at larger scales.
This scale-dependent behavior is consistent with both components of $C(x)$ being non-negligible.
Neither intra-patch smoothing nor inter-patch fragmentation alone dominates the family, so the final LLM/VLM preference depends on the backbone's ability to recover information from the visual representation.

\subsection{Full Per-Dataset Breakdown}
\label{sec:appx-perdataset}

\begin{table*}[!htbp]
\centering
\footnotesize
\caption{Per-dataset performance at the 4B, 8B, and 32B scales for the three inference strategies considered in Sec.~\ref{sec:exp}: text-only LLM, text-rendered-as-image VLM, and VLM with localized inverse transport (Foveation, FOV). Within each dataset and scale, the best score among the three methods is boldfaced. Macro rows aggregate over $S_2$/$S_3$ (the 4B visual-friendly partition held fixed across all three scale columns; per-scale partitions are reported separately in App.~\ref{sec:appx-crossarch}) and over all 24 datasets.}
\label{tab:appx-perdataset}
\vspace{0.5em}
\begin{tabular*}{\textwidth}{@{\extracolsep{\fill}} l ccc ccc ccc @{}}
\toprule[1.2pt]
& \multicolumn{3}{c}{4B} & \multicolumn{3}{c}{8B} & \multicolumn{3}{c}{32B} \\
\cmidrule(lr){2-4} \cmidrule(lr){5-7} \cmidrule(lr){8-10}
Dataset & LLM & VLM & FOV & LLM & VLM & FOV & LLM & VLM & FOV \\
\midrule
\multicolumn{10}{@{}l}{\textit{Multi-choice RC}} \\
RACE-H & 25.7 & 25.1 & \textbf{25.9} & \textbf{25.0} & 23.9 & 24.8 & 25.6 & \textbf{26.7} & 25.7 \\
QuALITY & \textbf{67.8} & 60.5 & 60.5 & \textbf{70.3} & 67.2 & 68.2 & \textbf{81.2} & 75.3 & 75.0 \\
LogiQA & \textbf{50.2} & 38.9 & 39.3 & \textbf{43.2} & 41.3 & 41.3 & \textbf{61.6} & 42.9 & 42.9 \\
DREAM & \textbf{88.0} & 71.1 & 71.1 & \textbf{86.4} & 81.9 & 81.5 & \textbf{95.1} & 85.4 & 85.9 \\
\midrule
\multicolumn{10}{@{}l}{\textit{Binary}} \\
BoolQ & \textbf{85.4} & 78.3 & 78.3 & \textbf{86.9} & 84.2 & 83.8 & \textbf{89.0} & 87.2 & 87.6 \\
IMDB & \textbf{94.6} & 82.4 & 82.4 & \textbf{94.7} & 86.7 & 86.7 & \textbf{94.6} & 85.7 & 85.7 \\
\midrule
\multicolumn{10}{@{}l}{\textit{Multi-class}} \\
DBpedia & 88.7 & \textbf{93.8} & \textbf{93.8} & 92.2 & \textbf{96.5} & 96.5 & 95.1 & \textbf{97.3} & 97.3 \\
Yahoo & 54.3 & 64.1 & \textbf{64.3} & 59.1 & \textbf{66.7} & 66.6 & 66.5 & \textbf{67.7} & 67.6 \\
Yelp & \textbf{53.4} & 45.8 & 45.8 & \textbf{53.9} & 53.2 & 53.2 & \textbf{57.6} & 55.3 & 55.3 \\
\midrule
\multicolumn{10}{@{}l}{\textit{NLI/verif.}} \\
ContractNLI & \textbf{51.5} & 41.9 & 42.6 & \textbf{53.3} & 41.1 & 40.3 & \textbf{57.8} & 41.3 & 39.7 \\
SciFact & 51.3 & 71.4 & \textbf{72.2} & 62.4 & \textbf{70.4} & 70.4 & 67.3 & \textbf{74.9} & 74.9 \\
\midrule
\multicolumn{10}{@{}l}{\textit{Short-answer QA}} \\
HotpotQA & 13.1 & 35.4 & \textbf{36.4} & 26.1 & \textbf{37.7} & 36.1 & 17.8 & 29.3 & \textbf{33.6} \\
MuSiQue & 8.0 & 19.0 & \textbf{20.1} & 16.3 & \textbf{22.8} & 21.8 & 11.4 & 18.9 & \textbf{21.4} \\
2WikiMHQA & 13.4 & 38.2 & \textbf{38.5} & 26.6 & \textbf{44.0} & 40.7 & 19.3 & 35.7 & \textbf{39.5} \\
CoQA & \textbf{41.5} & 39.2 & 39.3 & \textbf{52.3} & 48.6 & 48.8 & \textbf{41.6} & 25.3 & 26.0 \\
DROP & 18.5 & \textbf{22.0} & \textbf{22.0} & \textbf{35.6} & 26.2 & 26.4 & \textbf{23.9} & 21.1 & 21.3 \\
MultiFieldQA & \textbf{46.8} & 38.8 & 39.2 & \textbf{47.0} & 44.1 & 44.4 & 42.6 & 41.7 & \textbf{43.0} \\
QASPER & 32.7 & 30.1 & \textbf{33.0} & 34.7 & \textbf{45.0} & 44.4 & 32.2 & 31.7 & \textbf{32.4} \\
ReCoRD & \textbf{51.3} & 26.5 & 27.0 & \textbf{58.7} & 29.0 & 28.5 & \textbf{76.8} & 46.8 & 46.8 \\
\midrule
\multicolumn{10}{@{}l}{\textit{Summarization}} \\
CNN/DM & \textbf{34.0} & 24.1 & 24.1 & \textbf{34.6} & 26.8 & 26.8 & \textbf{34.3} & 28.2 & 28.2 \\
XSum & \textbf{18.1} & 16.8 & 16.8 & \textbf{19.0} & 17.4 & 17.4 & \textbf{19.1} & 17.7 & 17.7 \\
BillSum & 40.9 & \textbf{40.9} & 40.9 & 38.2 & \textbf{41.2} & 41.2 & 41.7 & \textbf{42.3} & 42.3 \\
GovReport & \textbf{42.0} & 32.5 & 32.5 & \textbf{39.8} & 37.1 & 37.1 & \textbf{43.6} & 41.8 & 41.8 \\
QMSum & \textbf{30.7} & 28.0 & 28.8 & \textbf{31.0} & 29.0 & 28.9 & 29.8 & 30.1 & \textbf{30.4} \\
\midrule
\multicolumn{10}{@{}l}{\textit{Macro averages}} \\
$S_2$ at 4B ($n=10$) & 34.7 & 44.0 & 44.7 & 41.6 & 47.4 & 46.9 & 40.1 & 44.6 & 45.6 \\
$S_3$ at 4B ($n=14$) & 54.0 & 44.6 & 44.8 & 55.1 & 49.1 & 49.1 & 58.9 & 50.3 & 50.4 \\
All ($n=24$) & 45.9 & 44.4 & 44.8 & 49.5 & 48.4 & 48.2 & 51.1 & 47.9 & 48.4 \\
\bottomrule[1.2pt]
\end{tabular*}
\vspace{0.6em}
\end{table*}

\noindent\textbf{The purpose of this subsection:}
This subsection serves three purposes.
First, it exposes the per-dataset outcomes behind the macro-averages reported in the main text, showing that the aggregate trends are not driven by a small number of outliers.
Second, it tests whether the same text-friendly and visual-friendly regimes persist across model scales.
Third, it separates the role of foveation from routing.
FOV tests whether localized inverse transport can repair high-cost visual regions, while the final path decision is still handled by the transport-efficiency rule.

Tab.~\ref{tab:appx-perdataset} reports the per-dataset scores for all 24 benchmarks, grouped by task family.
Within each dataset and scale, the best score among LLM, VLM, and FOV is boldfaced, making the text-friendly and visual-friendly regimes directly visible.

\textbf{\textit{Four cross-scale patterns are worth highlighting.}}

\emph{Intra-dominated, text-favored.}
Intra-sensitive families remain text-favored at every scale.
At 4B, DREAM ($88.0$ vs $71.1$), QuALITY ($67.8$ vs $60.5$), and LogiQA ($50.2$ vs $38.9$) in \emph{Multi-choice RC}, BoolQ ($85.4$ vs $78.3$) and IMDB ($94.6$ vs $82.4$) in \emph{Binary}, ContractNLI ($51.5$ vs $41.9$) in \emph{NLI/verif.}, and ReCoRD ($51.3$ vs $26.5$) in \emph{Short-answer QA} all show LLM path beating VLM path by at least $5$ points.
The same sign is preserved at 8B/32B for every dataset listed above.
This is consistent with the precision-cost term in Eq.~\ref{eq:cost}: outputs that depend on exact option-level, label-level, or span-level distinctions are vulnerable to within-patch lexical smoothing, regardless of backbone capacity.

\emph{Low-cost or visually recoverable, visual-friendly.}
The visual path is most competitive on tasks whose answer formats tolerate local lexical smoothing or whose evidence remains visually recoverable.
At 4B, DBpedia ($93.8$ vs $88.7$) and Yahoo ($64.1$ vs $54.3$) in \emph{Multi-class}, SciFact ($71.4$ vs $51.3$) in \emph{NLI/verif.}, and HotpotQA ($35.4$ vs $13.1$), MuSiQue ($19.0$ vs $8.0$), and 2WikiMultiHopQA ($38.2$ vs $13.4$) in \emph{Short-answer QA} all show the VLM path beating the LLM path.
These cases are precisely those where rendering preserves the page-level locality structure used by the visual path, while the answer format does not require exact intra-patch lexical fidelity.

\emph{Inter-dominated summarization, text-favored.}
Long-document summarization remains text-favored even at larger scales.
At 4B, CNN/DM ($34.0$ vs $24.1$), XSum ($18.1$ vs $16.8$), and GovReport ($42.0$ vs $32.5$) favor the LLM path.
The sign is preserved at 8B and 32B.
This reflects the coverage-cost term: summarization requires integrating evidence distributed across many rendered regions, so cross-patch fragmentation accumulates over the full input.
The exception is BillSum, where the VLM path is competitive at 8B and 32B, consistent with its more templated and locally compressible structure.

\emph{Foveation as bounded repair.}
Foveation provides a complementary but bounded repair mechanism.
It improves the 4B macro-average visual score from $44.4$ to $44.8$ and the 32B macro-average visual score from $47.9$ to $48.4$, with the largest gains concentrated in tasks where a small number of high-cost regions carry useful evidence: HotpotQA ($+1.0$), MuSiQue ($+1.1$), 2WikiMultiHopQA ($+0.3$), QASPER ($+2.9$), and QMSum ($+0.8$) at 4B.
At the same time, FOV does not uniformly dominate VLM, especially at 8B, confirming the role of the trigger in Sec.~\ref{sec:fov}.
Localized inverse transport is useful when recoverable local cost exceeds the additional token overhead, but it cannot replace the global path-selection decision.

\clearpage

%% file: Appendix/appx_rule.tex
\section{Decision-Rule Sensitivity and Cross-Scale Performance}
\label{sec:appx-rule}

\noindent\textbf{The purpose of this section:}
This section tests whether the transport-efficiency framework remains stable under threshold changes and model scaling.
The main paper reports the Qwen3-4B operating point used for the primary claim.
Here we provide post-hoc threshold sweeps and cross-scale diagnostics at Qwen3-4B, Qwen3-8B, and Qwen3-32B.
The analysis is intended to diagnose robustness and failure modes, not to refit the main Qwen3-4B decision rule.

\subsection{Threshold Sensitivity and Cross-Scale Diagnostics}
\label{sec:appx-rule-tau}

\begin{table}[!htbp]
\centering
\footnotesize
\renewcommand\arraystretch{1.10}
\caption{Decision-rule accuracy versus threshold $\tau$ across model scales. The 4B column reports the standard TE rule used for the main result. The 32B column uses the bounded-benefit variant defined in App.~\ref{sec:appx-rule-scale}. Bolded rows mark per-scale diagnostic peaks. The sweep shows that $\tau=1.28$ is stable for the main 4B setting, while larger-scale behavior is diagnostic of scale-dependent probe regimes.}
\label{tab:appx-tau-sweep}
\vspace{0.5em}
\begin{tabular*}{\linewidth}{@{\extracolsep{\fill}} c c c c @{}}
\toprule[1.2pt]
$\tau$ & 4B acc.\ & 8B acc.\ & 32B acc.\ \\
\midrule
$0.90$          & $16/24$ ($66.7\%$)                   & $11/24$ ($45.8\%$)                   & $9/24$ ($37.5\%$) \\
$1.20$          & $16/24$ ($66.7\%$)                   & $16/24$ ($66.7\%$)                   & $9/24$ ($37.5\%$) \\
$\mathbf{1.28}$ & $\mathbf{17/24}$ ($\mathbf{70.8\%}$) & $\mathbf{18/24}$ ($\mathbf{75.0\%}$) & $10/24$ ($41.7\%$) \\
$1.36$          & $15/24$ ($62.5\%$)                   & $15/24$ ($62.5\%$)                   & $13/24$ ($54.2\%$) \\
$1.45$          & $16/24$ ($66.7\%$)                   & $13/24$ ($54.2\%$)                   & $14/24$ ($58.3\%$) \\
$\mathbf{1.55}$ & $14/24$ ($58.3\%$)                   & $13/24$ ($54.2\%$)                   & $\mathbf{17/24}$ ($\mathbf{70.8\%}$) \\
$1.65$          & $14/24$ ($58.3\%$)                   & $13/24$ ($54.2\%$)                   & $15/24$ ($62.5\%$) \\
\bottomrule[1.2pt]
\end{tabular*}
\vspace{0.6em}
\end{table}

\vspace{1.0em}

\noindent\textbf{The purpose of this subsection:}
This subsection tests whether the main operating point $\tau=1.28$ is a fragile choice.
The sweep is a \textbf{\textit{post-hoc sensitivity analysis}}.
It evaluates how accuracy changes after the rule is fixed, and it is not used to fit $C(x)$, $(\alpha,\beta,\gamma)$, or any dataset-specific decision.

Sweeping $\tau$ on the standard $[0.90,1.40]$ grid with step $0.01$ gives 4B accuracies in $[0.58,0.71]$ and 8B diagnostic accuracies in $[0.46,0.75]$. Tab.~\ref{tab:appx-tau-sweep} additionally reports values at $\tau\in\{1.45, 1.55, 1.65\}$, included to inform the 32B bounded-benefit operating point selected in App.~\ref{sec:appx-rule-scale}.
At 4B, the near-optimum plateau with accuracy at least $16/24$ spans $\tau\in[1.12,1.31]$, and the peak plateau with accuracy $17/24$ spans $\tau\in[1.26,1.31]$.
The main operating point $\tau=1.28$ lies inside the 4B peak plateau, and the reported 4B result is therefore not a brittle threshold artifact.
We do not claim that $\tau=1.28$ transfers as a universal constant across scales.
The 8B and 32B columns of Tab.~\ref{tab:appx-tau-sweep} are diagnostic, used to characterize how the same transport-efficiency framework behaves under scale-dependent probe regimes.
Per-scale operating points are reported separately in App.~\ref{sec:appx-rule-scale}.
Tab.~\ref{tab:appx-tau-sweep} reports representative values from the sweep.

The main observation is that the 4B rule is locally flat around the chosen operating point.
A $\pm 0.03$ perturbation around $\tau=1.28$ leaves the 4B accuracy unchanged at $17/24$.
Across $\tau\in[1.20,1.30]$, the 4B rule remains at or above $16/24$.
This local flatness is a property of the 4B operating regime.
The analogous plateau at 8B and 32B is shifted, not centered at $\tau=1.28$.
This stability follows the decision geometry of Sec.~\ref{sec:te}: because $\mathrm{TE}(x)=\mathrm{ISR}(x)\cdot\mathrm{VCR}(x)$ varies smoothly with the probes, small movements of the threshold contour in Fig.~\ref{fig:te-plane} reassign only borderline benchmarks.

The 32B column is included as a stress test.
The 4B operating point does not transfer directly to 32B.
At $\tau=1.28$, the diagnostic 32B accuracy is only $10/24$.
This is consistent with the scale-dependent probe behavior in Tab.~\ref{tab:appx-params}: the small 32B precision weight $\alpha$ reduces the ability of $C(x)$ to penalize precision-sensitive inputs.
As a result, text-friendly inputs can remain above the 4B threshold.
The 32B result therefore motivates the bounded-benefit diagnostic extension in App.~\ref{sec:appx-rule-scale}, rather than contradicting the 4B rule.

\clearpage

\subsection{OT Decision Accuracy across Scales} \label{sec:appx-rule-scale}

\begin{table}[!htbp]
\centering
\footnotesize
\renewcommand\arraystretch{1.10}
\caption{OT decision-rule accuracy across model scales, with score and input-token impact relative to an always-LLM baseline. The 4B row uses the main operating point with $\tau=1.28$. The 8B row applies the same threshold as a cross-scale diagnostic. The 32B row uses the diagnostic bounded-benefit extension with $\tau_{32\mathrm{B}}=1.55$. Score columns are macro-averages of per-dataset scores under each routing strategy. Token columns are macro-averages of mean input tokens per sample.}
\label{tab:appx-rule-scale}
\vspace{0.5em}
\begin{tabular*}{\linewidth}{@{\extracolsep{\fill}} l c c c c c c c @{}}
\toprule[1.2pt]
& & & \multicolumn{2}{c}{Score} & \multicolumn{3}{c}{Input tokens} \\
\cmidrule(lr){4-5} \cmidrule(lr){6-8}
Scale & $\tau$ & OT acc.\ & LLM\,$\to$\,Routed & $\Delta$ score & LLM\,$\to$\,Routed & $\Delta$ tok & $\Delta$ tok (\%) \\
\midrule
4B  & $1.28$ & $\mathbf{17/24}$ ($\mathbf{70.8\%}$) & $45.9\to 47.4$ & $+1.5$ ($+3.3\%$) & $2092\to 1876$ & $-216$ & $-10.3\%$ \\
8B  & $1.28$ & $\mathbf{18/24}$ ($\mathbf{75.0\%}$) & $49.5\to 50.8$ & $+1.3$ ($+2.6\%$) & $2092\to 1861$ & $-231$ & $-11.0\%$ \\
32B & $1.55$ & $\mathbf{17/24}$ ($\mathbf{70.8\%}$) & $51.1\to 52.5$ & $+1.4$ ($+2.7\%$) & $2092\to 1938$ & $-154$ & $\phantom{0}-7.4\%$ \\
\bottomrule[1.2pt]
\end{tabular*}
\vspace{0.6em}
\end{table}

\vspace{0.6em}

\noindent\textbf{The purpose of this subsection:}
This subsection reports how the decision framework behaves as model scale changes.
The 4B row corresponds to the main operating point reported in the paper.
The 8B and 32B rows are diagnostic cross-scale operating points used to study how the same transport-efficiency framework behaves under scale-dependent probe regimes.

\paragraph{Scale-aware diagnostic extension at 32B.}
The main 4B operating point transfers well to the 8B diagnostic setting, but the same threshold overcommits to the visual path at 32B.
The reason is not a change in the framework itself, but a change in the probe-calibrated cost parameters.
At 32B, $\alpha=0.053$ and $\gamma=0.241$ (Tab.~\ref{tab:appx-params}), so the cost term has little precision signal and the structured-format bonus becomes large.
This pushes $\mathrm{TE}(x)$ upward on some text-friendly inputs.
To diagnose this failure mode, we use a bounded compression-benefit variant at 32B.
The standard rule is
\[
\mathrm{TE}(x)=\bigl(1+\gamma-C(x)\bigr)\cdot\mathrm{VCR}(x).
\]
For the 32B diagnostic extension, we replace $\mathrm{VCR}(x)$ by
\[
\mathrm{VCR}_{\mathrm{eff}}(x)=1+\min\bigl(\mathrm{VCR}(x)-1,\;0.30\bigr),
\]
and compute
\[
\mathrm{TE}_{32\mathrm{B}}(x)=\bigl(1+\gamma-C(x)\bigr)\cdot \mathrm{VCR}_{\mathrm{eff}}(x).
\]
This caps extreme compression benefits when the cost term has lost precision signal.
The decision rule remains a single threshold test, $\mathrm{TE}_{32\mathrm{B}}(x)\geq\tau_{32\mathrm{B}}$.
We report $\tau_{32\mathrm{B}}=1.55$ as the operating point selected from the diagnostic sweep in Tab.~\ref{tab:appx-tau-sweep}.
This extension is not part of main 4B claim.
It is used only to analyze how framework behaves when the probe parameters become scale-degenerate.

Tab.~\ref{tab:appx-rule-scale} reports the per-scale diagnostic operating point of the framework and its score-token tradeoff.
At 4B, the main operating point delivers a positive macro-average score gain of $+1.5$ points over the always-LLM baseline, while reducing macro-average input tokens by $10.3\%$.
At 8B, the diagnostic rule reaches $18/24$ under the same $\tau=1.28$ operating point, with a $+1.3$ point score gain and $11.0\%$ fewer input tokens.
At 32B, the diagnostic bounded-benefit rule recovers $17/24$ accuracy and yields a $+1.4$ point score gain at $7.4\%$ fewer input tokens.
The smaller token reduction at 32B reflects the conservative effect of bounding the compression benefit, which routes fewer datasets to the visual path than at 4B and 8B.

The 4B accuracy of $17/24$ matches the best supervised baseline reported in Tab.~\ref{tab:ablation-predictors}.
At 8B, the diagnostic rule reaches $18/24$ under the same $\tau=1.28$ operating point.
The 32B result should be read as a scale-aware diagnostic rather than a replacement for the main downstream-label-free result.
It shows that when probe parameters become degenerate, the same transport-efficiency framework identifies the source of failure and suggests a conservative correction.

\paragraph{Why does the token reduction vary non-monotonically with scale?}
The macro-average input-token reduction is $-10.3\%$ at 4B, $-11.0\%$ at 8B, and $-7.4\%$ at 32B.
This non-monotone shape reflects which datasets the rule sends to the visual arm at each scale, not a change in per-sample visual-token efficiency.
At 4B, the rule routes $9$ of $24$ datasets to the visual arm, including long-context QA and summarization datasets where each routed dataset saves several hundred tokens.
At 8B, the diagnostic rule routes the same $9$ datasets plus Yahoo, and several long-context visual encodings become slightly more compact, so the macro reduction grows to $-11.0\%$.
At 32B, the bounded-benefit rule is more conservative and routes only the four lowest-cost multi-hop QA datasets.
These routed datasets save many tokens individually, but the savings are averaged with $20$ unrouted datasets, producing the smaller aggregate reduction of $-7.4\%$.
This trend is consistent with the framework: as the LLM strengthens, fewer inputs justify the visual path, but the inputs that remain selected are those where absolute token savings are largest.

\subsection{External Path-Identification Diagnostic on VTCBench}
\label{sec:appx-rule-vtcbench}

\vspace{0.6em}

\begin{table}[!htbp]
\centering
\footnotesize
\renewcommand\arraystretch{1.10}
\caption{VTCBench subtask-level path identification under the 4B-calibrated $\tau=1.28$. LLM and VLM columns report macro-average scores under the native \texttt{contains\_all} metric. ``Better path'' is the higher-scoring of the two. The rule's majority routing matches the better path on the retrieval and reasoning subtasks. On the memory subtask, the high mean $\mathrm{VCR}$ pushes $\mathrm{TE}(x)$ above $\tau$ for nearly all samples, and the rule's majority routing disagrees with the better path. The non-tied accuracy column is reported on the subset of samples where the LLM and VLM scores strictly differ.}
\label{tab:appx-rule-vtcbench}
\vspace{0.5em}
\begin{tabular*}{\linewidth}{@{\extracolsep{\fill}} l c c c c c c c c @{}}
\toprule[1.2pt]
Subtask   & $n$    & mean $\mathrm{VCR}$ & LLM   & VLM   & rule\,$\to$\,vis\,(\%) & better path & subtask match & non-tied acc \\
\midrule
Retrieval & $800$  & $1.44$ & $99.4$ & $71.8$ & $39.6$ & text & $\checkmark$ & $39.6\%$ ($n_{\rm nt}=280$) \\
Reasoning & $800$  & $1.51$ & $29.9$ & $\phantom{0}6.6$ & $38.6$ & text & $\checkmark$ & $69.7\%$ ($n_{\rm nt}=234$) \\
Memory    & $600$  & $4.87$ & $30.7$ & $15.3$ & $96.3$ & text & $\times$     & $\phantom{0}5.8\%$ ($n_{\rm nt}=104$) \\
\bottomrule[1.2pt]
\end{tabular*}
\vspace{0.6em}
\end{table}

\vspace{0.6em}

\noindent\textbf{The purpose of this subsection:}
This subsection tests where the calibrated 4B routing rule remains directionally correct and where it fails on an external benchmark that did not participate in any parameter fit.
The goal is not to claim a routed-score improvement on VTCBench, but to diagnose whether the rule's \textbf{\textit{subtask-level decision}}, defined as the path receiving the majority of routed samples, agrees with the empirically stronger path on each subtask.

\paragraph{Setup.}
We evaluate Qwen3-4B on VTCBench~\cite{DBLP:journals/corr/abs-2512-15649} with the same 4B parameters $(\alpha,\beta,\gamma)=(0.213,0.627,0.069)$ and the same operating point $\tau=1.28$ used in the main paper.
No VTCBench label, score, or subtask outcome is used to fit $(\alpha,\beta,\gamma)$, $C(x)$, $\tau$, or any routing decision.
$\mathrm{VCR}(x)$ is computed per sample as the ratio of LLM-side text-token count to VLM-side rendered-image-token count, and $\mathrm{TE}(x)=(1+\gamma-C(x))\cdot \mathrm{VCR}(x)$ follows Sec.~\ref{sec:te}.
For each subtask we report the fraction of samples routed to the visual path, the empirically better path under macro LLM/VLM scores, whether the rule's majority routing matches that path, and the rule's accuracy on non-tied samples.
The non-tied subset contains only samples where the LLM and VLM scores strictly differ, which avoids inflating accuracy with ties at $0$ or $1$.

\paragraph{What this shows.}
On retrieval and reasoning, the same operating point used in the main paper routes the majority of samples to the empirically stronger path, without any re-tuning on VTCBench.
This is a subtask-level directional check rather than a routed-score result, since the table evaluates majority path identification and non-tied routing accuracy, not the final routed score.
The reasoning subtask also clears the non-tied accuracy threshold of $50\%$, reaching $69.7\%$.
Retrieval does not, reaching only $39.6\%$ on non-tied samples.
This means the retrieval row should be interpreted as a correct majority-path identification result, not as evidence of reliable per-sample discrimination.
The non-tied column is included precisely to make this distinction explicit.

\paragraph{Memory exposes a boundary of the multiplicative TE form.}
On memory, the rule's majority routing disagrees with the empirically stronger path.
The cause is structural rather than a result of missing threshold tuning.
The mean visual-compression ratio is $4.87$, with some samples reaching $\mathrm{VCR}>40$, so the multiplicative form $\mathrm{TE}(x)=(1+\gamma-C(x))\cdot \mathrm{VCR}(x)$ can become large even when the text path is already near its score ceiling.
A diagnostic sweep suggests that correcting memory would require $\tau\geq 3.0$, but such a threshold would also move retrieval and reasoning toward near always-text decisions.
This sweep is diagnostic only and is not used to select any reported operating point.
We therefore report memory as an external limitation of the current multiplicative TE form on near-ceiling, high-VCR text benchmarks.
This limitation is consistent with the paper's broader claim: VTC is useful only when compression benefit outweighs task-relevant transport loss, and high compression alone is not sufficient.

\clearpage

%% file: Appendix/appx_fov_fail.tex
\section{Limitations of Foveation at Scale}
\label{sec:appx-fov-fail}

\noindent\textbf{The purpose of this section:}
This section examines when transport-informed foveation helps and when it is limited.
The goal is not to present foveation as a uniformly beneficial add-on, but to test whether its gains and limitations are consistent with the same scale-dependent cost parameters that drive the global routing rule.

In Sec.~\ref{sec:fov}, foveation is defined as a localized inverse of the push-forward map, driven by the patch-level cost map
\[
C_q=\alpha W_q+\beta L_q(1-\mathrm{TRR}_q).
\]
Because this local map uses the same scalars $(\alpha,\beta)$ as the global cost $C(x)$, its effectiveness should vary with the scale-specific probe regime.
We therefore examine the calibrated probe parameters first, and then test whether the empirical FOV gains follow the same scale-dependent pattern.

\subsection{Scale-Dependent Probe Regimes}

\vspace{0.6em}

\begin{table}[!htbp]
\centering
\footnotesize
\caption{Per-scale calibrated parameters of $C(x)=\alpha W(x)+\beta L(x)(1-\mathrm{TRR}(x))$ and the structured-format bonus $\gamma$. At 8B, the coverage weight $\beta$ weakens, so $C_q$ is dominated by the precision term. At 32B, the precision weight $\alpha$ weakens, so $C_q$ is dominated by the coverage term.}
\label{tab:appx-params}
\vspace{0.5em}
\begin{tabular*}{\linewidth}{@{\extracolsep{\fill}} l c c c l @{}}
\toprule[1.2pt]
Scale & $\alpha$ (intra) & $\beta$ (inter) & $\gamma$ (struct.) & Cost map regime \\
\midrule
4B  & $0.213$ & $0.627$ & $0.069$ & both terms active \\
8B  & $0.455$ & $\mathbf{0.061}$ & $0.000$ & precision-dominated: $C_q\approx\alpha W_q$ \\
32B & $\mathbf{0.053}$ & $0.233$ & $0.241$ & coverage-dominated: $C_q\approx\beta L_q(1-\mathrm{TRR}_q)$ \\
\bottomrule[1.2pt]
\end{tabular*}
\vspace{0.6em}
\end{table}

\vspace{0.6em}

\noindent\textbf{The purpose of this subsection:}
This subsection documents the scale-specific probe parameters $(\alpha,\beta,\gamma)$ and identifies which transport-cost dimension remains reliable at each backbone scale.
These parameters determine both the global cost $C(x)$ and the patch-level foveation map $C_q$.

The downstream-label-free probes calibrate $(\alpha,\beta,\gamma)$ separately at each scale, with the values reported in Tab.~\ref{tab:appx-params}.
Two asymmetric regimes emerge beyond 4B.
At 4B, both $\alpha$ and $\beta$ are non-trivial, so the cost map carries information from both within-patch precision variance and cross-patch coverage dispersion.
Foveation can therefore identify either kind of high-cost region.
At 8B, $\beta$ becomes small, so the cost map is effectively precision-dominated and has little signal for identifying dispersed evidence patches.
At 32B, the situation is reversed: $\alpha$ becomes small while $\beta$ remains usable, so the cost map is coverage-dominated and is better suited to recovering evidence regions than fine-grained lexical distinctions.
This scale dependence is not a separate property of foveation.
It follows from the same probe-calibrated parameters that drive the global transport-efficiency rule.

\clearpage

\subsection{Empirical FOV across Scales} \label{sec:appx-fov-scale}

\begin{table*}[!htbp]
\centering
\footnotesize
\caption{Per-dataset foveation gain at three scales. Each scale shows two columns: the baseline VLM score and the FOV minus VLM gap in absolute points. Positive $\Delta_{\mathrm{Fov}}$ means foveation improves the visual path on that dataset. The All row reports paired averages over all 24 datasets at each scale after applying the trigger-zero convention in Sec.~\ref{sec:appx-fov-trigger}.}
\label{tab:appx-fov-at-scale}
\vspace{0.5em}
\begin{tabular*}{\textwidth}{@{\extracolsep{\fill}} l rr rr rr @{}}
\toprule[1.2pt]
& \multicolumn{2}{c}{4B} & \multicolumn{2}{c}{8B} & \multicolumn{2}{c}{32B} \\
\cmidrule(lr){2-3} \cmidrule(lr){4-5} \cmidrule(lr){6-7}
Dataset & VLM & $\Delta_{\mathrm{Fov}}$ & VLM & $\Delta_{\mathrm{Fov}}$ & VLM & $\Delta_{\mathrm{Fov}}$ \\
\midrule
\multicolumn{7}{@{}l}{\textit{Multi-choice RC}} \\
RACE-H & 25.1 & $+0.81$ & 23.9 & $+0.90$ & 26.7 & $-0.99$ \\
QuALITY & 60.5 & $+0.00$ & 67.2 & $+0.96$ & 75.3 & $-0.34$ \\
LogiQA & 38.9 & $+0.46$ & 41.3 & $+0.00$ & 42.9 & $+0.00$ \\
DREAM & 71.1 & $+0.05$ & 81.9 & $-0.34$ & 85.4 & $+0.44$ \\
\midrule
\multicolumn{7}{@{}l}{\textit{Binary}} \\
BoolQ & 78.3 & $+0.00$ & 84.2 & $-0.40$ & 87.2 & $+0.40$ \\
IMDB & 82.4 & $+0.00$ & 86.7 & $+0.00$ & 85.7 & $+0.00$ \\
\midrule
\multicolumn{7}{@{}l}{\textit{Multi-class}} \\
DBpedia & 93.8 & $+0.00$ & 96.5 & $+0.00$ & 97.3 & $+0.00$ \\
Yahoo & 64.1 & $+0.13$ & 66.7 & $-0.08$ & 67.7 & $-0.03$ \\
Yelp & 45.8 & $+0.00$ & 53.2 & $+0.00$ & 55.3 & $+0.00$ \\
\midrule
\multicolumn{7}{@{}l}{\textit{NLI/verif.}} \\
ContractNLI & 41.9 & $+0.68$ & 41.1 & $-0.77$ & 41.3 & $-1.54$ \\
SciFact & 71.4 & $+0.84$ & 70.4 & $-0.05$ & 74.9 & $+0.00$ \\
\midrule
\multicolumn{7}{@{}l}{\textit{Short-answer QA}} \\
HotpotQA & 35.4 & $+1.03$ & 37.7 & $-1.59$ & 29.3 & $+4.27$ \\
MuSiQue & 19.0 & $+1.13$ & 22.8 & $-0.99$ & 18.9 & $+2.53$ \\
2WikiMHQA & 38.2 & $+0.30$ & 44.0 & $-3.25$ & 35.7 & $+3.84$ \\
CoQA & 39.2 & $+0.05$ & 48.6 & $+0.11$ & 25.3 & $+0.68$ \\
DROP & 22.0 & $+0.00$ & 26.2 & $+0.13$ & 21.1 & $+0.23$ \\
MultiFieldQA & 38.8 & $+0.41$ & 44.1 & $+0.36$ & 41.7 & $+1.24$ \\
QASPER & 30.1 & $+2.93$ & 45.0 & $-0.61$ & 31.7 & $+0.70$ \\
ReCoRD & 26.5 & $+0.52$ & 29.0 & $-0.53$ & 46.8 & $+0.04$ \\
\midrule
\multicolumn{7}{@{}l}{\textit{Summarization}} \\
CNN/DM & 24.1 & $+0.00$ & 26.8 & $+0.00$ & 28.2 & $+0.00$ \\
XSum & 16.8 & $+0.00$ & 17.4 & $+0.00$ & 17.7 & $+0.00$ \\
BillSum & 40.9 & $+0.00$ & 41.2 & $+0.00$ & 42.3 & $+0.00$ \\
GovReport & 32.5 & $+0.00$ & 37.1 & $+0.00$ & 41.8 & $+0.00$ \\
QMSum & 28.0 & $+0.86$ & 29.0 & $-0.08$ & 30.1 & $+0.38$ \\
\midrule
All (paired $n$) & 44.4 & $\mathbf{+0.42}$ & 48.4 & $\mathbf{-0.26}$ & 47.9 & $\mathbf{+0.49}$ \\
\bottomrule[1.2pt]
\end{tabular*}
\vspace{0.6em}
\end{table*}

\vspace{0.6em}

\noindent\textbf{The purpose of this subsection:}
This subsection tests whether localized inverse transport improves visual path only when patch-level cost map contains the right signal for the task family.
The foveation choices are made by the label-free cost map described in Sec.~\ref{sec:fov}.
VLM and FOV scores reported here are used only for post-hoc evaluation of whether those choices recover useful information.

Tab.~\ref{tab:appx-fov-at-scale} reports, for each of the 24 benchmarks, the VLM score without foveation and the FOV minus VLM gap at 4B, 8B, and 32B.
A positive $\Delta_{\mathrm{Fov}}$ means that foveation improves the visual path, while a negative value means that the selected crops add token cost or noise without recovering enough useful information.
The aggregate paired gains are $\Delta_{\mathrm{Fov}}=+0.42$ at 4B, $-0.26$ at 8B, and $+0.49$ at 32B.
These values are paired over all 24 datasets at each scale and apply the trigger-zero reporting convention in Sec.~\ref{sec:appx-fov-trigger}.
The non-monotonic trajectory is consistent with the scale-dependent probe regimes in Tab.~\ref{tab:appx-params}.
Foveation is beneficial at 4B, where both precision and coverage terms are active.
It becomes negative at 8B, where the coverage weight $\beta$ is weak and the cost map has little signal for selecting dispersed evidence regions.
It becomes positive again at 32B, where the coverage signal becomes usable again and the largest gains concentrate on datasets whose evidence can be localized to a small number of useful regions.

The per-family pattern is more informative than the aggregate.
\emph{Short-answer QA} is the family with the clearest scale-dependent foveation behavior: its average foveation gain is positive at 4B, negative at 8B, and strongly positive at 32B.
This family contains multi-hop and span-style QA tasks where recoverable evidence often appears in localized regions of the rendered passage.
When the coverage term is weak at 8B, foveation can select regions that are locally dense but not necessarily evidence-bearing.
When the coverage signal becomes usable again at 32B, the same mechanism can identify evidence patches more effectively.
In contrast, \emph{Multi-choice RC} and \emph{NLI/verif.} are more sensitive to precision.
They become less favorable to foveation at 32B, where $\alpha$ is small and the cost map carries little precision signal.
Summarization shows only small foveation gains despite being coverage-heavy, because the relevant information is often diffuse across long documents rather than concentrated in a few crop-worthy regions.

\subsection{Discussion: Non-uniform Gains as Self-Consistency}
\label{sec:appx-fov-discussion}

\noindent\textbf{The purpose of this discussion:}
This discussion interprets the non-uniform foveation gains as a scale-aware limitation of the method rather than as an unrelated failure.
It connects the local behavior of $C_q$ back to the global transport-cost structure $C(x)$.

The fact that foveation does not produce a uniform gain at every scale is a limitation of a scale-blind deployment, but it is also consistent with the proposed framework.
The local refinement in Sec.~\ref{sec:fov} is, by construction, the patch-level instantiation of the same cost decomposition that drives the global decision in Sec.~\ref{sec:te}.
When a probe-calibrated parameter becomes weak at a particular scale, the cost map loses sensitivity to the corresponding failure mode, and foveation can select regions that do not recover enough information to offset their token cost.
The pattern in Tab.~\ref{tab:appx-fov-at-scale} therefore reads as follows.
When the local cost map retains the signal needed by a task family, foveation contributes.
When the relevant signal is weak, foveation contributes little or can hurt.
This supports the claim that $C(x)$ and $C_q$ are two linked instantiations of the same transport-cost structure rather than independent heuristics that only agree at 4B.

A practical consequence is that a scale-blind foveation deployment is not the right interface beyond the 4B setting.
A better deployment is to gate foveation by the same scale-aware probes that gate routing.
Foveation should be enabled only when the relevant cost dimension remains reliable at the target scale.
We leave this scale-conditioned foveation gate to future work.
The parameter table in Tab.~\ref{tab:appx-params} provides the information needed to implement it.

\clearpage

\subsection{Sample-Level Visualization}
\label{sec:appx-fov-vis}

\noindent\textbf{The purpose of this subsection:}
The aggregate gains in Tab.~\ref{tab:appx-fov-at-scale} are summary statistics.
This subsection complements them with qualitative sample-level evidence that the cost map localizes recoverable evidence and that the resulting crops can correct VLM-baseline errors.
The examples are selected from datasets where foveation produces a sizeable positive macro gain, so they should be read as illustrative cases rather than as an unbiased estimate of average performance.

For visualization, we show the rendered page, a normalized patch-cost overlay, and the crop region selected by foveation.
The displayed heatmap emphasizes the coverage component $\beta L_q(1-\mathrm{TRR}_q)$ because it is directly interpretable from the rendered text.
The deployed foveation rule, however, uses the full patch-level cost map $C_q$ defined in Sec.~\ref{sec:fov}.
Thus, the figures should be interpreted as visual explanations of where the selected evidence lies, not as a separate decision rule.

\textbf{\textit{Interpreting the hot patches.}}
The cost-map overlay marks where question keywords concentrate after redundancy is discounted (BM25-style query-patch locality $L_q$ damped by patch redundancy $\mathrm{TRR}_q$), not the literal pixel location of the gold answer.
The two regions coincide on extractive QA, where the gold answer string is co-located with question keywords on the visible page (the MultiFieldQA panels and the extractive QASPER s1 panel), but can diverge in two structural cases.
First, for inference-based questions whose gold answer paraphrases or summarizes the supporting text (QASPER s2 and s3, both yes/no), hot patches mark the topic-relevant evidence the model needs to reason from, not the answer string itself.
Second, for multi-hop questions whose disambiguating evidence sits on a later rendered page than the first page shown here (HotpotQA s2 and s3), hot patches highlight one branch of the multi-hop chain that lies on the visible page, while the gold-answer string may not appear on this page at all.
In all cases, both the cost map and the selected crop are computed in the rendering coordinate system at inference time and are never selected with reference to the gold answer.

\paragraph{Visualized samples.}
We visualize three QASPER samples, three HotpotQA samples, and three MultiFieldQA samples.
The QASPER and HotpotQA examples show cases where the baseline VLM scores at or near $0$ and foveation lifts the prediction to $1$.
The MultiFieldQA examples show partial F1 improvements on long-form QA.
Each sample yields one figure with three stacked panels: rendered page, patch-cost overlay, and selected crop box.
The corresponding question, gold answer, and predicted strings are consolidated in Tab.~\ref{tab:appx-fov-qasper-details}, Tab.~\ref{tab:appx-fov-hotpotqa-details}, and Tab.~\ref{tab:appx-fov-mfqa-details}.

\clearpage

\vspace{1.0em}

\begin{table*}[!htbp]
\centering
\footnotesize
\caption{QASPER foveation examples corresponding to Figs.~\ref{fig:appx-fov-qasper-s1}--\ref{fig:appx-fov-qasper-s3}. The QASPER evaluation pipeline records F1 only, so predicted text is omitted. Long strings are truncated.}
\label{tab:appx-fov-qasper-details}
\vspace{0.4em}
\renewcommand{\arraystretch}{1.15}
\begin{tabular*}{\textwidth}{@{\extracolsep{\fill}} l p{0.48\linewidth} p{0.16\linewidth} p{0.20\linewidth} @{}}
\toprule[1.2pt]
Fig. & Question & Gold & F1 change \\
\midrule
\ref{fig:appx-fov-qasper-s1}
& Which other tasks are evaluated?
& product category classification and review headline generation
& $0.26 \rightarrow 1.00$ \\

\ref{fig:appx-fov-qasper-s2}
& Did they use crowdsourcing for the annotations?
& No
& $0.00 \rightarrow 1.00$ \\

\ref{fig:appx-fov-qasper-s3}
& Are the recommendations specific to a region?
& No
& $0.00 \rightarrow 1.00$ \\
\bottomrule[1.2pt]
\end{tabular*}
\vspace{0.4em}
\end{table*}

\vspace{1.0em}

\begin{table*}[!htbp]
\centering
\footnotesize
\caption{HotpotQA foveation examples corresponding to Figs.~\ref{fig:appx-fov-hotpotqa-s1}--\ref{fig:appx-fov-hotpotqa-s3}. The baseline VLM gives incorrect short answers, while VLM+Fov recovers the gold entity. Long strings are truncated.}
\label{tab:appx-fov-hotpotqa-details}
\vspace{0.4em}
\renewcommand{\arraystretch}{1.18}
\begin{tabular*}{\textwidth}{@{\extracolsep{\fill}} l p{0.30\linewidth} p{0.13\linewidth} p{0.22\linewidth} p{0.22\linewidth} @{}}
\toprule[1.2pt]
Fig. & Question & Gold & VLM baseline & VLM+Fov \\
\midrule
\ref{fig:appx-fov-hotpotqa-s1}
& What pop band features an Argentine actress that appeared in the television series ``Chiquititas''?
& Erreway
& F1 $=0.00$: ``The pop band is `Tinys Angels'.''
& F1 $=1.00$: ``Erreway'' \\

\ref{fig:appx-fov-hotpotqa-s2}
& Which species is native to Asia, Corokia or Rhodotypos?
& Rhodotypos
& F1 $=0.00$: ``Corokia''
& F1 $=1.00$: ``Rhodotypos'' \\

\ref{fig:appx-fov-hotpotqa-s3}
& Which was formed first, Noori or Test Icicles?
& Noori
& F1 $=0.00$: ``Test Icicles''
& F1 $=1.00$: ``Noori'' \\
\bottomrule[1.2pt]
\end{tabular*}
\vspace{0.4em}
\end{table*}

\vspace{1.0em}

\begin{table*}[!htbp]
\centering
\footnotesize
\caption{MultiFieldQA foveation examples corresponding to Figs.~\ref{fig:appx-fov-mfqa-s1}--\ref{fig:appx-fov-mfqa-s3}. VLM+Fov improves answer specificity by re-examining high-cost evidence regions. Long strings are truncated.}
\label{tab:appx-fov-mfqa-details}
\vspace{0.4em}
\renewcommand{\arraystretch}{1.18}
\begin{tabular*}{\textwidth}{@{\extracolsep{\fill}} l p{0.28\linewidth} p{0.20\linewidth} p{0.24\linewidth} p{0.24\linewidth} @{}}
\toprule[1.2pt]
Fig. & Question & Gold & VLM baseline & VLM+Fov \\
\midrule
\ref{fig:appx-fov-mfqa-s1}
& Which air unit did Goodwin command during the initial landings of Marines on Saipan?
& VC-10 Squadron
& F1 $=0.22$: ``The air unit Goodwin commanded during the initial landings \dots was VC-10 Squadron.''
& F1 $=1.00$: ``VC-10 Squadron'' \\

\ref{fig:appx-fov-mfqa-s2}
& What was the conclusion of the study?
& The conclusion was that fruit consumption may provide a protective effect for mercury exposure \dots
& F1 $=0.26$: ``The document contains multiple studies on mercury exposure from fish consumption \dots''
& F1 $=0.58$: ``The conclusion of the study was that fruit consumption may provide a protective effect \dots'' \\

\ref{fig:appx-fov-mfqa-s3}
& According to the text, what is Toby Schindelbeck's observation about the police?
& Toby Schindelbeck's observation is that the police say they are not paid enough to enforce the laws \dots
& F1 $=0.48$: ``Toby Schindelbeck observes that the police are not paid enough to enforce the laws \dots''
& F1 $=0.52$: ``Toby Schindelbeck observes that the police are not paid enough to enforce the laws \dots'' \\
\bottomrule[1.2pt]
\end{tabular*}
\vspace{0.4em}
\end{table*}

\clearpage

\begin{figure}[!htbp]
\centering
\includegraphics[width=\linewidth]{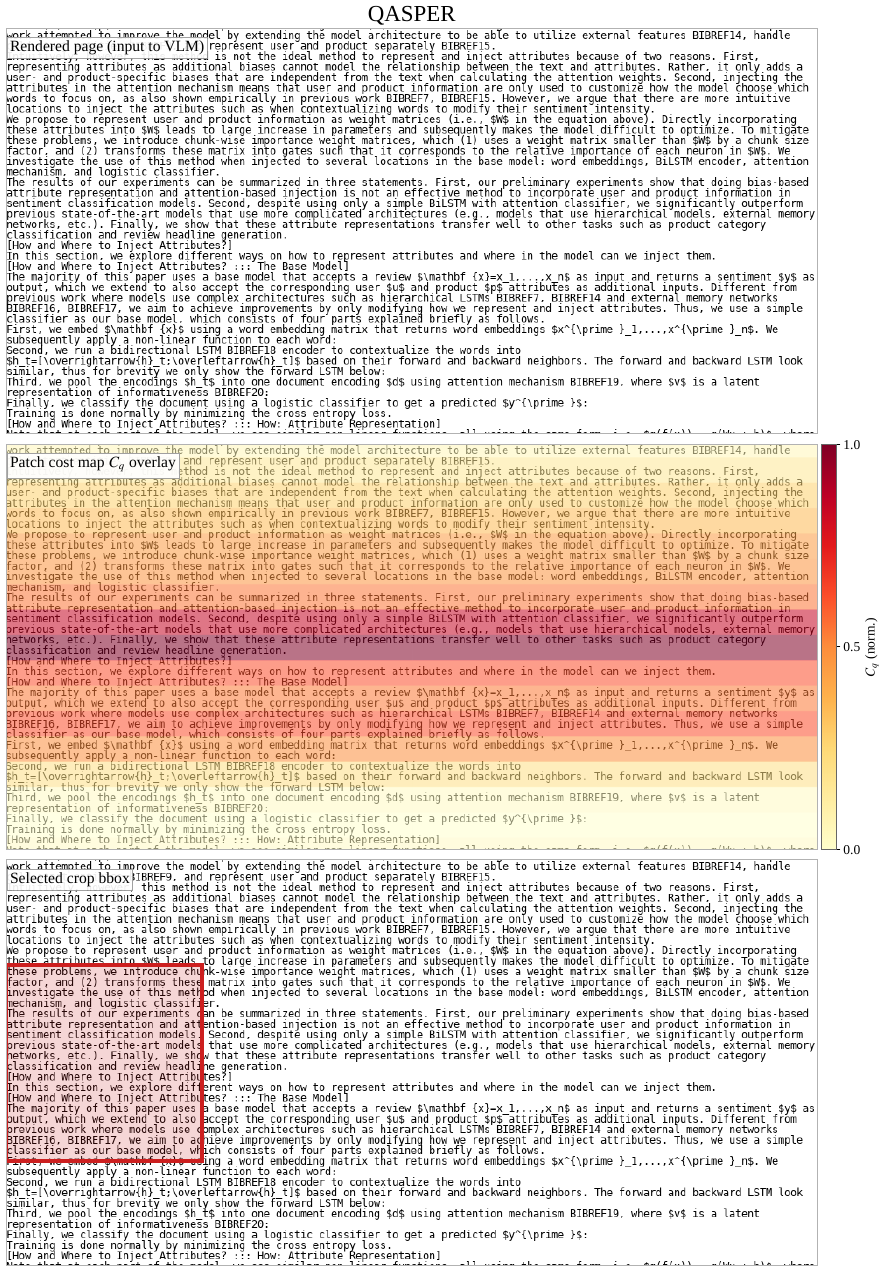}
\caption{QASPER example 1. The three panels show the rendered page, the normalized patch-cost overlay, and the selected crop box used by foveation. Hot patches mark question-keyword concentration after redundancy is discounted, not literal answer locations.}
\label{fig:appx-fov-qasper-s1}
\end{figure}

\begin{figure}[!htbp]
\centering
\includegraphics[width=\linewidth]{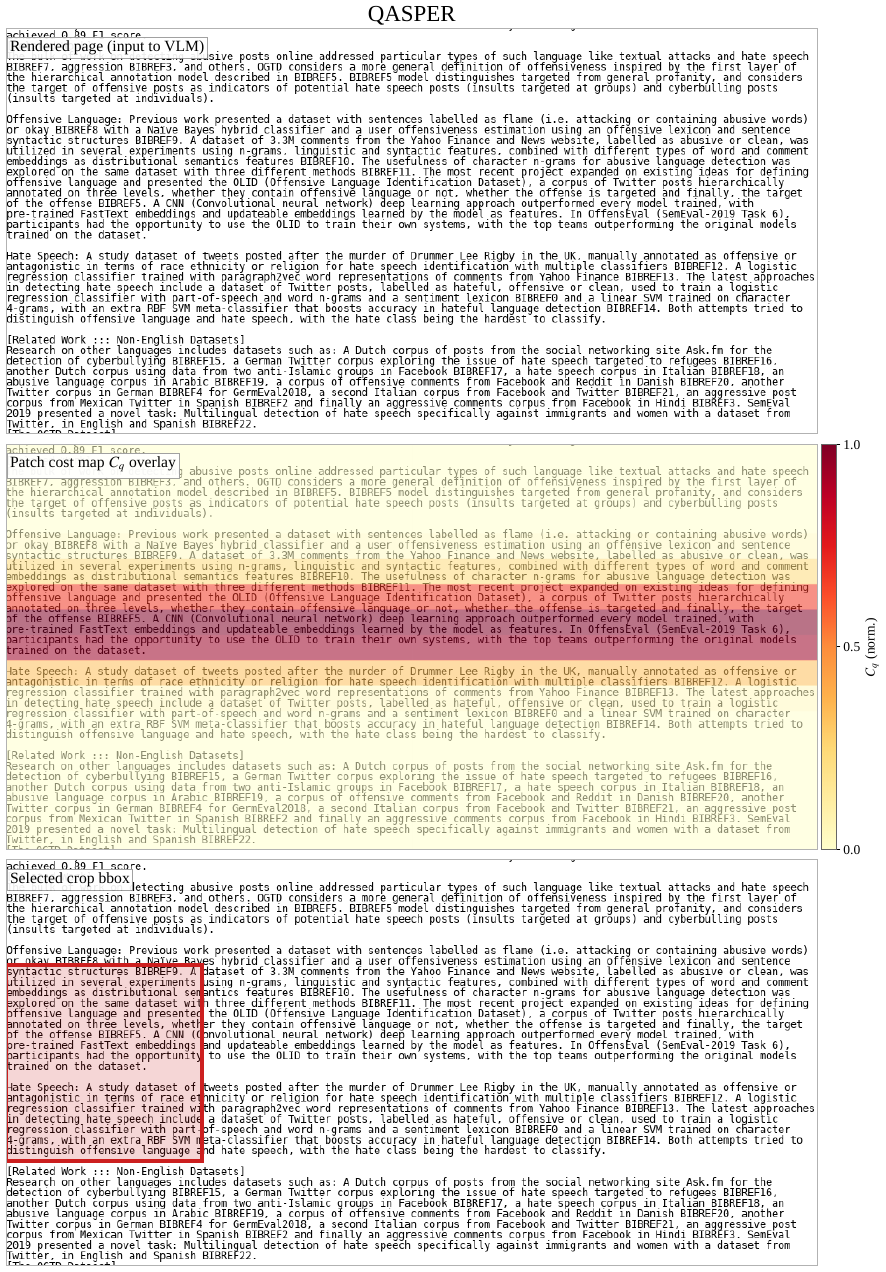}
\caption{QASPER example 2. The three panels show the rendered page, the normalized patch-cost overlay, and the selected crop box used by foveation. Hot patches mark question-keyword concentration after redundancy is discounted, not literal answer locations.}
\label{fig:appx-fov-qasper-s2}
\end{figure}

\begin{figure}[!htbp]
\centering
\includegraphics[width=\linewidth]{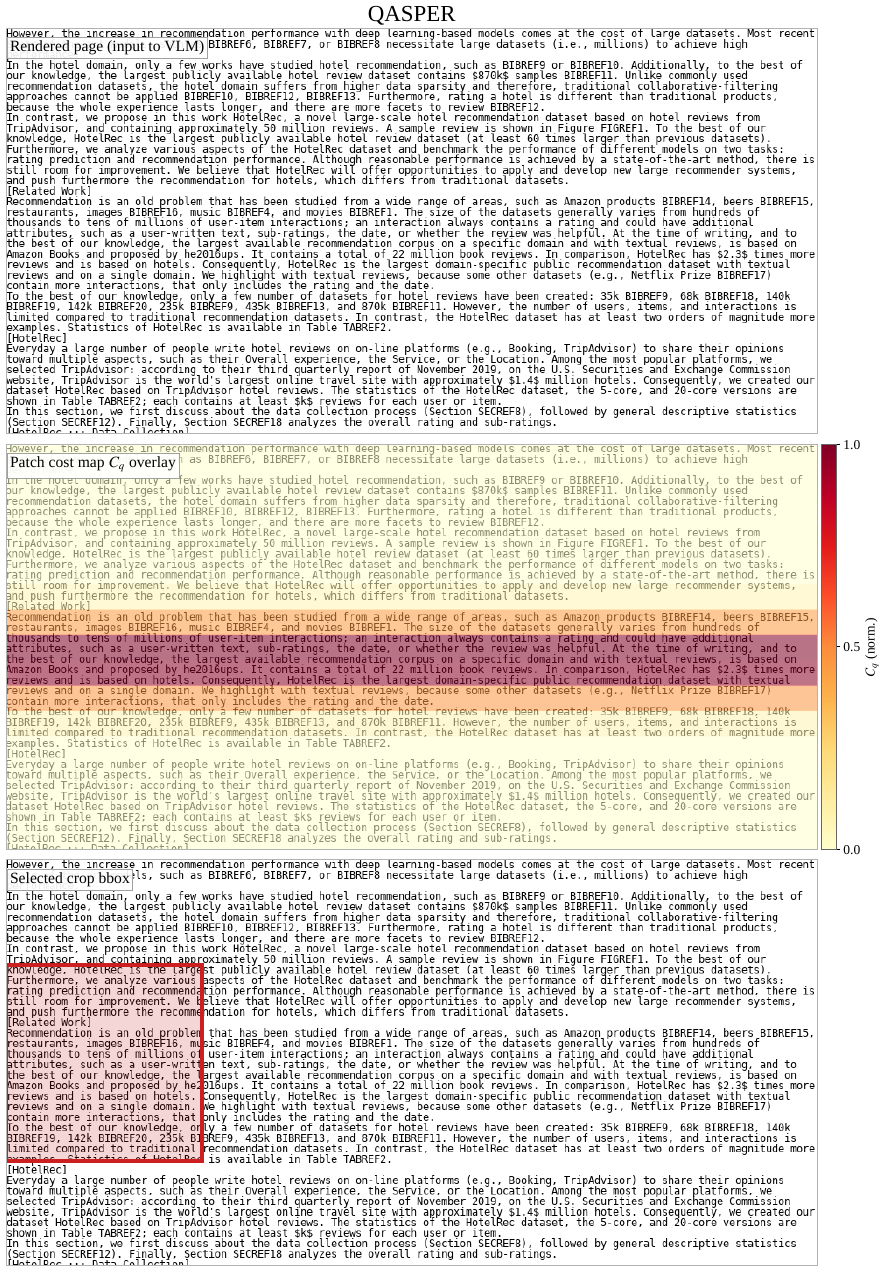}
\caption{QASPER example 3. The three panels show the rendered page, the normalized patch-cost overlay, and the selected crop box used by foveation. Hot patches mark question-keyword concentration after redundancy is discounted, not literal answer locations.}
\label{fig:appx-fov-qasper-s3}
\end{figure}

\begin{figure}[!htbp]
\centering
\includegraphics[width=\linewidth]{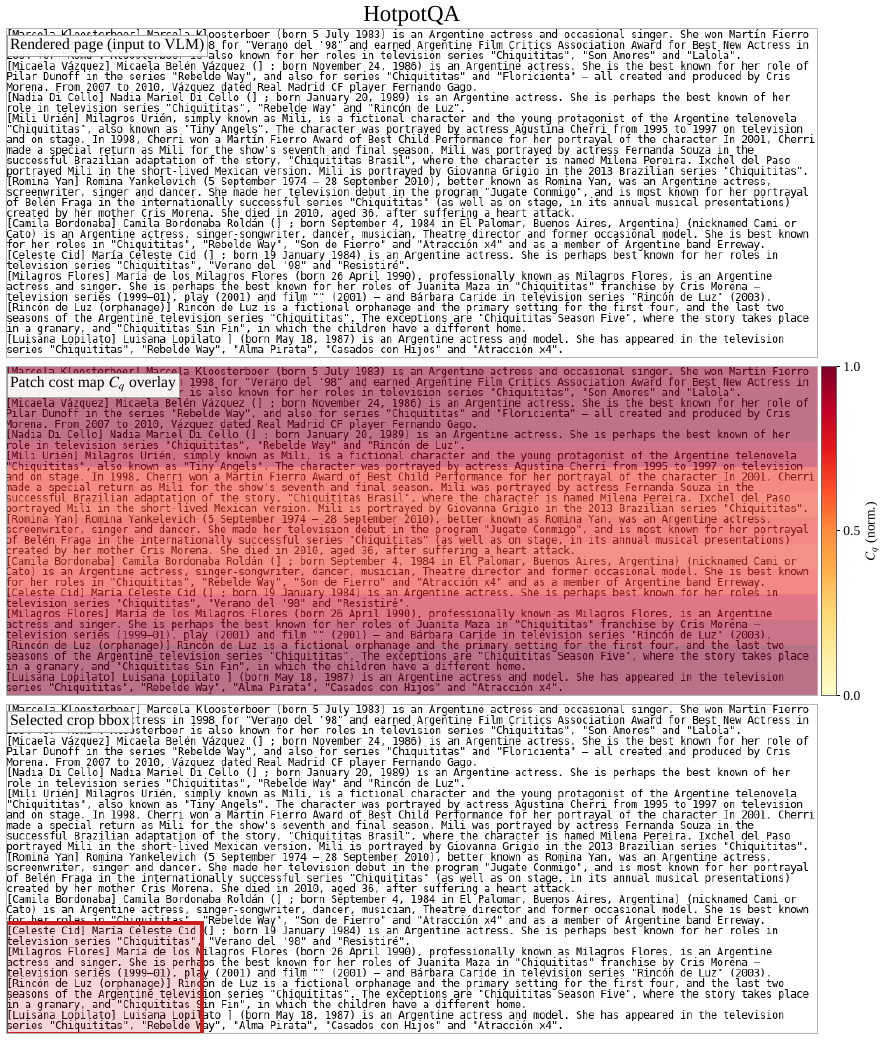}
\caption{HotpotQA example 1. The three panels show the rendered page, the normalized patch-cost overlay, and the selected crop box used by foveation. Hot patches mark question-keyword concentration after redundancy is discounted, not literal answer locations.}
\label{fig:appx-fov-hotpotqa-s1}
\end{figure}

\begin{figure}[!htbp]
\centering
\includegraphics[width=\linewidth]{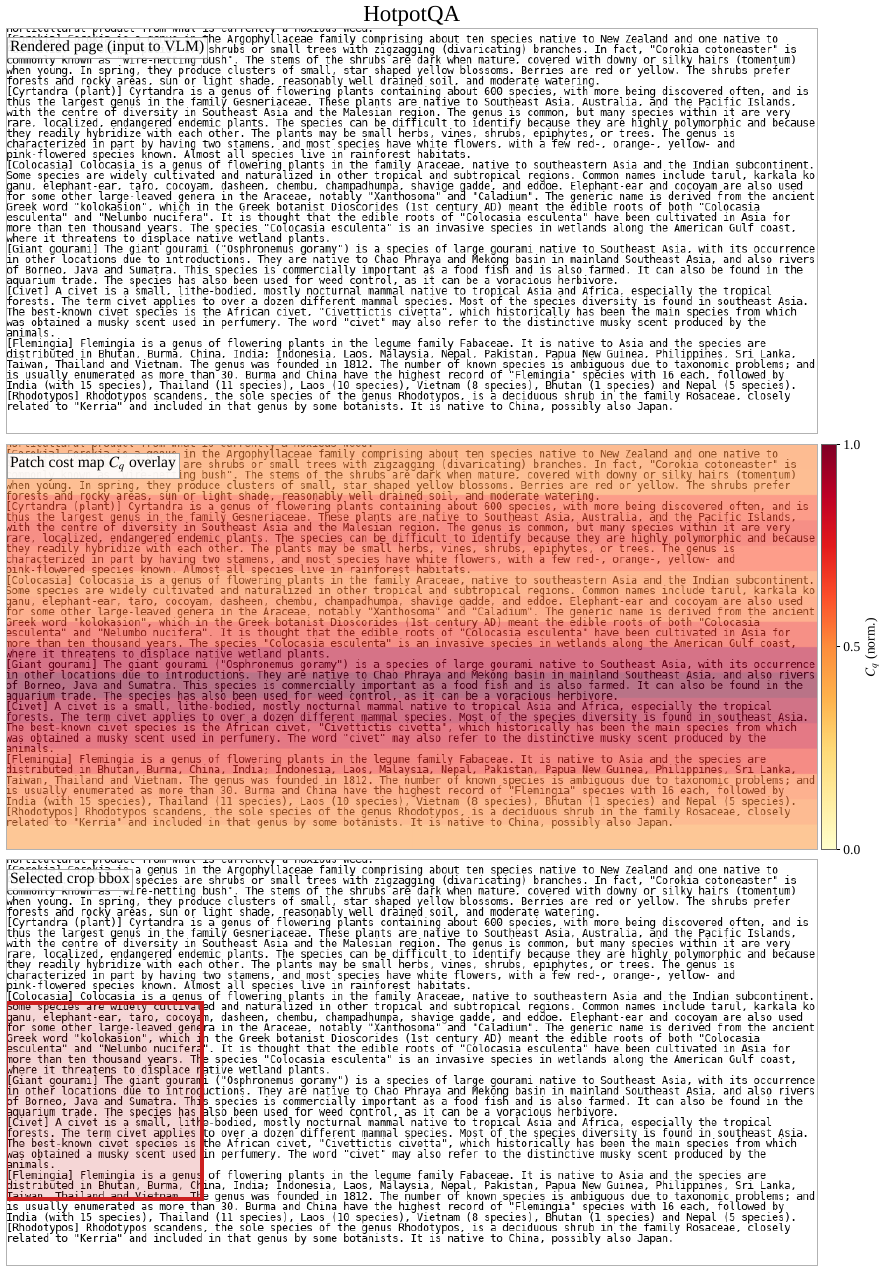}
\caption{HotpotQA example 2. The three panels show the rendered page, the normalized patch-cost overlay, and the selected crop box used by foveation. Hot patches mark question-keyword concentration after redundancy is discounted, not literal answer locations.}
\label{fig:appx-fov-hotpotqa-s2}
\end{figure}

\begin{figure}[!htbp]
\centering
\includegraphics[width=\linewidth]{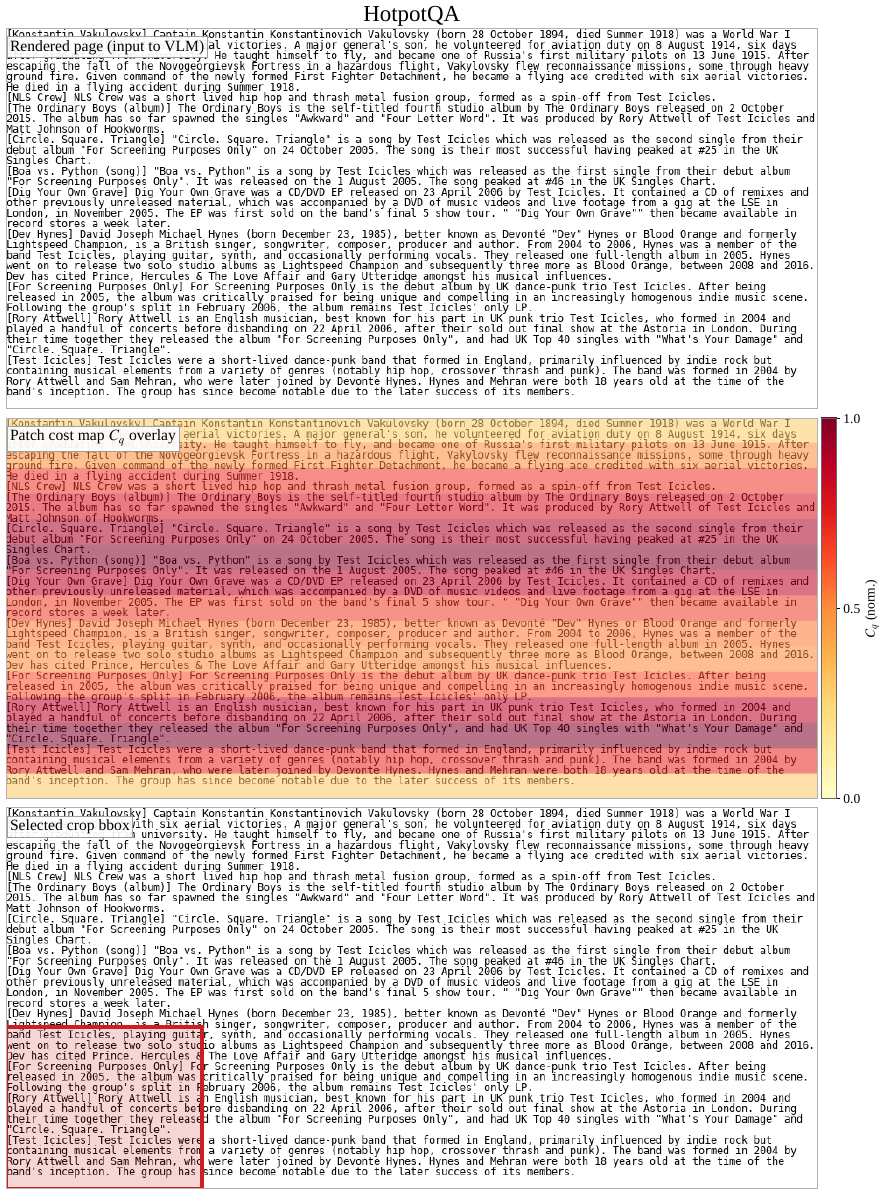}
\caption{HotpotQA example 3. The three panels show the rendered page, the normalized patch-cost overlay, and the selected crop box used by foveation. Hot patches mark question-keyword concentration after redundancy is discounted, not literal answer locations.}
\label{fig:appx-fov-hotpotqa-s3}
\end{figure}

\begin{figure}[!htbp]
\centering
\includegraphics[width=\linewidth]{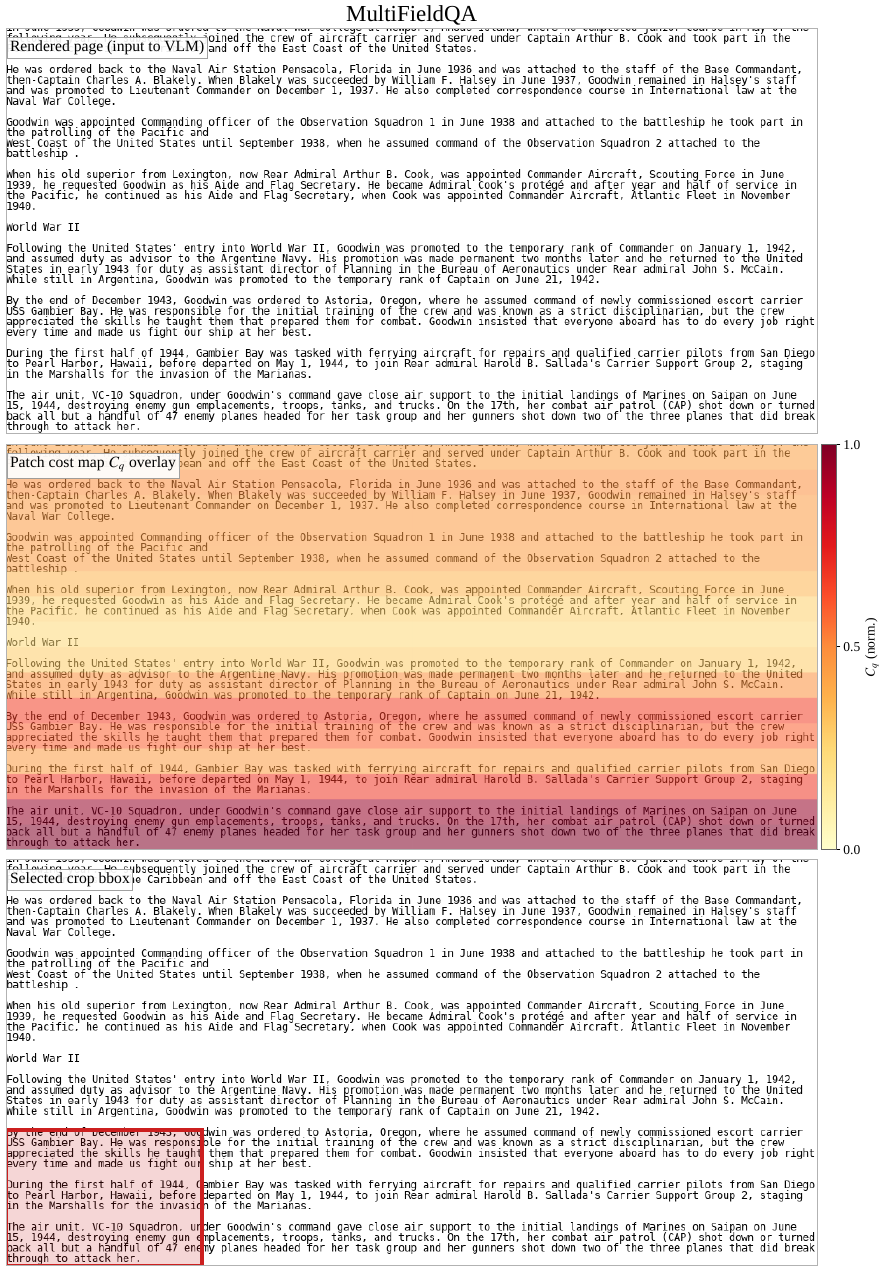}
\caption{MultiFieldQA example 1. The three panels show the rendered page, the normalized patch-cost overlay, and the selected crop box used by foveation. Hot patches mark question-keyword concentration after redundancy is discounted, not literal answer locations.}
\label{fig:appx-fov-mfqa-s1}
\end{figure}

\begin{figure}[!htbp]
\centering
\includegraphics[width=\linewidth]{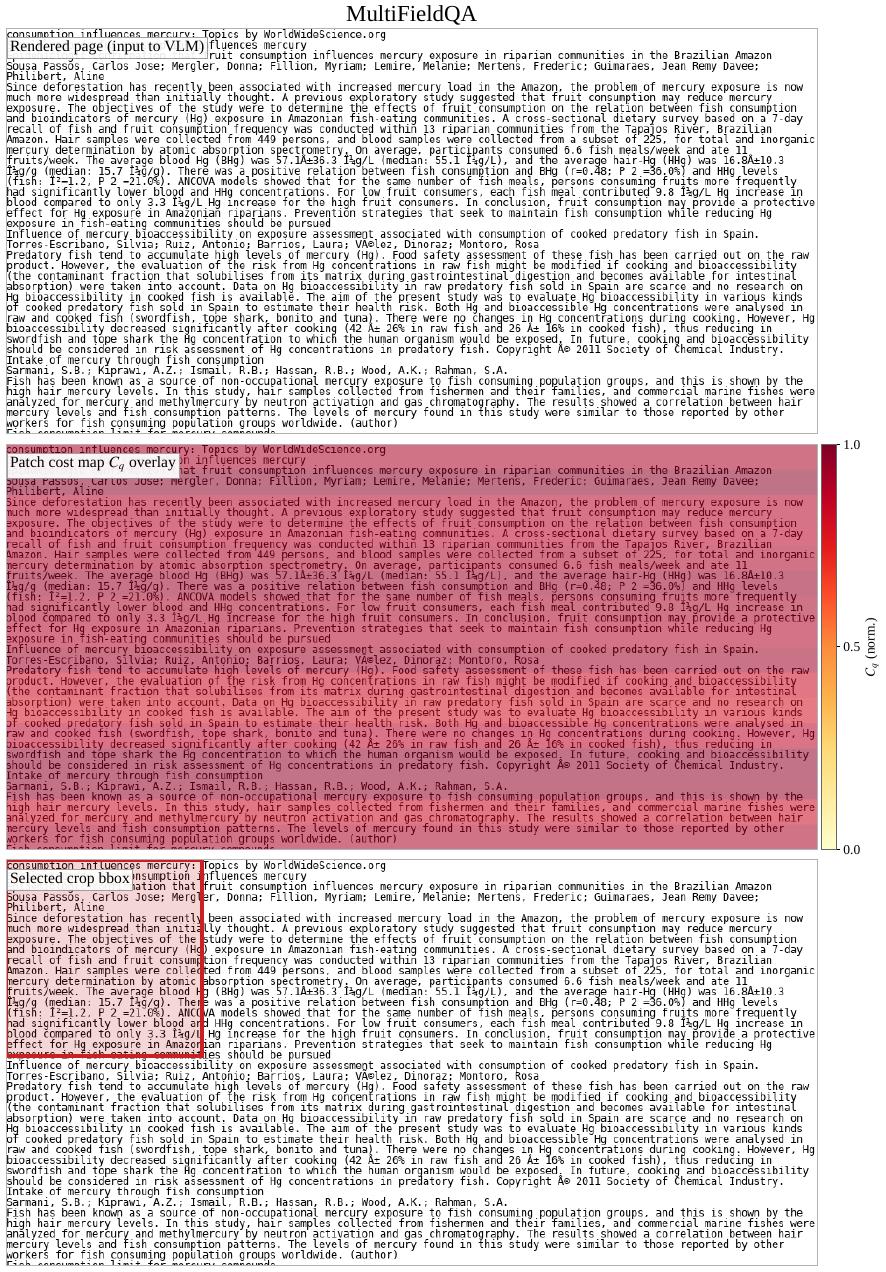}
\caption{MultiFieldQA example 2. The three panels show the rendered page, the normalized patch-cost overlay, and the selected crop box used by foveation. Hot patches mark question-keyword concentration after redundancy is discounted, not literal answer locations.}
\label{fig:appx-fov-mfqa-s2}
\end{figure}

\begin{figure}[!htbp]
\centering
\includegraphics[width=\linewidth]{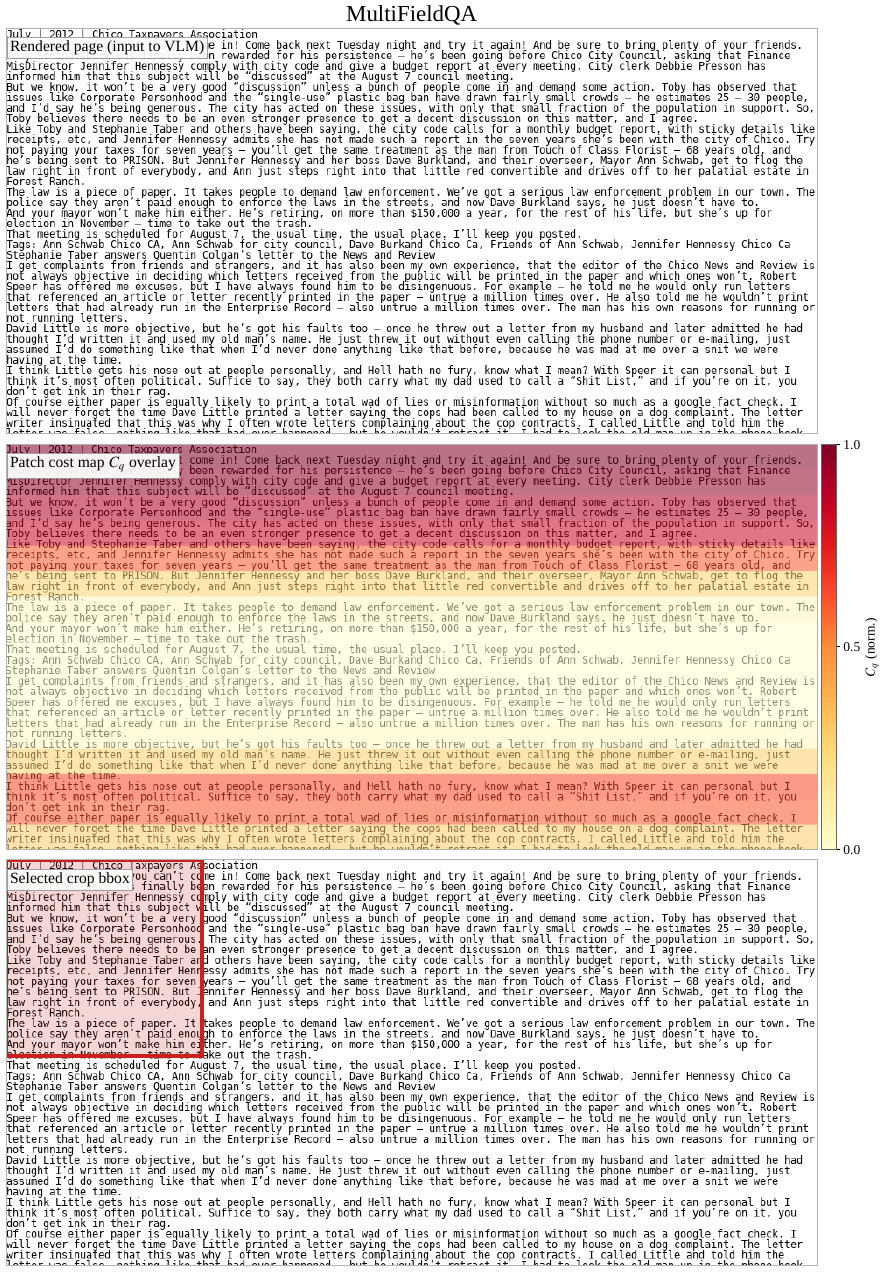}
\caption{MultiFieldQA example 3. The three panels show the rendered page, the normalized patch-cost overlay, and the selected crop box used by foveation. Hot patches mark question-keyword concentration after redundancy is discounted, not literal answer locations.}
\label{fig:appx-fov-mfqa-s3}
\end{figure}

\clearpage

\subsection{Foveation Trigger Statistics} \label{sec:appx-fov-trigger}

\begin{figure}[!t]
\centering
\includegraphics[width=\linewidth]{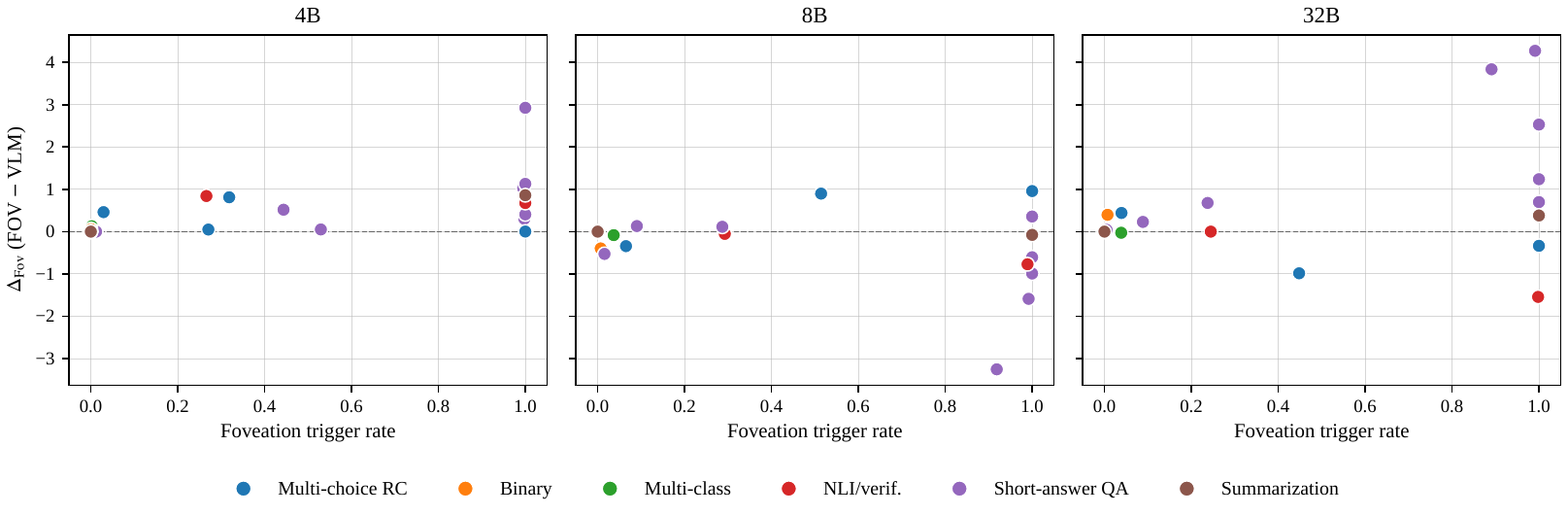}
\caption{Per-dataset relationship between the foveation trigger rate and the foveation gain $\Delta_{\mathrm{Fov}}$ at each backbone scale. Each point is one benchmark, colored by task family. Datasets with near-zero trigger rate cluster around $\Delta_{\mathrm{Fov}}=0$ at every scale. Datasets with high trigger rate carry most of the scale-dependent positive and negative $\Delta_{\mathrm{Fov}}$ changes, including the largest 32B short-answer-QA gains and the largest 8B short-answer-QA losses.}
\label{fig:appx-fov-trigger-scatter}
\end{figure}

\noindent\textbf{The purpose of this subsection:}
The foveation gain $\Delta_{\mathrm{Fov}}$ in Tab.~\ref{tab:appx-fov-at-scale} is a per-dataset macro statistic, but the underlying decision is made per sample.
This subsection documents the per-sample trigger that gates whether an input is actually foveated.
It also explains why datasets with near-zero trigger rate have $\Delta_{\mathrm{Fov}}\approx 0$ by construction, while high-trigger datasets carry most of the positive and negative scale-dependent changes.

\paragraph{Trigger definition.}
We say the foveation trigger fires on a sample $x$ when the transport-efficiency inequality of Eq.~\ref{eq:fov-trigger} is satisfied.
Equivalently, the relative cost recovered by the selected patches must exceed the relative token overhead:
\[
\frac{\sum_{k}\Delta C_k}{\mathrm{ISR}(x)}
>
\frac{n_c}{n_v},
\qquad
\mathrm{ISR}(x)=1+\gamma-C(x),
\]
where the sum runs over the patches added by the greedy region-selection procedure of Sec.~\ref{sec:fov}, and $n_c$, $n_v$ are the additional and base visual-token counts.
In practice, the trigger is evaluated only when the cost map $\{C_q\}$ contains a localized hot region.
We use the criterion $\max_q C_q/\overline{C_q}>2.5$.
If the cost map is approximately uniform, no region can recover enough cost to justify extra visual tokens, so the sample is forwarded through the base VLM path.
All quantities in the trigger are computed from probe-calibrated, downstream-label-free features.

\paragraph{Convention for trigger-zero benchmarks.}
If the trigger never fires on a benchmark, every sample is forwarded along the unfoveated VLM path by construction.
In that case, the FOV pipeline and the VLM pipeline are identical for reporting purposes, and $\Delta_{\mathrm{Fov}}=0$.
We therefore report $\Delta_{\mathrm{Fov}}=+0.00$ on every scale-benchmark cell whose trigger rate is $0.0\%$ in Tab.~\ref{tab:appx-fov-trigger}.
This avoids interpreting small numerical differences from independent evaluation runs as foveation effects.
The same convention is applied to Tab.~\ref{tab:appx-fov-at-scale}, Fig.~\ref{fig:appx-fov-trigger-scatter}, and the aggregate statistics in this section.

\begin{table*}[!htbp]
\centering
\footnotesize
\caption{Per-dataset foveation trigger rate at three scales together with the resulting per-dataset $\Delta_{\mathrm{Fov}}$ from Tab.~\ref{tab:appx-fov-at-scale}. The trigger rate is the fraction of samples for which the foveation trigger of Eq.~\ref{eq:fov-trigger} fires. The macro-average row averages over all 24 datasets at each scale.}
\label{tab:appx-fov-trigger}
\vspace{0.5em}
\begin{tabular*}{\textwidth}{@{\extracolsep{\fill}} l rrr rrr @{}}
\toprule[1.2pt]
& \multicolumn{3}{c}{Trigger rate} & \multicolumn{3}{c}{$\Delta_{\mathrm{Fov}}$} \\
\cmidrule(lr){2-4} \cmidrule(lr){5-7}
Dataset & 4B & 8B & 32B & 4B & 8B & 32B \\
\midrule
\multicolumn{7}{@{}l}{\textit{Multi-choice RC}} \\
RACE-H & 31.9\% & 51.4\% & 44.8\% & $+0.81$ & $+0.90$ & $-0.99$ \\
QuALITY & 100.0\% & 100.0\% & 100.0\% & $+0.00$ & $+0.96$ & $-0.34$ \\
LogiQA & 2.9\% & 0.0\% & 0.0\% & $+0.46$ & $+0.00$ & $+0.00$ \\
DREAM & 27.1\% & 6.5\% & 3.9\% & $+0.05$ & $-0.34$ & $+0.44$ \\
\midrule
\multicolumn{7}{@{}l}{\textit{Binary}} \\
BoolQ & 0.0\% & 0.7\% & 0.7\% & $+0.00$ & $-0.40$ & $+0.40$ \\
IMDB & 0.0\% & 0.0\% & 0.0\% & $+0.00$ & $+0.00$ & $+0.00$ \\
\midrule
\multicolumn{7}{@{}l}{\textit{Multi-class}} \\
Yahoo & 0.2\% & 3.7\% & 3.9\% & $+0.13$ & $-0.08$ & $-0.03$ \\
DBpedia & 0.0\% & 0.0\% & 0.0\% & $+0.00$ & $+0.00$ & $+0.00$ \\
Yelp & 0.0\% & 0.0\% & 0.0\% & $+0.00$ & $+0.00$ & $+0.00$ \\
\midrule
\multicolumn{7}{@{}l}{\textit{NLI/verif.}} \\
SciFact & 26.6\% & 29.3\% & 24.5\% & $+0.84$ & $-0.05$ & $+0.00$ \\
ContractNLI & 100.0\% & 98.9\% & 99.8\% & $+0.68$ & $-0.77$ & $-1.54$ \\
\midrule
\multicolumn{7}{@{}l}{\textit{Short-answer QA}} \\
2WikiMHQA & 99.8\% & 91.9\% & 89.1\% & $+0.30$ & $-3.25$ & $+3.84$ \\
HotpotQA & 99.5\% & 99.2\% & 99.1\% & $+1.03$ & $-1.59$ & $+4.27$ \\
MuSiQue & 100.0\% & 100.0\% & 100.0\% & $+1.13$ & $-0.99$ & $+2.53$ \\
DROP & 1.2\% & 9.0\% & 8.8\% & $+0.00$ & $+0.13$ & $+0.23$ \\
QASPER & 100.0\% & 100.0\% & 100.0\% & $+2.93$ & $-0.61$ & $+0.70$ \\
CoQA & 52.9\% & 28.7\% & 23.7\% & $+0.05$ & $+0.11$ & $+0.68$ \\
MultiFieldQA & 100.0\% & 100.0\% & 100.0\% & $+0.41$ & $+0.36$ & $+1.24$ \\
ReCoRD & 44.4\% & 1.6\% & 0.5\% & $+0.52$ & $-0.53$ & $+0.04$ \\
\midrule
\multicolumn{7}{@{}l}{\textit{Summarization}} \\
BillSum & 0.0\% & 0.0\% & 0.0\% & $+0.00$ & $+0.00$ & $+0.00$ \\
XSum & 0.0\% & 0.0\% & 0.0\% & $+0.00$ & $+0.00$ & $+0.00$ \\
QMSum & 100.0\% & 100.0\% & 100.0\% & $+0.86$ & $-0.08$ & $+0.38$ \\
GovReport & 0.0\% & 0.0\% & 0.0\% & $+0.00$ & $+0.00$ & $+0.00$ \\
CNN/DM & 0.0\% & 0.0\% & 0.0\% & $+0.00$ & $+0.00$ & $+0.00$ \\
\midrule
Macro avg & 41.1\% & 38.4\% & 37.5\% & $+0.42$ & $-0.26$ & $+0.49$ \\
\bottomrule[1.2pt]
\end{tabular*}
\vspace{0.6em}
\end{table*}

Tab.~\ref{tab:appx-fov-trigger} reports the per-dataset trigger rate at all three scales, alongside the per-dataset $\Delta_{\mathrm{Fov}}$ that it produces.
The trigger rate varies strongly across datasets.
It is $0\%$ on Binary tasks, most Multi-class tasks, and most short-summary tasks, but approaches $100\%$ on QASPER, MuSiQue, MultiFieldQA, ContractNLI, QMSum, QuALITY, HotpotQA, and 2WikiMultiHopQA.
The bimodal distribution is meaningful.
When the cost map has no hot region, the trigger suppresses foveation.
When the cost map identifies one or two coverage-cost peaks, which is typical of long-context QA and meeting summarization, nearly every sample triggers.

\paragraph{Foveation abstains on long-document summarization.}
A specific consequence of the trigger design is that foveation does \emph{not} attempt to repair long-document summarization.
On BillSum, XSum, GovReport, and CNN/DM the trigger rate is exactly $0\%$ at all three scales, so the FOV pipeline reproduces the VLM pipeline by construction and $\Delta_{\mathrm{Fov}}=+0.00$.
This behavior is consistent with the theory rather than an oversight.
Summarization is the canonical \emph{inter}-dominated regime: the relevant evidence is distributed across many rendered regions, so no single patch carries enough recoverable cost to satisfy the trigger inequality of Eq.~\ref{eq:fov-trigger}.
Adding a small number of higher-resolution patches cannot repair fragmentation that is spread across the entire document.
We therefore interpret the zero-trigger summarization cells as a deliberate abstention, not as a degenerate failure: the rule chooses not to spend extra visual tokens when the cost map provides no localizable target.
The lone summarization exception, QMSum, has a $100\%$ trigger rate because its meeting-style transcripts concentrate the answer-relevant span around a small number of dialogue turns.

The scatter in Fig.~\ref{fig:appx-fov-trigger-scatter} shows that $\Delta_{\mathrm{Fov}}$ is bounded by the trigger rate.
A dataset with trigger rate close to zero cannot have a large $|\Delta_{\mathrm{Fov}}|$ because the FOV pipeline degenerates to the VLM pipeline on most of its samples.
This is visible in the scatter cloud near the origin at every scale.
The non-trivial $\Delta_{\mathrm{Fov}}$ values, both the positive ones at 4B and 32B and the negative ones at 8B, all originate from datasets in the high-trigger-rate region.
The trigger therefore acts as a downstream-label-free filter that concentrates foveation on the inputs whose cost map signals it has something to recover.
The empirical $\Delta_{\mathrm{Fov}}$ behavior of Tab.~\ref{tab:appx-fov-at-scale} is consistent with this structure: positive at 4B where the cost map has both terms, negative at 8B where the coverage signal is weak in the relevant dataset population, and positive again at 32B where the coverage signal becomes usable again.

\clearpage

%% file: Appendix/appx_rendering.tex
\section{Rendering Protocol and Font Size Sensitivity}
\label{sec:appx-rendering}

\begin{table}[!htbp]
\centering
\footnotesize
\renewcommand\arraystretch{1.15}
\caption{Font size sensitivity of the base visual path at three scales. Scores are macro-averages over the 24-benchmark suite without foveation. The default setting is $12$\,pt.}
\label{tab:appx-rendering-scale}
\vspace{0.5em}
\begin{tabular*}{0.7\textwidth}{@{\extracolsep{\fill}} l c c c r @{}}
\toprule[1.2pt]
& \multicolumn{3}{c}{All score} & \\
\cmidrule(lr){2-4}
Font size & 4B & 8B & 32B & Tokens \\
\midrule
font10            & 42.4 & 46.4 & 45.8 & 1{,}185 \\
font12 (Default)  & 44.4 & 48.4 & 47.9 & 1{,}486 \\
font14            & 45.3 & 49.7 & 49.8 & 1{,}716 \\
\bottomrule[1.2pt]
\end{tabular*}
\vspace{0.6em}
\end{table}

\vspace{0.6em}

\noindent\textbf{The purpose of this section:}
This section documents the deterministic rendering protocol used to construct visual-path inputs and tests whether the visual-path behavior follows the expected font size tradeoff.
The goal is not to tune rendering per dataset, but to verify that a single global rendering knob changes score and token count in a predictable way.

\noindent\textbf{Rendering setup.}
For all visual-path runs, we render the input text as a single-column black-on-white RGB document image.
The default setting uses Roboto-Regular at $12$\,pt, line spacing $1.0$, no extra inter-line gap, and antialiased glyph rasterization.
The page width and height are sized adaptively to fit the rendered text, then snapped to the encoder patch grid and capped at $928$ pixels per dimension.
Longer inputs are paginated, with interior pages reaching the full $928\times 928$ pixel bound.
The rendered image is passed to the Qwen3-VL visual encoder, which uses $16\times 16$ raw patches followed by a $2\times 2$ spatial merge inside the encoder.

Among these rendering choices, font size directly controls the average number of characters that fall into one visual patch.
It therefore changes the strength of within-patch aggregation $T_{\mathrm{intra}}$ in the push-forward map.
Within our transport view, smaller fonts pack more characters into each patch and should increase lexical smoothing, while larger fonts reduce smoothing at the cost of more visual tokens.
We test this prediction by sweeping font size in $\{10,12,14\}$ while keeping the font family, layout, color scheme, page sizing rule, patch size, and spatial merge factor fixed.
The default $12$\,pt setting is used globally in all primary experiments.

The sweep shows a consistent score-token tradeoff at all three scales.
Shrinking the font from $12$\,pt to $10$\,pt reduces the visual token count but lowers the macro score.
Increasing the font to $14$\,pt raises the macro score but increases token cost.
The score change from $10$\,pt to $14$\,pt is $+2.9$ points at 4B, $+3.3$ points at 8B, and $+4.0$ points at 32B.
The direction is the same at every scale.

This behavior supports the transport interpretation.
Font size is not a hidden tuning parameter that changes unpredictably across tasks.
It acts as a global rendering knob controlling the precision-token tradeoff induced by within-patch aggregation.
The fixed $12$\,pt setting used in the main experiments provides a balanced operating point between smaller, more compressed renderings and larger, less compressed renderings.
Because the same global rendering rule is applied to all datasets and scales, the reported routing and foveation results are not produced by dataset-specific rendering choices.

\clearpage

%% file: Appendix/appx_latency.tex
\section{Runtime and Memory Measurements}
\label{sec:appx-latency}

\begin{table*}[!htbp]
\centering
\footnotesize
\renewcommand\arraystretch{1.12}
\caption{Runtime and peak-memory measurements on six representative benchmarks at the 4B scale. Latency is reported as seconds per sample. Peak memory is measured as GPU memory in GB. The routed strategy uses the label-free routing decision and reports the measured latency and memory of the selected path. The macro-average row is the unweighted average over the six datasets.}
\label{tab:appx-latency}
\vspace{0.5em}
\begin{tabular*}{\textwidth}{@{\extracolsep{\fill}} l l rrrr rrrr @{}}
\toprule[1.2pt]
& & \multicolumn{4}{c}{Latency (sec/sample)} & \multicolumn{4}{c}{Peak memory (GB)} \\
\cmidrule(lr){3-6} \cmidrule(lr){7-10}
Dataset & Length regime & LLM & VLM & FOV & Routed & LLM & VLM & FOV & Routed \\
\midrule
RACE-H    & short  & $0.16$ & $0.12$ & $0.13$ & $0.12$ & $8.7$  & $9.3$  & $9.3$  & $9.3$ \\
BoolQ     & short  & $0.08$ & $0.13$ & $0.13$ & $0.08$ & $8.2$  & $8.8$  & $8.8$  & $8.2$ \\
DBPedia   & short  & $0.05$ & $0.11$ & $0.11$ & $0.05$ & $8.1$  & $8.7$  & $8.7$  & $8.1$ \\
SciFact   & medium & $0.18$ & $0.17$ & $0.18$ & $0.17$ & $8.8$  & $9.3$  & $9.4$  & $9.3$ \\
HotpotQA  & medium & $1.16$ & $0.41$ & $0.42$ & $0.41$ & $9.6$  & $10.2$ & $10.2$ & $10.2$ \\
GovReport & long   & $9.59$ & $6.38$ & $6.34$ & $6.38$ & $12.9$ & $22.4$ & $22.4$ & $22.4$ \\
\midrule
Macro avg & ---    & $1.87$ & $1.22$ & $1.22$ & $1.20$ & $9.4$  & $11.5$ & $11.5$ & $11.3$ \\
\bottomrule[1.2pt]
\end{tabular*}
\vspace{0.6em}
\end{table*}

\vspace{0.5em}

\noindent\textbf{The purpose of this section:}
This section complements the token-count analysis in the main paper with wall-clock runtime and GPU-memory measurements.
The goal is not to claim that the visual path is always faster, but to test whether the token reductions reported in the main results can translate into practical latency reductions on representative short-, medium-, and long-context benchmarks.

\paragraph{Protocol.}
To obtain implementation-level runtime measurements, we instrument the evaluation pipeline by wrapping
\texttt{transformers.GenerationMixin.generate} to record per-call wall-clock
time using \texttt{time.perf\_counter} and peak GPU memory using
\texttt{torch.cuda.max\_memory\_allocated}, without changing the generation
outputs.
We measure on six benchmarks selected to span the input-length spectrum
of the 24-benchmark suite: RACE-H (multi-choice RC, short), BoolQ (binary,
short), DBPedia (multi-class, short), SciFact (claim verification, medium),
HotpotQA (multi-hop QA, medium), and GovReport (long-document summarization,
long).
All runs use Qwen3-4B/Qwen3-VL-4B-Instruct with $12$\,pt rendering, on a
single NVIDIA L40S GPU, in the same offline-inference configuration as the main
experiments.
For the routed strategy, we compute the label-free routing decision first and
then assign the measured latency of the selected path.
This avoids re-running identical generations and does not use task labels or
model scores.
The routing computation itself is a lightweight CPU-side calculation and is
not included in the table.

\paragraph{What the latency table shows.}
Three patterns are worth highlighting.
First, on \textbf{\textit{short-context}} tasks (RACE-H, BoolQ, DBPedia), all
strategies sit within $0.05$--$0.16$ sec/sample.
At this length range, neither the text nor the visual path uniformly dominates
wall-clock latency.
The routed strategy selects text for BoolQ and DBPedia and visual for RACE-H
under the label-free TE rule, yielding latency close to the faster measured
path on these short-context rows.
Second, on \textbf{\textit{longer-context}} tasks (HotpotQA, GovReport), the visual
path is meaningfully faster than the LLM path: $0.41$ vs. $1.16$ sec/sample on
HotpotQA ($2.8\times$ faster) and $6.38$ vs. $9.59$ sec/sample on GovReport
($1.5\times$ faster).
Long inputs make the LLM decoder pay for many text tokens, while the rendered
visual representation reduces the effective sequence length.
Third, the \textbf{\textit{macro-average}} routed latency is $1.20$ sec/sample, a
$36\%$ reduction relative to the LLM baseline ($1.87$).
It is also slightly lower than the always-visual baseline ($1.22$), although this difference is small after rounding.
Thus, the routed strategy preserves much of the runtime benefit of the visual
path on inputs where it is selected, while reverting to the cheaper text path
when rendering plus visual encoding would add overhead.

\paragraph{Foveation cost.}
On the six datasets in Tab.~\ref{tab:appx-latency}, FOV differs from VLM by at
most $0.04$ sec/sample.
This small runtime difference reflects that foveation uses a bounded crop
budget and is activated only for selected regions.
The runtime of FOV is therefore close to the runtime of the base visual path
on this representative subset, and it does not change the macro-level latency
picture.

\paragraph{Memory.}
Peak GPU memory is dominated by the longest input rather than by the routing
rule.
On GovReport, the VLM path retains rendered pages and visual tokens in memory,
reaching a peak of $22.4$ GB, while the LLM path peaks at $12.9$ GB.
On all other datasets, the two paths are within $0.5$--$1.0$ GB of each other.
Because foveation uses a bounded crop budget and does not retain extra crops
beyond the base visual representation, FOV peak memory is within $0.1$ GB of
VLM in every case.
None of the strategies exceeded the available memory on the L40S GPU in this
measurement.

\paragraph{Interpretation.}
These measurements support the efficiency claim made in the main text.
Token reduction is not merely an accounting statistic: on longer-context inputs, it translates into lower wall-clock latency.
At the same time, the measurements also show why routing is necessary.
The visual path is not uniformly faster on short inputs, and it can require more memory on long rendered documents.
The label-free decision rule therefore plays a practical role beyond accuracy, selecting the visual path where its token savings are likely to matter while retaining the text path where rendering overhead would dominate.

\clearpage

%% file: Appendix/appx_crossarch.tex
\section{Cross-Architecture Validation: InternVL3.5}
\label{sec:appx-crossarch}

\noindent\textbf{The purpose of this section:}
This section tests whether the task-dependent pattern of visual-versus-text behavior described by the framework is specific to the Qwen3-VL family used in the headline experiments, or whether it transfers to a different visual encoder.
We evaluate \textbf{\textit{InternVL3.5}}~\cite{DBLP:journals/corr/abs-2508-18265}, a vision-language model that pairs an InternViT visual encoder with the same Qwen3 LLM backbones used in our main experiments.
The 4B and 8B InternVL3.5 models use Qwen3-4B and Qwen3-8B respectively, and the 38B model uses Qwen3-32B (verified from the released checkpoint configuration).
This pairing is convenient because it keeps the LLM backbone matched at every scale and provides a diagnostic view of how changing the visual encoding stack affects the visual path.
The text path through the LLM backbone is identical between Qwen3-VL-$x$B and the InternVL3.5 variant that shares the same Qwen3 LLM, so visual-path differences primarily reflect the visual encoder.

\paragraph{Protocol.}
We evaluate InternVL3.5 using the same evaluation harness as the main experiments, adapting only the model-loading and inference wrappers required by InternVL.
Every benchmark uses identical prompt templates, scoring code, and rendering parameters (Roboto-Regular, $12$\,pt, adaptive page sizing capped at $928$\,px) as the Qwen3-VL runs.
The only deviation is that InternVL's native multi-image API (\texttt{chat(num\_patches\_list=...)}) is used to handle paginated inputs, which performs the model-specific dynamic tiling expected by InternVL.
We evaluate the full $24$-benchmark suite at three scales: Qwen3-VL-4B / InternVL3.5-4B (Qwen3-4B LLM), Qwen3-VL-8B / InternVL3.5-8B (Qwen3-8B LLM), and Qwen3-VL-32B / InternVL3.5-38B (Qwen3-32B LLM).
Foveation is not run on InternVL because the current foveation implementation assumes Qwen3-VL's patch grid and ViT feature layout.
We therefore restrict the cross-architecture comparison to baseline visual paths.
At each scale we use the scale-specific visual-friendly partition: $S_2$ at scale $\sigma$ is the set of datasets where the visual arm (max of VLM and FOV) wins LLM under the bolding rule of Tab.~\ref{tab:appx-perdataset} at that scale.
This gives $|S_2|=10$ at 4B, $|S_2|=8$ at 8B, and $|S_2|=11$ at 32B.
The per-scale composition shifts as the LLM backbone scales, as described below.

\vspace{0.6em}

\begin{table}[!htbp]
\centering
\footnotesize
\renewcommand\arraystretch{1.10}
\caption{Macro-average scores at the Qwen3-4B / InternVL3.5-4B scale. ``LLM'' is text-only Qwen3-4B. ``Qwen-VL'' is Qwen3-VL-4B, ``+Fov'' is Qwen-VL with foveation, and ``InternVL'' is InternVL3.5-4B baseline. $\Delta_{\mathrm{Int}}$ is InternVL minus Qwen-VL. The $S_2 / S_3$ split is the 4B visual-friendly partition ($10$ datasets visual-wins, $14$ datasets text-wins, See Tab.~\ref{tab:appx-perdataset}). }
\label{tab:appx-crossarch-4b}
\vspace{0.5em}
\begin{tabular*}{0.92\linewidth}{@{\extracolsep{\fill}} l c c c c c @{}}
\toprule[1.2pt]
Group & LLM & Qwen-VL & +Fov & InternVL & $\Delta_{\mathrm{Int}}$ \\
\midrule
All ($n\!=\!24$)                              & $45.9$ & $44.4$ & $44.8$ & $46.9$ & $+2.5$ \\
$S_2$ visual-friendly at 4B ($n\!=\!10$)      & $34.7$ & $44.0$ & $44.7$ & $43.8$ & $-0.2$ \\
$S_3$ text-friendly at 4B ($n\!=\!14$)        & $54.0$ & $44.6$ & $44.8$ & $49.1$ & $+4.5$ \\
\midrule
Multi-choice RC ($n\!=\!4$)                   & $57.9$ & $48.9$ & $49.2$ & $52.9$ & $+4.0$ \\
Binary ($n\!=\!2$)                            & $90.0$ & $80.4$ & $80.4$ & $91.9$ & $+11.5$ \\
Multi-class ($n\!=\!3$)                       & $65.5$ & $67.9$ & $68.0$ & $68.8$ & $+0.9$ \\
NLI/verif.\ ($n\!=\!2$)                       & $51.4$ & $56.7$ & $57.4$ & $57.1$ & $+0.4$ \\
Short-answer QA ($n\!=\!8$)                   & $28.2$ & $31.1$ & $31.9$ & $33.9$ & $+2.8$ \\
Summarization ($n\!=\!5$)                     & $33.1$ & $28.5$ & $28.6$ & $27.5$ & $-0.9$ \\
\bottomrule[1.2pt]
\end{tabular*}
\vspace{0.6em}
\end{table}

\begin{table}[!htbp]
\centering
\footnotesize
\renewcommand\arraystretch{1.10}
\caption{Macro-average scores at the Qwen3-8B / InternVL3.5-8B scale. The $S_2 / S_3$ split is the 8B visual-friendly partition: $S_2$ at 8B contains $8$ datasets where the visual path wins under the bolding rule of Tab.~\ref{tab:appx-perdataset} at 8B. This is the 4B set minus \texttt{drop} and \texttt{race}, since both flip to $S_3$ at 8B as the 8B LLM closes the gap. All scores are in percent.}
\label{tab:appx-crossarch-8b}
\vspace{0.5em}
\begin{tabular*}{0.92\linewidth}{@{\extracolsep{\fill}} l c c c c c @{}}
\toprule[1.2pt]
Group & LLM & Qwen-VL & +Fov & InternVL & $\Delta_{\mathrm{Int}}$ \\
\midrule
All ($n\!=\!24$)                              & $49.5$ & $48.4$ & $48.2$ & $49.3$ & $+0.9$ \\
$S_2$ visual-friendly at 8B ($n\!=\!8$)       & $44.5$ & $53.0$ & $52.2$ & $47.6$ & $-5.4$ \\
$S_3$ text-friendly at 8B ($n\!=\!16$)        & $52.0$ & $46.1$ & $46.1$ & $50.2$ & $+4.1$ \\
\midrule
Multi-choice RC ($n\!=\!4$)                   & $56.2$ & $53.6$ & $54.0$ & $54.2$ & $+0.7$ \\
Binary ($n\!=\!2$)                            & $90.8$ & $85.4$ & $85.2$ & $92.5$ & $+7.0$ \\
Multi-class ($n\!=\!3$)                       & $68.4$ & $72.1$ & $72.1$ & $70.2$ & $-1.9$ \\
NLI/verif.\ ($n\!=\!2$)                       & $57.9$ & $55.8$ & $55.3$ & $62.9$ & $+7.2$ \\
Short-answer QA ($n\!=\!8$)                   & $37.2$ & $37.2$ & $36.4$ & $38.3$ & $+1.1$ \\
Summarization ($n\!=\!5$)                     & $32.5$ & $30.3$ & $30.3$ & $27.8$ & $-2.5$ \\
\bottomrule[1.2pt]
\end{tabular*}
\vspace{0.6em}
\end{table}

\begin{table}[!htbp]
\centering
\footnotesize
\renewcommand\arraystretch{1.10}
\caption{Macro-average scores at the Qwen3-32B / InternVL3.5-38B scale. Both models share the same Qwen3-32B LLM backbone, so visual-path differences primarily reflect the InternViT-vs-Qwen-VL encoder swap. The $S_2 / S_3$ split is the 32B visual-friendly partition: $S_2$ at 32B contains $11$ datasets where the visual arm (max of VLM and FOV) wins LLM under the bolding rule of Tab.~\ref{tab:appx-perdataset} at 32B. Compared to 8B, \texttt{race}, \texttt{multifieldqa}, and \texttt{qmsum} enter $S_2$, while \texttt{drop} remains outside $S_2$ as the 32B LLM closes the precision gap. All scores are in percent.}
\label{tab:appx-crossarch-32b}
\vspace{0.5em}
\begin{tabular*}{0.92\linewidth}{@{\extracolsep{\fill}} l c c c c c @{}}
\toprule[1.2pt]
Group & LLM & Qwen-VL & +Fov & InternVL & $\Delta_{\mathrm{Int}}$ \\
\midrule
All ($n\!=\!24$)                              & $51.1$ & $47.9$ & $48.4$ & $48.8$ & $+0.9$ \\
$S_2$ visual-friendly at 32B ($n\!=\!11$)     & $40.8$ & $45.1$ & $46.2$ & $42.2$ & $-2.9$ \\
$S_3$ text-friendly at 32B ($n\!=\!13$)       & $59.7$ & $50.3$ & $50.3$ & $54.3$ & $+4.0$ \\
\midrule
Multi-choice RC ($n\!=\!4$)                   & $65.8$ & $57.6$ & $57.3$ & $58.5$ & $+0.9$ \\
Binary ($n\!=\!2$)                            & $91.8$ & $86.5$ & $86.7$ & $93.6$ & $+7.1$ \\
Multi-class ($n\!=\!3$)                       & $73.1$ & $73.4$ & $73.4$ & $72.9$ & $-0.5$ \\
NLI/verif.\ ($n\!=\!2$)                       & $62.5$ & $58.1$ & $57.3$ & $56.2$ & $-1.9$ \\
Short-answer QA ($n\!=\!8$)                   & $33.2$ & $31.3$ & $33.0$ & $34.0$ & $+2.7$ \\
Summarization ($n\!=\!5$)                     & $33.7$ & $32.0$ & $32.1$ & $29.2$ & $-2.8$ \\
\bottomrule[1.2pt]
\end{tabular*}
\vspace{0.6em}
\end{table}

\begin{table}[!htbp]
\centering
\footnotesize
\renewcommand\arraystretch{1.10}
\caption{Per-dataset cross-architecture gain $\Delta_{\mathrm{Int}} = \mathrm{InternVL} - \mathrm{Qwen\text{-}VL}$ at three scales. Positive values mean InternVL outperforms Qwen-VL on that dataset. The 32B column compares Qwen3-VL-32B against InternVL3.5-38B (both share the Qwen3-32B LLM). Tab.~\ref{tab:appx-crossarch-4b}, Tab.~\ref{tab:appx-crossarch-8b}, and Tab.~\ref{tab:appx-crossarch-32b} report the macro and family-level aggregates from the same data. The win/loss/tie counts cited in this section use a tie threshold of $|\Delta_{\mathrm{Int}}| \leq 0.5$. Per-dataset $\Delta_{\mathrm{Int}}$ values are rounded to one decimal from full-precision differences. Their 24-row average can differ from the macro $\Delta_{\mathrm{Int}}$ in Tabs.~\ref{tab:appx-crossarch-4b}--\ref{tab:appx-crossarch-32b} by up to $\pm0.1$ when the full-precision macro lies near a 1dp display boundary. This affects the 32B row of the macro tables: the 24-row average of the per-dataset values shown here is $+0.83$, which rounds to $+0.8$, while subtracting the 1dp macros in Tab.~\ref{tab:appx-crossarch-32b} gives $+0.9$.}
\label{tab:appx-crossarch-perdataset}
\vspace{0.5em}
\begin{tabular*}{0.65\linewidth}{@{\extracolsep{\fill}} l r r r @{}}
\toprule[1.2pt]
Dataset & 4B $\Delta_{\mathrm{Int}}$ & 8B $\Delta_{\mathrm{Int}}$ & 32B $\Delta_{\mathrm{Int}}$ \\
\midrule
\multicolumn{4}{@{}l}{\textit{Multi-choice RC}} \\
RACE-H        & $+1.1$  & $+2.2$  & $-0.1$  \\
QuALITY       & $-9.6$  & $-12.4$ & $-16.3$ \\
LogiQA        & $+9.7$  & $+7.7$  & $+13.8$ \\
DREAM         & $+14.8$ & $+5.2$  & $+6.2$  \\
\midrule
\multicolumn{4}{@{}l}{\textit{Binary}} \\
BoolQ         & $+9.7$  & $+5.1$  & $+3.2$  \\
IMDB          & $+13.4$ & $+8.9$  & $+11.1$ \\
\midrule
\multicolumn{4}{@{}l}{\textit{Multi-class}} \\
DBpedia       & $-1.2$  & $-4.0$  & $-3.1$  \\
Yahoo         & $-7.5$  & $-8.8$  & $-6.2$  \\
Yelp          & $+11.5$ & $+7.1$  & $+7.7$  \\
\midrule
\multicolumn{4}{@{}l}{\textit{NLI/verif.}} \\
ContractNLI   & $-1.1$  & $+9.0$  & $-2.7$  \\
SciFact       & $+2.0$  & $+5.4$  & $-1.1$  \\
\midrule
\multicolumn{4}{@{}l}{\textit{Short-answer QA}} \\
HotpotQA      & $-10.5$ & $-6.3$  & $-0.5$  \\
MuSiQue       & $-4.3$  & $-2.3$  & $+3.5$  \\
2WikiMHQA     & $-5.5$  & $-5.1$  & $+2.6$  \\
CoQA          & $+15.1$ & $+5.1$  & $+6.3$  \\
DROP          & $+38.6$ & $+36.1$ & $+9.8$  \\
MultiFieldQA  & $-12.0$ & $-15.7$ & $-16.9$ \\
QASPER        & $-14.8$ & $-21.8$ & $-3.7$  \\
ReCoRD        & $+15.5$ & $+18.7$ & $+20.3$ \\
\midrule
\multicolumn{4}{@{}l}{\textit{Summarization}} \\
CNN/DM        & $+7.6$  & $+3.8$  & $+4.0$  \\
XSum          & $+0.4$  & $-0.6$  & $-0.5$  \\
BillSum       & $-0.2$  & $-0.7$  & $-0.5$  \\
GovReport     & $-5.9$  & $-10.6$ & $-10.6$ \\
QMSum         & $-6.6$  & $-4.4$  & $-6.4$  \\
\bottomrule[1.2pt]
\end{tabular*}
\vspace{0.6em}
\end{table}

\paragraph{Macros are within $\pm 2.5$ points on the full suite.}
Tab.~\ref{tab:appx-crossarch-4b}, Tab.~\ref{tab:appx-crossarch-8b}, and Tab.~\ref{tab:appx-crossarch-32b} report macros at the 4B, 8B, and 32B scales; Tab.~\ref{tab:appx-crossarch-perdataset} reports the per-dataset $\Delta_{\mathrm{Int}}$ values that back the per-dataset claims below.
On the full $24$-benchmark macro, InternVL leads Qwen-VL by $+2.5$ at 4B, $+0.9$ at 8B, and $+0.9$ at 32B.
The per-dataset win-loss counts of InternVL versus Qwen-VL are $11$\,/\,$11$\,/\,$2$ ties at 4B, $12$\,/\,$12$\,/\,$0$ at 8B, and $11$\,/\,$9$\,/\,$4$ ties at 32B (Tab.~\ref{tab:appx-crossarch-perdataset}, $|\Delta_{\mathrm{Int}}|\leq 0.5$ tie threshold), so the two architectures are evenly matched at the dataset level at every scale even when the macro is close.
The framework's main qualitative finding, that VTC behavior is sharply task-dependent, replicates under InternViT at all three scales: on both architectures some tasks favor the visual path, others favor the text path, and compression ratio alone does not predict which case applies.

\paragraph{The visual-friendly partition itself is scale-dependent.}
$|S_2|$ shifts non-monotonically across scales: $10$ at 4B, $8$ at 8B, and $11$ at 32B.
The transitions are interpretable from the cost decomposition.
From 4B to 8B, \texttt{race} and \texttt{drop} flip to text-wins because the 8B LLM closes the precision-sensitive gap on these tasks.
From 8B to 32B, \texttt{race}, \texttt{multifieldqa}, and \texttt{qmsum} enter $S_2$.
\texttt{race} re-enters because the baseline visual path slightly exceeds the LLM at 32B, while \texttt{multifieldqa} and \texttt{qmsum} enter because the 32B foveation regime restores a usable coverage signal (App.~\ref{sec:appx-fov-fail}).
\texttt{drop} stays out as the 32B LLM remains stronger on this precision-sensitive task.
Eight datasets are visual-friendly at every scale and form the stable core of $S_2$: \texttt{2wikimhqa}, \texttt{hotpotqa}, \texttt{scifact}, \texttt{musique}, \texttt{yahoo}, \texttt{dbpedia}, \texttt{billsum}, and \texttt{qasper}.
We therefore split $S_2 / S_3$ per scale rather than imposing a single partition uniformly.

\paragraph{InternVL gives back where Qwen-VL leads, and recovers where Qwen-VL trails.}
The most informative comparison is the per-scale $S_2 / S_3$ rows.
The same direction holds at every scale: InternVL underperforms Qwen-VL on $S_2$ where Qwen-VL has a strong visual lead, and outperforms Qwen-VL on $S_3$ where Qwen-VL is below the LLM.
At 4B the pattern is mild ($S_2$: $-0.2$, $S_3$: $+4.5$), at 8B it is most pronounced ($S_2$: $-5.4$, $S_3$: $+4.1$), and at 32B it remains visible ($S_2$: $-2.9$, $S_3$: $+4.0$).
Two dataset families illustrate this contrast.
Long-document QA, which tends to be visual-friendly under Qwen-VL (\texttt{qasper}, \texttt{multifieldqa}, \texttt{hotpotqa}), shows large InternVL deficits at 4B ($-14.8$, $-12.0$, $-10.5$; Tab.~\ref{tab:appx-crossarch-perdataset}).
Precision-sensitive cloze and reasoning tasks where Qwen-VL trails the LLM (\texttt{drop}, \texttt{record}, \texttt{dream}) show large InternVL gains.
The two ViT families therefore have different but partially complementary strengths at every scale we tested.
This suggests that Qwen-VL's native visual encoder is better aligned with cross-page coverage on long documents, while InternViT appears stronger on short precision-sensitive inputs.

\paragraph{Cross-architecture intervention changes a different axis from foveation.}
At 4B, switching the visual encoder from Qwen3-VL's native ViT to InternViT yields $+2.5$ points on the full macro, compared with the $+0.4$ foveation gain reported in App.~\ref{sec:appx-fov-fail}.
At 8B, foveation is slightly negative ($-0.3$, consistent with the scale-dependent probe regimes of App.~\ref{sec:appx-fov-fail}), while the InternVL swap still yields $+0.9$.
However, the per-dataset profiles of the two interventions are very different.
Foveation produces small, sample-conservative changes bounded by its trigger gate, with the largest 4B gains on evidence-localized QA tasks such as \texttt{qasper} ($+2.9$), \texttt{hotpotqa} ($+1.0$), and \texttt{musique} ($+1.1$).
The InternVL swap is a high-variance intervention (Tab.~\ref{tab:appx-crossarch-perdataset}): $+38.6$ on \texttt{drop} and $+15.5$ on \texttt{record} at 4B, but $-14.8$ on \texttt{qasper} and $-12.0$ on \texttt{multifieldqa} at the same scale.
This asymmetry is consistent with the two operations being conceptually different.
Foveation refines the within-patch resolution of a fixed ViT according to a probe-derived cost map.
Switching architectures replaces the entire push-forward map with a differently trained one.
The two are not substitutes.

\paragraph{Per-family pattern is consistent with the cost decomposition.}
The family rows of Tab.~\ref{tab:appx-crossarch-4b}, Tab.~\ref{tab:appx-crossarch-8b}, and Tab.~\ref{tab:appx-crossarch-32b} show structured differences that hold across scales.
Binary classification favors InternVL substantially at every scale ($+11.5$ at 4B, $+7.0$ at 8B, $+7.1$ at 32B), and multi-choice RC and short-answer QA favor InternVL by smaller margins.
Summarization favors Qwen-VL at every scale ($-0.9$, $-2.5$, $-2.8$), in line with summarization being inter-cost-dominated and harder to repair by either intervention.
NLI/verif.\ is less stable across scales, moving from an InternVL advantage at 4B and 8B to a Qwen-VL advantage at 32B.
We do not claim a quantitative explanation of these per-family differences.
We report them so that readers using a different VLM family can anticipate where the visual path's behavior may shift.

\paragraph{What this section does and does not claim.}
This section reports a sample-paired cross-architecture replication.
The result we want to convey is that the framework's main object, a task-dependent visual-versus-text gap whose ordering is largely set by intra-patch precision and inter-patch coverage, is reproduced under a different ViT family at three backbone scales.
We do not claim that $C(x)$ predicts the exact magnitude of the InternVL gap on each dataset, since the magnitudes are architecture-specific.
We also do not claim that InternVL3.5 should replace Qwen3-VL as the headline backbone.
The two are within $\pm 2.5$ macro points, with a balanced win-loss split, and the choice between them is a deployment decision rather than a research claim.
What the cross-architecture data adds is direct evidence against the worry that VTC task-dependent behavior is an artifact of one specific ViT.
The same pattern appears under InternViT, with predictable family-level differences.

\clearpage